\documentclass[10pt,twocolumn,letterpaper]{article}

\usepackage[pagenumbers]{cvpr} 

\definecolor{cvprblue}{rgb}{0.21,0.49,0.74}
\usepackage[pagebackref,breaklinks,colorlinks,allcolors=cvprblue]{hyperref}

\usepackage[table]{xcolor}
\usepackage{amssymb} 
\usepackage{pifont}  

\usepackage{multirow}
\usepackage{tabularx}
\usepackage{booktabs}
\definecolor{excellent}{RGB}{75,185,95}
\definecolor{great}{RGB}{145,205,95}
\definecolor{good}{RGB}{235,225,95}
\definecolor{fair}{RGB}{255,200,95} 
\definecolor{poor}{RGB}{255,145,95}
\definecolor{verypoor}{RGB}{240,95,90}

\def\confName{CVPR}
\def\confYear{2027}

\title{GDD: A Remote Sensing Image Deepfake Detection and Manipulation Localization Dataset}

\author{Jacob Arndt \quad
        Philipe Dias \quad
        Debvrat Varshney \quad
        Sudarshan Srinivasan \quad
        Dalton Lunga \\\\
        Oak Ridge National Laboratory\\
        {\tt\small \{arndtjw, ambroziodiap, varshneyd, srinivasans, lungadd\}@ornl.gov} 
}

\begin{document}
\maketitle
\begin{abstract}

Verifying image authenticity is a critical need for a variety of applications and decision-making in the remote sensing (RS) domain, yet it is becoming more challenging as advancements in generative AI enable producing increasingly realistic synthetic overhead data. However, existing RS deepfake benchmarks prioritize sheer volume at the expense of visual realism and diversity, while also lacking rigorous out-of-distribution (OOD) testing protocols to evaluate real-world robustness. To address this, we introduce the Geospatial Deepfakes Dataset (GDD), a benchmark for RS deepfake detection and manipulation localization. Rather than relying on uncurated, redundant synthetic generation, GDD comprises images across diverse modern generative models and real image sources, alongside manipulations from diffusion inpainting, splicing, and copy-move operations. Crucially, GDD establishes an OOD evaluation protocol for deepfake detection comprising unseen real image sources and generative models. We benchmark five deepfake detectors and two localization models across intra- and cross-dataset regimes and reveal that detectors trained on high-volume, uncurated benchmarks overfit to dataset-specific artifacts, whereas training on GDD yields superior cross-dataset generalization despite its compact scale. This benchmarking also exposes critical failure modes in current state-of-the-art methods. Combined, GDD and its evaluation protocol establish a high-quality standard for developing and stress-testing detection and localization models in this safety-critical RS domain. The dataset is publicly available\footnotemark[1]. 

\end{abstract}

\footnotetext[1]{\url{https://huggingface.co/datasets/geodf/gdd}}
\section{Introduction}
\label{sec:intro}

Remote sensing (RS) imagery plays a critical role in a wide range of applications, including supply chain and transportation planning, infrastructure and environmental monitoring, disaster response, urban planning, and national security. As advances in generative AI models, particularly diffusion-based models, enable unprecedented levels of visual and spatial fidelity, the threat of localized satellite image manipulations and realistic yet synthetic imagery has escalated from theoretical risk to a pressing reality as documented in academic literature \cite{zhao2021deep, sui2024fldcf, tan2025rsfake}. Manipulated geospatial data, altered maps, and fake satellite imagery can now pass manual visual inspection, undermine trust and confidence in downstream geospatial analysis and decisions, and adversely impact automated pipelines. Consequently, developing robust RS deepfake detection and manipulation localization methods is a pressing need in the geospatial domain \cite{irwin_intelligence_2023, darpa_deepfake_2024, tucker_newest_2019}.


Developing robust detection frameworks relies fundamentally on standardized, high-quality benchmark datasets. However, RS deepfake and manipulation benchmarks remain scarce and severely limited in both scope and quality. Two existing large-scale RS deepfake detection datasets \cite{dmaer, tan2025rsfake} achieve sample counts in the millions, yet suffer from poor visual quality, lack diversity in generative models, and draw real imagery from only a single source. As a result, models trained on poorly curated, high-volume data overfit to trivial generation signatures and singular real image datasets, creating a false sense of performance while failing under real-world deployment. Furthermore, only two datasets support the task of manipulation localization \cite{arndt2026, sui2024fldcf}, and none provide controlled evaluation protocols with well-documented test splits to assess out-of-distribution (OOD) generalization across unseen real image sources and unseen generative models. Beyond these dataset-level constraints, a notable benchmarking gap persists, as recent state-of-the-art advancements in deepfake detection and localization from broader computer vision have yet to be systematically applied or evaluated in remote sensing contexts.


\begin{figure*}[t!]
    \centering
    \includegraphics[width=0.98\linewidth]{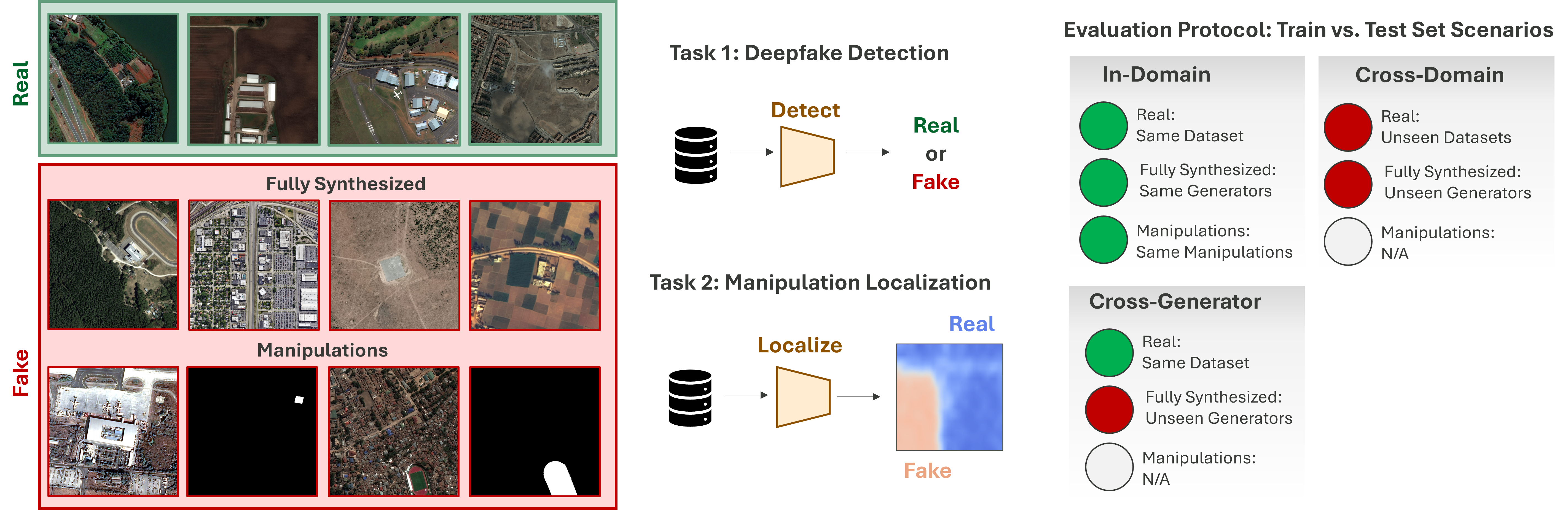}
    \caption{GDD consists of real high-resolution aerial and satellite imagery and fake imagery. The fake imagery includes fully synthesized examples created using generative models and locally manipulated images with ground truth masks developed through splicing, copy-move, and diffusion model inpainting. An evaluation protocol is proposed for improved assessment of deepfake detection  generalization under various settings. The dataset supports training and evaluation of both deepfake detection and manipulation localization methods.}
    \label{fig:overview}
\end{figure*}

To address these limitations, we propose the Geospatial Deepfakes Dataset (GDD) (Figure~\ref{fig:overview}), a RS deepfake detection and manipulation localization benchmark designed to prioritize diagnostic rigor over volume. Rather than relying on uncurated, redundant synthetic generation, GDD provides carefully controlled splits spanning real images from multiple sensors and datasets, diverse state-of-the-art generative models, and both fully synthesized and locally manipulated imagery. The manipulated imagery includes examples developed through traditional splicing and copy-move operations to more modern inpainting approaches that leverage diffusion models. In addition, GDD establishes an evaluation protocol comprising three testing splits to systematically assess deepfake detection generalization under OOD scenarios of {\em (i) unseen real image sources}, {\em (ii) unseen generative models}, and {\em (iii) combined unseen real image source and unseen generative models}. In total, GDD contains $18,410$ unique images to advance the training and evaluation of robust detectors and localization models in the remote sensing domain. The main contributions of this paper are summarized as follows:


\begin{itemize}
    \item We develop and release GDD, a RS deepfake detection and manipulation localization dataset featuring diverse real image sources, state-of-the-art generative models, and multiple manipulation operations.
    \item We establish an OOD evaluation protocol designed to systematically assess detector robustness against unseen real image sources and generative models.
    \item We establish intra- and cross-dataset baselines across five deepfake detectors and two manipulation localization models, exposing critical failure modes in current methods and limitations of current RS datasets.
    \item We demonstrate that GDD serves as both the most effective training benchmark for cross-dataset generalization and the most challenging test sets in RS. Specifically, we show that models trained on GDD generalize best to external datasets, whereas models trained on existing large-scale datasets degrade sharply on GDD.
\end{itemize}

\section{Related Work}
\label{sec:related-work}

\subsection{Datasets}
Two types of datasets are relevant to our work: deepfake detection datasets, consisting of fully synthesized AI-generated images; and manipulation localization datasets, consisting of images that have been partially modified. 
In Table~\ref{tab:related-datasets}, we  provide summary statistics of existing satellite image deepfake and manipulation datasets along with our proposed GDD dataset.

\begin{table*}[h]
    \centering
    \scriptsize
    \begin{tabular}{lccccrr}
        \toprule
        \textbf{Dataset}  & \textbf{Det.} & \textbf{Loc.} &  \textbf{Real Sources} &  \textbf{Generator/Manipulation Types}  & \textbf{\# Real} & \textbf{\# Fake}\\
        \midrule
        fmow-fake-small \cite{arndt2026}  & \textcolor{red}{\ding{55}} & \textcolor{green}{\checkmark} & fMoW & Splice, Splice Inpainting (RSPaint) & 30 & 30 \\
        FLDCF \cite{sui2024fldcf} & \textcolor{red}{\ding{55}} & \textcolor{green}{\checkmark} & Vaihingen, LoveDA & Inpainting (LaMa, Repaint) & 2,846 & 8,350 \\
        ``Novel Dataset'' \cite{ziulu2025} & \textcolor{green}{\checkmark} & \textcolor{red}{\ding{55}} & Google Earth & GAN & 0 & 1,792  \\
        FSI \cite{zhao2021deep} & \textcolor{green}{\checkmark} & \textcolor{red}{\ding{55}} & Google Earth & GAN & 4,032 & 4,032 \\
        RSFAKE-1M \cite{tan2025rsfake} & \textcolor{green}{\checkmark} & \textcolor{red}{\ding{55}} & fMoW & Diffusion & 500,000 & 500,000 \\
         DM-AER \cite{dmaer} & \textcolor{green}{\checkmark} & \textcolor{red}{\ding{55}} & Million-AID & GAN  & 140,000 & 1,000,000 \\
        \midrule
        \multirow{2}{*}{GDD (ours)} & \multirow{2}{*}{\textcolor{green}{\checkmark}} & \multirow{2}{*}{\textcolor{green}{\checkmark}} & fMoW, MillionAID, xBD & Diffusion, GAN, Splice Inpainting & \multirow{2}{*}{8,500} & \multirow{2}{*}{9,910} \\
         & & & DOTA, INRIA, OrbitalNet & Splice, Copy-move, Fully Regenerated & & \\
        \bottomrule
    \end{tabular}
    \caption{Remote sensing deepfake and image manipulation datasets. (\textit{Det.}: detection, \textit{Loc.}: localization)}
    \label{tab:related-datasets}
\end{table*}

\noindent\textbf{Satellite Image Deepfake Datasets.}
Generative adversarial networks (GANs) \cite{goodfellow2020generative} have been commonly used for synthesizing fake satellite images to create deepfake datasets. For example, the FSI \cite{zhao2021deep} and DM-AER \cite{dmaer} datasets use CycleGAN and StyleGAN2, respectively. In \cite{ziulu2025}, the FSI dataset is expanded by including additional fake images synthesized using pix2pix. More recently, diffusion models \cite{ho2020denoising} are being used to synthesize satellite imagery. The RSFAKE-1M \cite{tan2025rsfake} dataset, for example, is the largest satellite image deepfake dataset available and consists of $500,000$ fake images synthesized using different conditional diffusion models \cite{khanna2024diffusionsat, sastry2024geosynth, espinosa2023generate, tang2024crs} initialized from pretrained Stable Diffusion \cite{rombach2022high} checkpoints and finetuned on satellite or aerial imagery.

\noindent\textbf{Satellite Image Manipulation Localization Datasets.}
Image manipulation localization datasets provide pixel-level annotations for regions altered through editing operations such as copy-move, splicing, and inpainting. In the broader computer vision literature, several well-established benchmarks exist for general image content, including the Columbia \cite{hsu2006detecting}, CASIA \cite{dong2013casia}, IMD2020 \cite{novozamsky2020imd2020}, and AutoSplice \cite{jia2023autosplice} datasets. 
In contrast, the landscape of satellite image manipulation datasets remains significantly underdeveloped. To our knowledge the FLDCF dataset \cite{sui2024fldcf} and the fmow-fake-small \cite{arndt2026} are the only publicly available datasets in this domain. FLDCF includes inpainting examples using RePaint and LaMa, while fmow-fake-small includes splices and spliced inpainting using the RSPaint model. The work of \cite{yarlagadda2018satellite} introduces object splices for satellite imagery with variations in object size and category (e.g., aircraft, clouds), but unfortunately it is not publicly available.

\begin{figure*}[ht]
    \centering
    \begin{subfigure}{0.49\textwidth}
        \centering
        \includegraphics[width=\linewidth]{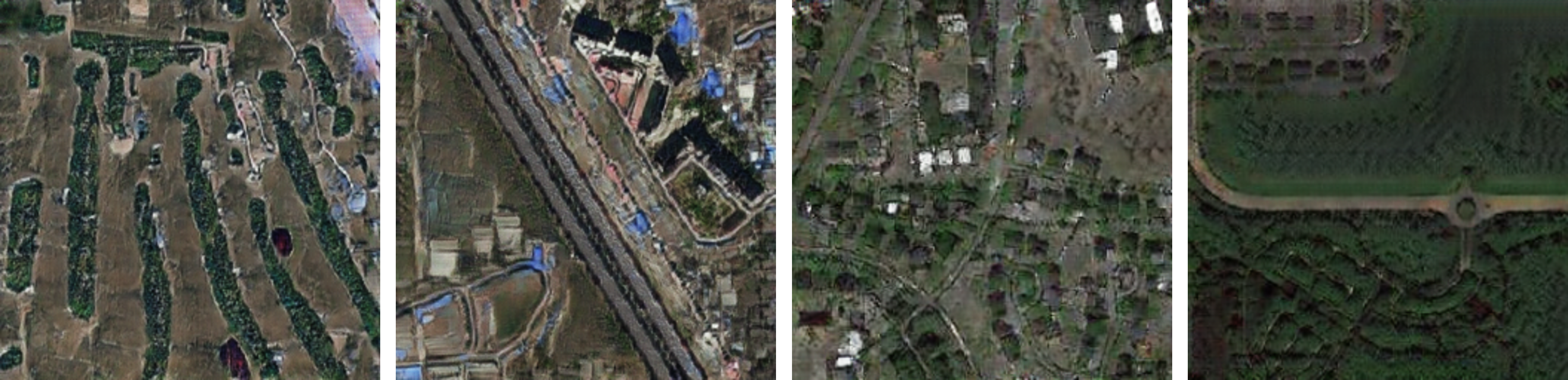}
        \caption{FSI \cite{zhao2021deep} (CycleGAN)}
    \end{subfigure}
    \hfill
    \begin{subfigure}{0.49\textwidth}
        \centering
        \includegraphics[width=\linewidth]{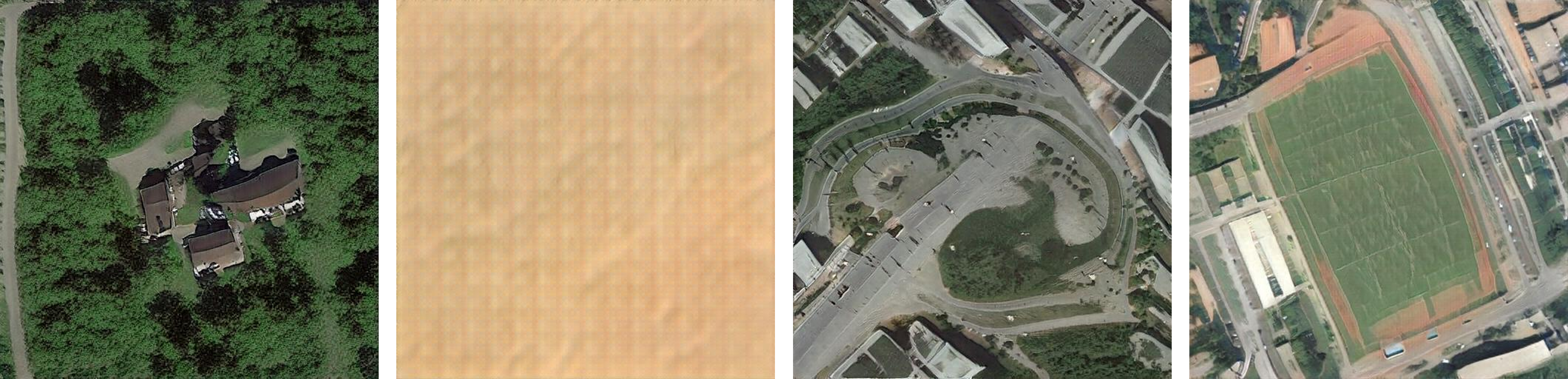}
        \caption{DM-AER \cite{dmaer} (StyleGAN2)}
    \end{subfigure}
    
    \vspace{0.20cm}
    \begin{subfigure}{0.49\textwidth}
        \centering
        \includegraphics[width=\linewidth]{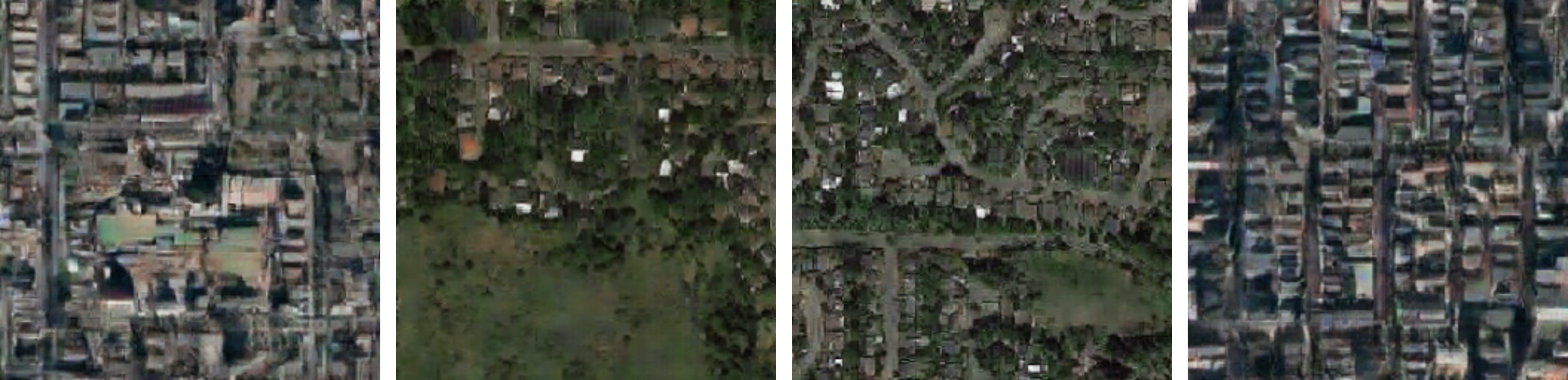}
        \caption{``Novel Dataset'' \cite{ziulu2025} (pix2pix)}
    \end{subfigure}
    \hfill
    \begin{subfigure}{0.49\textwidth}
        \centering
        \includegraphics[width=\linewidth]{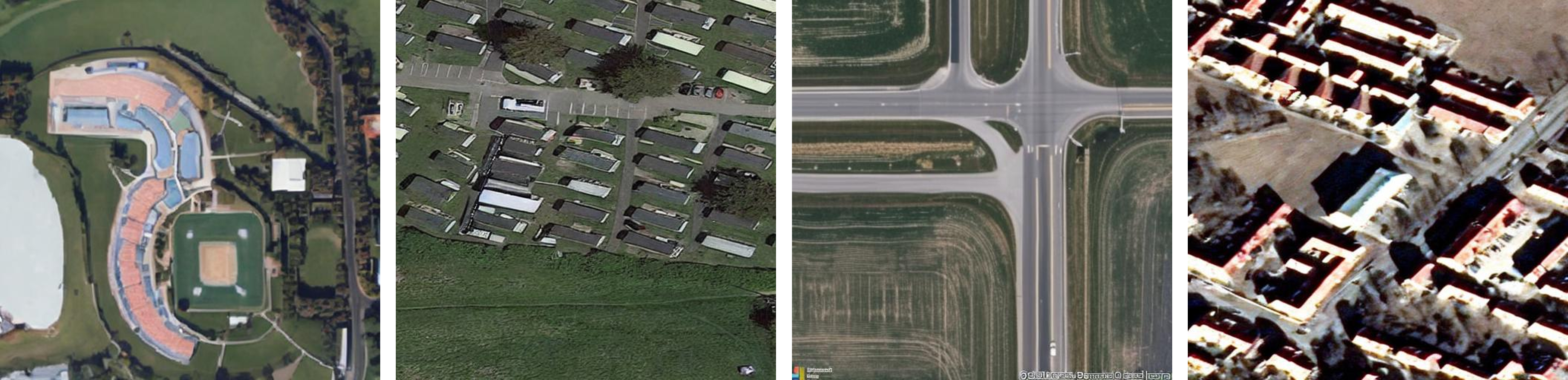}
        \caption{RSFAKE-1M Dataset \cite{tan2025rsfake} (Diffusion based models)}
    \end{subfigure}
    
    \vspace{0.20cm}
    \begin{subfigure}{0.49\textwidth}
        \centering
        \includegraphics[width=\linewidth]{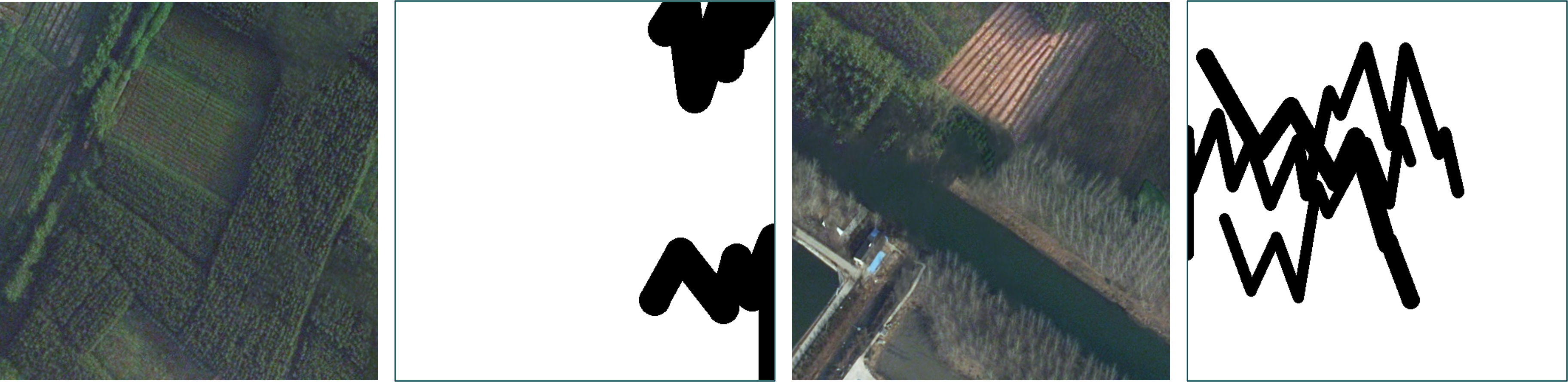}
        \caption{FLDCF \cite{sui2024fldcf} (Repaint). Image and ground truth mask.}
    \end{subfigure}
    \hfill
    \begin{subfigure}{0.49\textwidth}
        \centering
        \includegraphics[width=\linewidth]{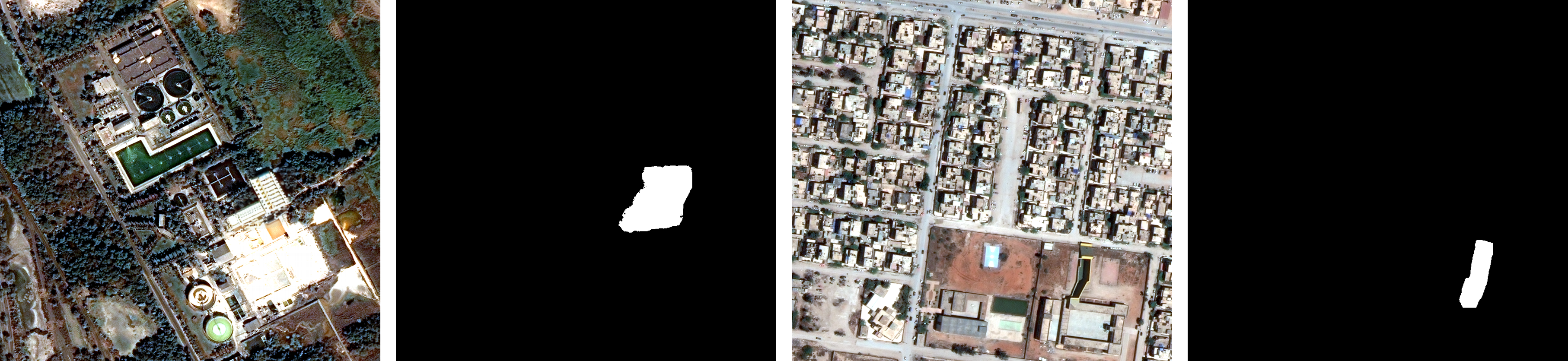}
        \caption{fMoW-Fake-Small \cite{arndt2026} (RSPaint). Image and ground truth mask.}
    \end{subfigure}
    \caption{Visual examples of fake images from existing satellite image deepfake and image manipulation localization datasets. While existing datasets establish the overhead view correctly, noticeable texture and semantic artifacts are visible. For existing manipulation localization datasets (Subfigures e. and f.), either the masks are unrealistic or the dataset is too small for meaningful benchmarking.}
    \label{fig:examples-related-work}
\end{figure*}

While the above-mentioned satellite image deepfake and manipulation datasets have enabled incremental progress, they all have notable limitations. First, as illustrated in Figure~\ref{fig:examples-related-work}, many examples in these datasets exhibit relatively low perceptual quality, including blurred textures, repetitive patterns, and inconsistencies in fine-grained spatial details. Second, some of the datasets are small in size (e.g. fmow-fake-small and ``Novel Dataset''), limiting their utility in robust benchmarking and training of generalizable models. Third, many of the datasets use only a single generative model or class of generative model (either GAN or diffusion) to create fake images (e.g. DM-AER, FSI, ``Novel Dataset'', RSFAKE-1M), limiting the dataset's utility for evaluating cross-generator generalization performance. Fourth, manipulation types and masks do not present realistic manipulation scenarios (see Figure~\ref{fig:examples-related-work}). For example, FLDCF includes manipulations using randomized masks and the inpainting attempts to recover the reference image rather than inpainting specific objects. Fifth, the existing datasets lack thorough prescribed evaluation protocols, meaning extensive generalization tests are not possible. These limitations at the dataset level reveal a significant research gap: in the remote sensing domain, there is insufficient data and evaluations to reliably measure progress in deepfake detection and manipulation localization.

\subsection{Benchmarking}


\textbf{Detection.} Deepfake detection is a binary classification task which aims to determine whether a given image is real or fake. Existing approaches can be broadly categorized based on the type of features they exploit. \emph{High-level feature-based methods} focus on learning semantic cues. For instance, CNNDet \cite{wang2020cnn} trains a ResNet-50 on ProGAN-generated images and improves generalization through augmentations such as JPEG compression and Gaussian blur. UnivFD \cite{ojha2023towards} leverages a frozen pretrained CLIP-ViT encoder to construct a database of real and fake image embeddings, performing classification via nearest-neighbor search. In turn, SimLBR \cite{dhakal2026simlbr} utilizes a frozen pretrained DINOv3 model to extract embeddings and trains only a small MLP for classification. During training, embeddings from real images are modified by adding controlled amounts of fake image embeddings, encouraging a tighter decision boundary around real images. \emph{Low-level feature-based methods} instead rely on subtle artifacts introduced during image synthesis, such as frequency-domain features \cite{frank2020leveraging}, texture \cite{liu2020global, zhong2023patchcraft}, noise patterns \cite{liu2022detecting}, and camera characteristics. \emph{Hybrid methods} aim to combine the strengths of both paradigms by jointly modeling semantic content and low-level signals. Co-Spy \cite{cheng2025co} and AIDE \cite{yan2025a}, for example, integrate texture, noise, and high-level representations within unified architectures, demonstrating improved robustness and generalization across diverse manipulation methods.

\noindent\textbf{Localization.}
Manipulation localization aims to identify which pixels in a given image have been modified. 
Similar to detection, these approaches can be categorized based on whether they rely on low-level forensic cues, high-level semantic features, or a combination of both. State-of-the-art approaches such as TruFor \cite{guillaro2023trufor}, UnionFormer \cite{li2024unionformer}, and the approach of \citet{su2024novel} all use dual-stream approaches leveraging both low-level cues and high-level semantic features. Low-level features are obtained from a NoisePrint model \cite{cozzolino2019noiseprint}, steganalysis rich model (SRM) filters \cite{fridrich2012rich}, or constrained  \cite{bayar2018constrained} or Bayer \cite{bayar2016deep} convolution layers, while high-level semantic cues are learned from the RGB input or obtained from pretrained image encoders. MMFusion \cite{triaridis2024exploring} adapts the TruFor \cite{guillaro2023trufor} model with a fusion strategy and incorporates additional forensic features derived from NoisePrint++, SRM, and Bayer convolution. 

\noindent\textbf{Evaluation Protocols.}
A significant challenge in deepfake detection and manipulation localization is generalizing to images produced by unseen generators. Evaluation protocols typically elucidate detector generalization limitations through a series out-of-distribution tests, offering insights into how a model could behave in unpredictable real-world situations. Cross-generator evaluations, as performed in \cite{wang2020cnn} for example, test a detector on images produced from unseen generators. In the manipulation localization setting, a similar procedure termed cross-forgery evaluation is typically performed. The evaluation protocol 2 in \cite{he2021forgerynet} is an example, where the model is tested on unseen manipulations. Most recently, an extensive evaluation protocol was proposed in \cite{tan2026veritas} that includes evaluating performance for in-domain, cross-generator, cross-forgery, and cross-domain scenarios. The cross-domain situation extends the cross-generator and cross-forgery cases by including real images from unseen datasets during training. 
\section{Dataset Construction}
\label{sec:dataset-construction}


\subsection{Dataset Source Imagery}

Our dataset is derived from several existing high-resolution remote sensing image datasets including Functional Map of the World (fMoW) \cite{christie2018functional}, Million-AID \cite{long2021creating}, OrbitalNet \cite{swan2025orbital}, DOTA \cite{xia2018dota,ding2021object}, xBD \cite{gupta2019xbd}, and the INRIA \cite{inria-dataset} dataset. By combining imagery from these sources, the dataset exhibits substantial diversity in terms of acquisition sensor, platform, viewing geometry, image formats, compression artifacts, spatial resolutions, geographic regions, and landcover types. 
See Supplementary~\ref{appendix:dataset-comp}, Table~\ref{tab:real-dataset-info} for further details regarding these datasets.

\subsection{Real Images}
To populate the ``real'' portion of our dataset, we randomly sample images across all sources. Qualitative examples of real image sources included in GDD can be found in Supplementary~\ref{appendix:visual-examples} Figure~\ref{fig:qual-real-examples}.


\subsection{Full Image Synthesis}
Selecting generators for a deepfake detection dataset is a critical design choice, as generative models introduce artifacts and fingerprints that classification models can exploit. As a result, cross-generator generalization remains a major challenge in deepfake detection \cite{ojha2023towards}, where models trained on a narrow set of generators often fail when evaluated on unseen generators. Furthermore, the rapid progress and evolution of generative vision models necessitates deepfake benchmarks that incorporate these modern architectures and advancements to maintain relevance. To address these challenges, we construct the synthetic portion of our dataset using a diverse suite of recent generators spanning disparate architectures, training datasets, and control mechanisms.

The fully synthesized portion of our dataset is created using eight recent image generators, including a single GAN and several diffusion models. For GAN-based synthesis, we include R3GAN \cite{huang2024gan}. For diffusion-based synthesis, we include general-purpose models (Stable Diffusion 3.5 \cite{esser2024scaling}, Flux2-dev \cite{blackforestlabs2025flux2}, and Flux2-klein-base-4b \cite{blackforestlabs2025flux2}) alongside remote sensing specific models (RSPaint \cite{immanuel2025tackling}, NeDS \cite{zheng2025neural}, DiffusionSat \cite{khanna2024diffusionsat}, and GeoSynth \cite{sastry2024geosynth}). Together, these models encompass diverse architectural designs (U-Net and Diffusion Transformer), training formulations (denoising diffusion and flow matching), and conditioning strategies (acquisition metadata embedding, ControlNet).

We deploy these generators under two settings: zero-shot synthesis and domain-specific training. Four models (SD 3.5, Flux2-dev, Flux2-klein-base-4b, and RSPaint) are used zero-shot without additional finetuning. The remaining four models are trained from scratch or finetuned on remote sensing imagery. R3GAN is trained from scratch on the fMoW dataset. NeDS is a finetune of Stable Diffusion 2.1 on the xBD dataset. DiffusionSat and GeoSynth are a finetune of Stable Diffusion 2.1 on an in-house dataset of WorldView-2, WorldView-3, GeoEye-1, and QuickBird-2 imagery. For all text-to-image models trained on satellite data, text conditioning is provided by captions generated via Qwen3-VL-8b-Instruct \cite{bai2025qwen3}. We design a domain-specific prompt (see Supplementary~\ref{appendix:text-prompt}) to produce detailed descriptions of objects, land cover, and spatial context. For zero-shot generation, prompts are randomly sampled from the pool of captions generated for the fMoW dataset. Examples of fully synthesized images included in GDD can be found in Supplementary~\ref{appendix:visual-examples} Figure~\ref{fig:qual-generated-examples}. 


\subsection{Manipulation} We organize the manipulations in GDD into three top-level \textit{types}: copy-move, splice, and fully regenerated images. Copy-move manipulations are further divided into \textit{variants}: object-level and simple operations. Splicing manipulations are divided into \textit{variants}: object-level, simple, and inpainting-based manipulations. Fully regenerated images form a separate category, as they do not involve compositing of an existing source image. The fully-regenerated and splice-inpainting examples leverage the pretrained RSPaint \cite{immanuel2025tackling} model. These different approaches for producing inpainting manipulations with a diffusion model cover the two different paradigms also outlined in \cite{mareen2026tgif2}. Given these \textit{types} and \textit{variants}, GDD contains a total of six possible type-variant manipulations, including splice-simple, splice-object, copy-move-simple, copy-move-object, splice-inpainting, and fully-regenerated. By including traditional splicing and copy-move operations alongside modern generative inpainting, GDD enables evaluation across both classical pixel-level forgery and modern diffusion-based inpainting and synthesis. Furthermore, the dataset's manipulation variants span from low-level, non-semantic random manipulations to targeted semantic alterations. Visual examples of manipulations included in GDD can be found in Supplementary~\ref{appendix:visual-examples} Figure~\ref{fig:qual-manipulation-examples}.

For manipulations that involve compositing existing real image content, we define a binary source mask $M_s$ that specifies the pixels or region to be extracted from a source image $I_s$. The selected region is then inserted into a target image $I_t$ to produce a manipulated image $I_m$. A binary ground-truth mask $M_m$ indicates the pixels in the manipulated image that were modified. 

\subsubsection{Manipulation Types}

\textbf{Copy-Move.}
For copy-move manipulations, we copy content from an image and paste it at a different location within the same image. The source image and target image are the same $I_s = I_t$. 

\noindent\textbf{Splice.}
For splicing, we copy pixels from a source image and paste them into a different image. The source image and target image are different $I_s \neq I_t$.

\noindent\textbf{Fully Regenerated.} 
For fully regenerated images, we directly use the output image of a diffusion-based inpainting model as the manipulated image. In this case, the ground-truth manipulation mask $M_m$ corresponds to the targeted region of the inpainting operation that is provided to the diffusion model.

\subsubsection{Manipulation Variants}

\textbf{Simple.}
For simple manipulations, the source mask $M_s$ is generated by randomly selecting a square region of randomly selected size from the source image. The selected region is then inserted into a target image at a random location. This procedure is used to generate both simple copy-move and simple splice manipulations. Thus, simple manipulations represent arbitrary region-level modifications rather than the transfer of identifiable semantic entities.

\noindent\textbf{Object.}
For object-based manipulations, the source mask $M_s$ corresponds to semantically meaningful regions (objects or stuff classes) extracted from the source imagery. We curate a diverse collection of objects and stuff regions from the source imagery using the Segment Anything Model (SAM). The resulting collection spans a range of semantic categories (e.g. cars, planes, bridges, roads, forest, and agriculture fields) and object sizes (small, medium, large, extra large) and is used to generate both object-level copy-move and object-level splice manipulations.

\noindent\textbf{Inpainted.}
The inpainted variant applies only to the splice type. Given an image $I_t$ and a user-defined mask specifying a region to be modified, we first apply a diffusion-based inpainting model to generate a new image with inpainting modifications applied to the masked region. This produces a fully synthesized inpainted image. We then extract the diffusion model generated content corresponding to the masked region and splice it into the original authentic image $I_t$. The user-defined mask therefore specifies both the region used for inpainting and the ground-truth manipulation mask $M_m$. We use the RSPaint \cite{immanuel2025tackling} model and follow the inpainting and splicing processes described in \cite{arndt2026}.

\subsubsection{Manipulation Post-Processing}
To improve the visual quality of object copy-move, object splicing, and inpainting splicing manipulations, we apply a blending operation between the target image and the splice/copy-move content. The blending step softens the boundary between the splice/copy-move content and target image background, creating a more natural transition between manipulated and authentic image regions. The target mask (location in target image $I_t$ where pixels are to be inserted) is first normalized to $[0, 1]$ and smoothed with a Gaussian filter to produce a feathered mask $\mathbf{\alpha}$. The source splice/copy-move content $I_\text{src}$ and target image $I_t$ are then blended using the feathered mask to create the final manipulated image: $I_m = \mathbf{\alpha} \odot I_\text{src} + (1 - \mathbf{\alpha}) \odot I_t$.

\section{Evaluation Protocol}
\label{sec:evaluation_protocol}
We develop three test splits to enable systematic evaluation of deepfake detection and generalization. The definitions of these test splits are similar to \cite{tan2026veritas} and are:

\begin{itemize}
    \item \textbf{in-domain:} \textit{Real images} are sampled from the same datasets as the training split. \textit{Fully synthesized images} are generated from the same generators as the training split. \textit{Manipulated images} are created using the same source datasets and manipulation types as those found in the training split.
    \item \textbf{cross-generator:} \textit{Real images} are sampled from the same datasets as the training split. \textit{Fully synthesized images} are generated with unseen/new generators that are not present in the training split.
    \item \textbf{cross-domain:} \textit{Real images} are sampled from different datasets not present in the training split. \textit{Fully synthesized images} are generated with unseen/new generators that are not present in the training split. 
\end{itemize}

\section{Dataset Composition and Analysis}
\label{sec:dataset-composition}

\textbf{Dataset Statistics.} The dataset consists of 18,410 unique images total, partitioned across training, validation, and the three testing splits described in Section~\ref{sec:evaluation_protocol}. The deepfake detection portion of the dataset includes 16,000 unique images total (8,500 real, 7,500 fake). The manipulation localization portion of the dataset consists of 4,910 images total (2,410 unique manipulated images and 2,500 authentic images drawn from the real portion of the GDD). We present a detailed breakdown of real image sources, generators, manipulation types, and image counts per split in Supplementary~\ref{appendix:dataset-comp}. 

\begin{figure}[h!]
    \centering
    \begin{subfigure}[t]{0.23\textwidth}
        \centering
        \includegraphics[width=\linewidth]{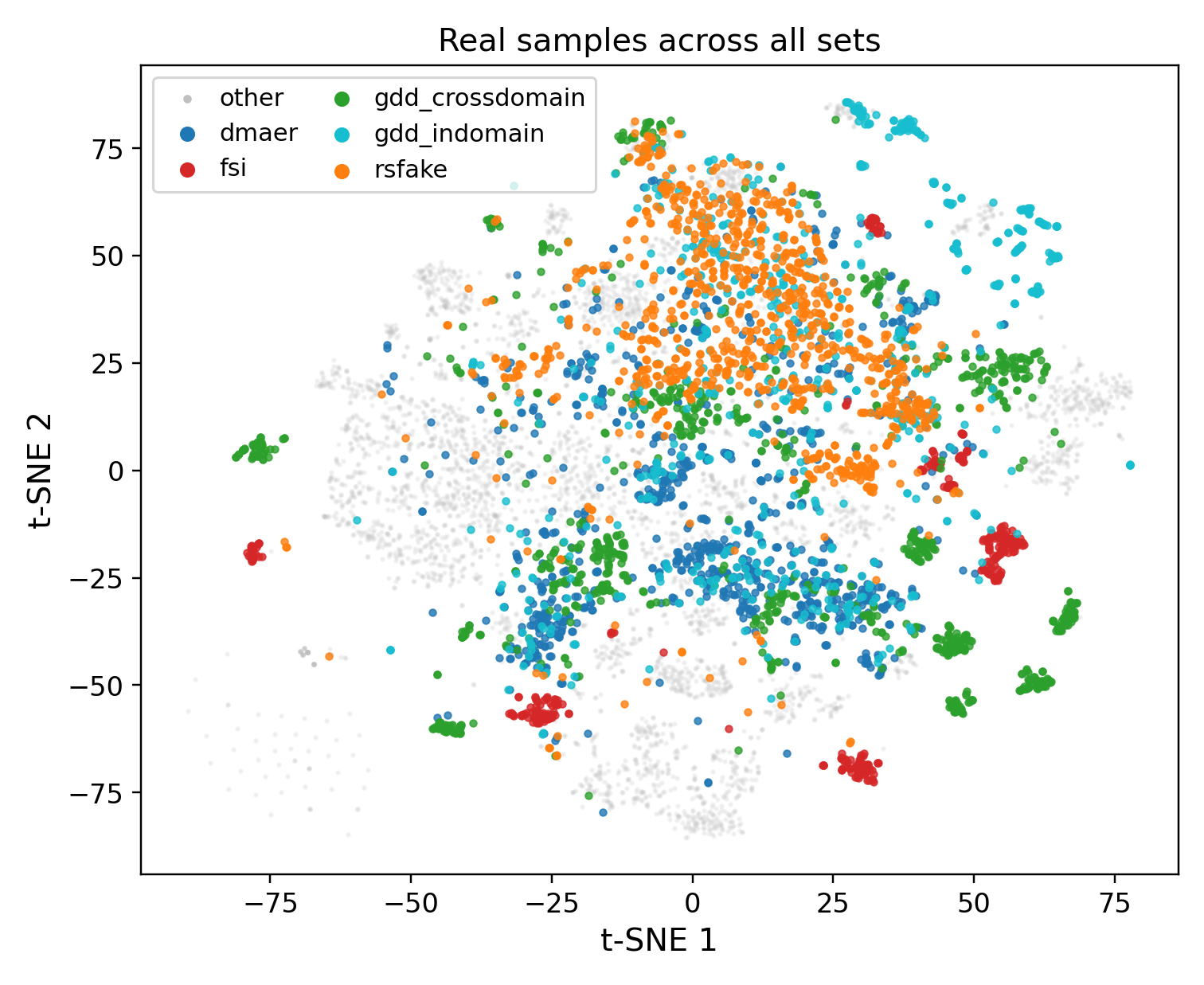}
        \caption{Real images by dataset}
        \label{fig:tsne_real}
    \end{subfigure}
    \hfill
    \begin{subfigure}[t]{0.23\textwidth}
        \centering
        \includegraphics[width=\linewidth]{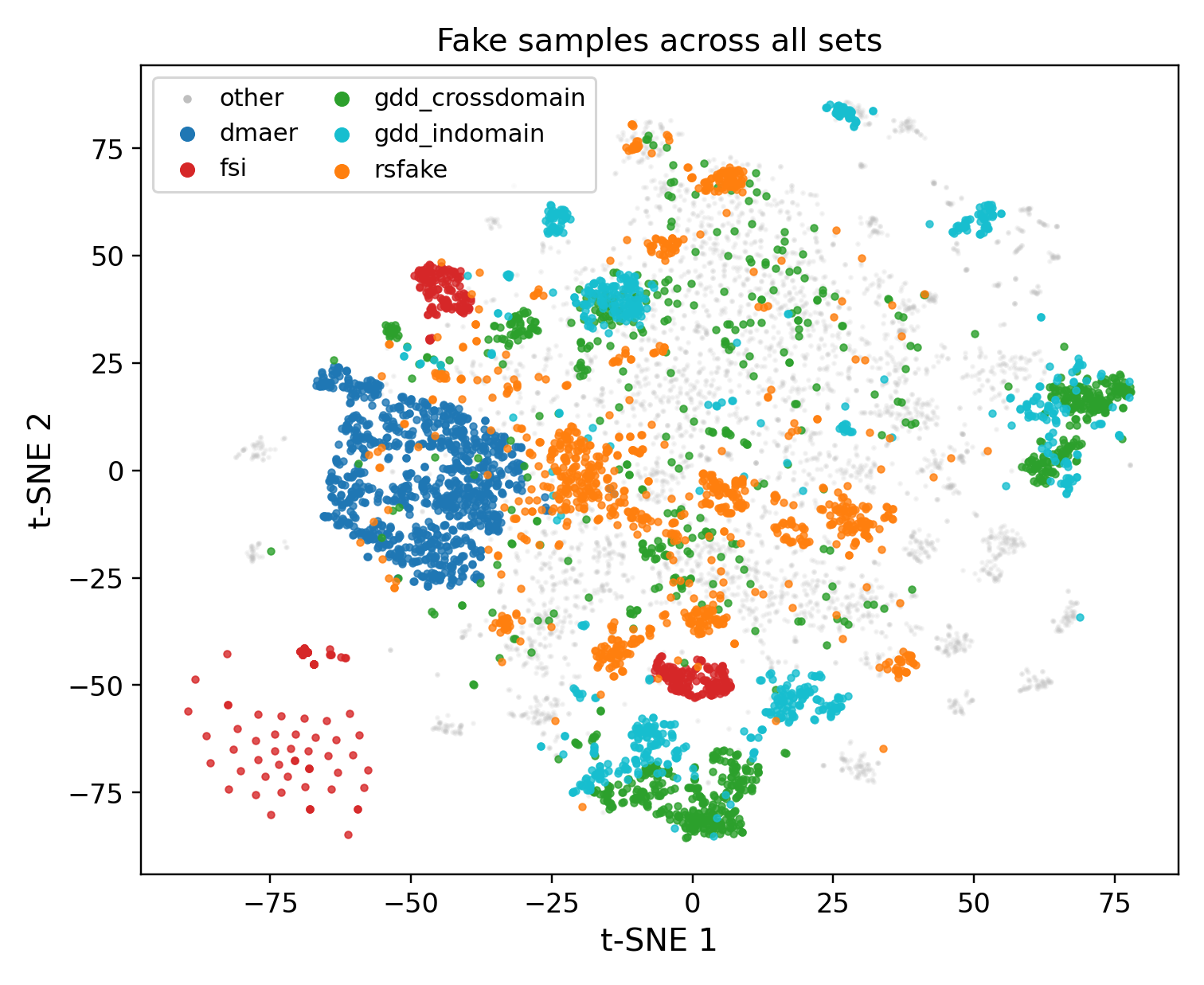}
        \caption{Fake images by dataset}
        \label{fig:tsne_fakes}
    \end{subfigure}
    \caption{t-SNE visualization comparing GDD feature distributions and other RS deepfake detection datasets.}
    \label{fig:tsne-main-comparison}
\end{figure}

\noindent\textbf{Diversity and Distribution.}
In Figure~\ref{fig:tsne-main-comparison}, we present t-SNE projections of DINOv3-Large-SAT [CLS] token embeddings for real and fake imagery in GDD alongside existing RS deepfake detection datasets. The real portion of GDD exhibits a substantially wider spread and multi-modal distribution compared to FSI, DM-AER, and RSFAKE-1M. This improved spread is a result of GDD's multi-source composition, incorporating imagery across diverse real RS datasets comprised of different sensors, land cover types, spatial resolutions, and acquisition conditions. Conversely, existing RS deepfake datasets are constrained by a single or two sources, resulting in limited breadth. Similarly, within the synthetic partition, GDD's fully synthesized fakes span a broad and almost continuous region of the feature space. In contrast, fakes from FSI and DM-AER collapse into distinct, isolated clusters (1 to 3 modes), exposing limitations in diversity of fake images and generative models. By incorporating a variety of modern diffusion and GAN architectures, GDD prevents bias and a small number of modes that are inherent in prior benchmarks. Additional t-SNE visualizations and analysis are found in Supplementary~\ref{appendix:tsne}.

\section{Experiments}
\label{sec:experiments}
\subsection{Setup}

\textbf{Deepfake Detection Models.} We consider five state-of-the-art detectors in the benchmark including UnivFD \cite{ojha2023towards}, AIDE \cite{yan2025a}, Co-SPY \cite{cheng2025co}, and SimLBR \cite{dhakal2026simlbr}. In addition, we propose a simple detector (hereafter referred to as DINOv3-sup), composed of a frozen DINOv3 \cite{simeoni2025dinov3} backbone with a trainable classification head consisting of a single linear layer. We use the DINOv3-L checkpoints trained on LVD.


\noindent\textbf{Manipulation Localization Models.} We train and evaluate two state-of-the-art manipulation localization models including TruFor \cite{guillaro2023trufor} and MMFusion \cite{triaridis2024exploring}.

To aid in interpretation of the results, we classify the performance of detectors and localization models using the following color-coded scale:
\colorbox{excellent}{\textcolor{black}{\textbf{Excellent}}}: $\geq$0.99,
\colorbox{great}{\textcolor{black}{\textbf{Great}}}: 0.90-0.99,
\colorbox{good}{\textcolor{black}{\textbf{Good}}}: 0.80-0.90,
\colorbox{fair}{\textcolor{black}{\textbf{Fair}}}: 0.70-0.80,
\colorbox{poor}{\textcolor{black}{\textbf{Poor}}}: 0.60-0.70, and
\colorbox{verypoor}{\textcolor{black}{\textbf{Very Poor}}}: $<$0.60.

\subsection{Cross-Dataset Deepfake Detection}
In the cross-dataset setting, we benchmark the five detectors across four datasets including FSI, DM-AER, RSFAKE-1M, and GDD. For each setting, a detector is trained on one dataset and evaluated on the test splits of all remaining datasets. For the GDD, we use the in-domain test split. In contrast to RSFAKE-1M and GDD, DM-AER and FSI have imbalanced real/fake dataset composition. To enable a fair comparison across datasets, we report the balanced accuracy defined as the average of the accuracies on real and fake samples: $(Acc_{Real}+Acc_{Fake})/2$, where $Acc_{Real}$ and $Acc_{Fake}$ denote the respective accuracies. Table~\ref{tab:cross_dataset_generalization_deepfake_detection} summarizes the results obtained for each detector under in-dataset and cross-dataset testing settings. 

\noindent\textbf{Model Oriented Analysis.} 
On average, the DINO-based SimLBR and DINOv3-sup outperform the CLIP-based models (AIDE, Co-Spy and UnivFD) across all scenarios. This indicates DINOv3-derived features are more informative for remote sensing deepfake detection than CLIP-derived ones, which we conjecture is due to 
its pretraining scheme that better captures fine-grained features as compared to CLIP's image-level contrastive learning. This is particularly evident by contrasting the performances of UnivFD and DINOv3-sup, as both consist of simple classification heads on top of extracted features.

AIDE's and Co-Spy's combination of CLIP-derived features with low-level DCT and VAE-based local features enables near perfect scores for within-dataset train and test scenarios. However, both exhibit a tendency to overfit as poor accuracies are observed for cross-dataset tests. For AIDE, this is particularly evident when trained on FSI, a dataset of notoriously poor quality and low diversity of fake samples. Co-Spy reveals a strong sensitivity to dataset imbalance, with poor results across all test datasets when trained on DM-AER. SimLBR is the top performer when considering balanced accuracy, but with the caveat of reduced generalization to unseen real image sources as discussed in Section~\ref{sec:gdd_detection}. 


\noindent\textbf{Training Dataset Transferability.} As shown in the aggregate accuracies across all models according to training dataset (last column in Table~\ref{tab:cross_dataset_generalization_deepfake_detection}), GDD's diversity in real image sources and combination of both GAN- and diffusion-based generators enables training models with higher generalization. 
While GDD contains only approximately 2\% as many images as RSFAKE-1M, all detectors show improved cross-dataset generalization when trained on GDD compared to the RSFAKE-1M.

\noindent\textbf{Test Dataset Difficulty.} The \textit{average test set difficulty} scores found in the last row of Table~\ref{tab:cross_dataset_generalization_deepfake_detection} reveal that GDD provides the most challenging test set for cross-dataset generalization experiments compared to existing remote sensing deepfake detection datasets.

\begin{table}[h]
\centering
\resizebox{\columnwidth}{!}{%
\begin{tabular}{lcccc|cc}
\toprule
\textbf{Method} & \textbf{FSI} & \textbf{DM-AER} & \textbf{RSFAKE-1M} & \textbf{GDD} & \textbf{Avg.} & \textbf{All Model Avg.} \\
\midrule
\rowcolor{gray!30}
\multicolumn{7}{l}{\textbf{Trained on FSI}} \\
UnivFD & \cellcolor{excellent}0.99 & \cellcolor{poor}0.61 & \cellcolor{verypoor}0.55 & \cellcolor{verypoor}0.45 & \cellcolor{verypoor}\textbf{0.54} & \cellcolor{poor} \\
AIDE & \cellcolor{excellent}1.00 & \cellcolor{poor}0.61 & \cellcolor{verypoor}0.41 & \cellcolor{verypoor}0.48 & \cellcolor{verypoor}\textbf{0.50} & \cellcolor{poor} \\
Co-Spy & \cellcolor{great}0.98 & \cellcolor{fair}0.71 & \cellcolor{poor}0.62 & \cellcolor{verypoor}0.54 & \cellcolor{poor}\textbf{0.62} & \cellcolor{poor} \\
DINOv3-sup & \cellcolor{excellent}1.00 & \cellcolor{fair}0.72 & \cellcolor{verypoor}0.57 & \cellcolor{verypoor}0.57 & \cellcolor{poor}\textbf{0.62} & \cellcolor{poor} \\
SimLBR & \cellcolor{excellent}0.99 & \cellcolor{good}0.82 & \cellcolor{poor}0.64 & \cellcolor{poor}0.64 & \cellcolor{fair}\textbf{0.70} & \cellcolor{poor}\multirow{-5}{*}{\textbf{0.60}} \\
\midrule
\rowcolor{gray!30}
\multicolumn{7}{l}{\textbf{Trained on DM-AER}} \\
UnivFD & \cellcolor{poor}0.63 & \cellcolor{great}0.90 & \cellcolor{verypoor}0.58 & \cellcolor{poor}0.62 & \cellcolor{poor}\textbf{0.61} & \cellcolor{poor} \\
AIDE & \cellcolor{poor}0.63 & \cellcolor{excellent}1.00 & \cellcolor{verypoor}0.54 & \cellcolor{verypoor}0.54 & \cellcolor{verypoor}\textbf{0.57} & \cellcolor{poor} \\
Co-Spy & \cellcolor{verypoor}0.51 & \cellcolor{fair}0.70 & \cellcolor{verypoor}0.51 & \cellcolor{verypoor}0.57 & \cellcolor{verypoor}\textbf{0.53} & \cellcolor{poor} \\
DINOv3-sup & \cellcolor{good}0.83 & \cellcolor{excellent}0.99 & \cellcolor{poor}0.68 & \cellcolor{fair}0.71 & \cellcolor{fair}\textbf{0.74} & \cellcolor{poor} \\
SimLBR & \cellcolor{poor}0.64 & \cellcolor{great}0.93 & \cellcolor{fair}0.77 & \cellcolor{fair}0.77 & \cellcolor{fair}\textbf{0.72} & \cellcolor{poor}\multirow{-5}{*}{\textbf{0.64}} \\
\midrule
\rowcolor{gray!30}
\multicolumn{7}{l}{\textbf{Trained on RSFAKE-1M}} \\
UnivFD & \cellcolor{verypoor}0.50 & \cellcolor{verypoor}0.48 & \cellcolor{good}0.88 & \cellcolor{verypoor}0.47 & \cellcolor{verypoor}\textbf{0.48} & \cellcolor{poor} \\
AIDE & \cellcolor{verypoor}0.52 & \cellcolor{poor}0.60 & \cellcolor{great}0.98 & \cellcolor{poor}0.68 & \cellcolor{poor}\textbf{0.60} & \cellcolor{poor} \\
Co-Spy & \cellcolor{verypoor}0.52 & \cellcolor{verypoor}0.52 & \cellcolor{great}0.91 & \cellcolor{poor}0.66 & \cellcolor{verypoor}\textbf{0.57} & \cellcolor{poor} \\
DINOv3-sup & \cellcolor{fair}0.77 & \cellcolor{fair}0.79 & \cellcolor{great}0.97 & \cellcolor{poor}0.69 & \cellcolor{fair}\textbf{0.75} & \cellcolor{poor} \\
SimLBR & \cellcolor{good}0.81 & \cellcolor{fair}0.79 & \cellcolor{great}0.96 & \cellcolor{fair}0.72 & \cellcolor{fair}\textbf{0.77} & \cellcolor{poor}\multirow{-5}{*}{\textbf{0.63}} \\
\midrule
\rowcolor{gray!30}
\multicolumn{7}{l}{\textbf{Trained on GDD}} \\
UnivFD & \cellcolor{verypoor}0.59 & \cellcolor{verypoor}0.56 & \cellcolor{verypoor}0.47 & \cellcolor{poor}0.65 & \cellcolor{verypoor}\textbf{0.54} & \cellcolor{fair} \\
AIDE & \cellcolor{poor}0.67 & \cellcolor{poor}0.61 & \cellcolor{poor}0.61 & \cellcolor{excellent}0.99 & \cellcolor{poor}\textbf{0.63} & \cellcolor{fair} \\
Co-Spy & \cellcolor{verypoor}0.56 & \cellcolor{fair}0.76 & \cellcolor{fair}0.75 & \cellcolor{great}0.95 & \cellcolor{poor}\textbf{0.69} & \cellcolor{fair} \\
DINOv3-sup & \cellcolor{good}0.80 & \cellcolor{great}0.90 & \cellcolor{fair}0.73 & \cellcolor{great}0.99 & \cellcolor{good}\textbf{0.81} & \cellcolor{fair} \\
SimLBR & \cellcolor{good}0.85 & \cellcolor{great}0.91 & \cellcolor{good}0.83 & \cellcolor{great}0.94 & \cellcolor{good}\textbf{0.86} & \cellcolor{fair}\multirow{-5}{*}{\textbf{0.71}} \\
\midrule
\textbf{Avg. Test Set Difficulty} & \cellcolor{poor}\textbf{0.66} & \cellcolor{poor}\textbf{0.69} & \cellcolor{poor}\textbf{0.62} & \cellcolor{poor}\textbf{0.61} & \multicolumn{2}{l}{} \\
\bottomrule
\end{tabular}%
}
\caption{Cross-dataset generalization performance. Scores indicate balanced accuracy. Diagonal entries correspond to within-dataset evaluation (train and test on the same dataset) and are excluded from the cross-dataset averages.}
\label{tab:cross_dataset_generalization_deepfake_detection}
\end{table}


\subsection{Deepfake Detection on GDD Test Splits}\label{sec:gdd_detection}
In addition to in-domain test results, GDD's cross-generator and cross-domain test sets enable controlled evaluation of transferability across unseen real image sources and generators. Results reported in Table~\ref{tab:gdd_crossdomain_breakdown} present overall accuracy of deepfake detectors trained on the GDD training split and evaluated on the in-domain (ID), cross-generator (CG), and cross-domain (CD) test splits. With the exception of UnivFD, all detectors exhibit a reduction in overall accuracy when exposed to GDD's CG and CD test scenarios. Co-Spy and SimLBR experience a strong drop in accuracy on the CD test split, where real images from unseen sources are introduced. This is a result of the latent blending regularization scheme that enables learning a tight decision boundary around the distribution of real images seen during training. Our benchmark reveals that this can be a problematic approach in remote sensing where real images can significantly vary as a result of different or new sensors and image preprocessing procedures. In Supplementary~\ref{supp:deepfake-detection} we present additional experiments including an ablation on DINOv3 checkpoints, comparing with non-RS pretrained deepfake detection models, and visualizations of top 20 hardest and easiest to detect real and fake images present in GDD.

\begin{table}[t]
    \centering
    \small
    \setlength{\tabcolsep}{4pt}
    \renewcommand{\arraystretch}{1.2}
    \begin{tabular}{l ccc}
    \toprule
    \textbf{Method} & \textbf{ID} & \textbf{CG} & \textbf{CD} \\
    \midrule
    UnivFD 
        & \cellcolor{poor}0.65 \scriptsize{(0.55 / 0.75)} 
        & \cellcolor{poor}0.63 \scriptsize{(0.55 / 0.72)} 
        & \cellcolor{fair}0.70 \scriptsize{(0.68 / 0.72)} \\
    AIDE 
        & \cellcolor{excellent}0.99 \scriptsize{(0.99 / 0.99)} 
        & \cellcolor{good}0.89 \scriptsize{(0.98 / 0.79)} 
        & \cellcolor{good}0.81 \scriptsize{(0.82 / 0.79)} \\
    Co-SPY 
        & \cellcolor{great}0.95 \scriptsize{(0.90 / 1.00)} 
        & \cellcolor{good}0.86 \scriptsize{(0.89 / 0.83)} 
        & \cellcolor{poor}0.69 \scriptsize{(0.54 / 0.83)} \\
    DINOv3-sup 
        & \cellcolor{excellent}0.99 \scriptsize{(0.99 / 0.99)} 
        & \cellcolor{good}0.85 \scriptsize{(0.97 / 0.72)} 
        & \cellcolor{good}0.82 \scriptsize{(0.93 / 0.72)} \\
    SimLBR 
        & \cellcolor{great}0.94 \scriptsize{(0.93 / 0.95)} 
        & \cellcolor{good}0.85 \scriptsize{(0.92 / 0.78)} 
        & \cellcolor{fair}0.79 \scriptsize{(0.79 / 0.78)} \\
    \bottomrule
    \end{tabular}
    \caption{Performance on GDD test splits across In-Domain (\textbf{ID}), Cross-Generator (\textbf{CG}), and Cross-Domain (\textbf{CD}) evaluations. Scores indicate overall accuracy and class-wise \textit{Real / Fake} accuracy in parentheses.}
    \label{tab:gdd_crossdomain_breakdown}
\end{table}

\subsection{Manipulation Localization.}
In Table~\ref{tab:manipulation-localization-our-dataset}, we evaluate manipulation localization performance using pixel-level F1 scores (computed via best threshold per image and a fixed threshold of $0.5$). We benchmark pretrained checkpoints of TruFor~\cite{guillaro2023trufor} and MMFusion~\cite{triaridis2024exploring} on both the GDD In-Domain (ID) test split and the FLDCF test split. Additionally, we conduct cross-dataset generalization experiments by training TruFor on each dataset's training split and evaluate performance on the test split of another dataset.

The results show that GDD is a challenging and non-trivial manipulation localization dataset that overcomes saturation limitations and the limited diversity of manipulations present in the FLDCF dataset. When trained on FLDCF, TruFor achieves nearly perfect performance on its own test set at $0.951$ F1 fixed. However, when evaluated on GDD-ID, performance drops to $0.266$ F1 fixed, highlighting that models trained on FLDCF overfit to limited manipulation patterns and fail to generalize. Conversely, training on GDD yields robust, non-trivial performance on GDD-ID of $0.795$ F1 while avoiding the ceiling effects observed in FLDCF. 

\begin{figure}
    \centering
    \includegraphics[width=\linewidth]{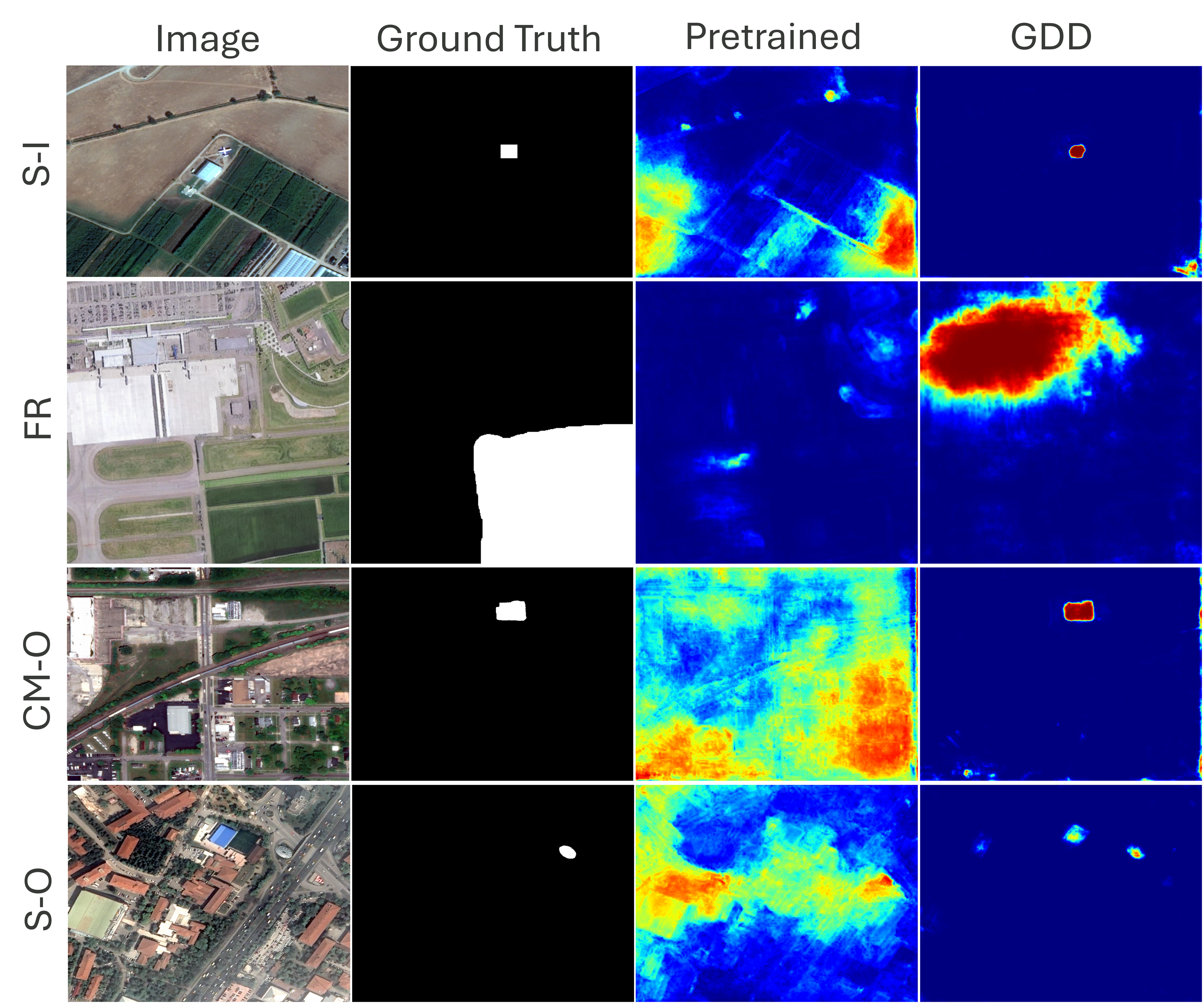}
    \caption{Localization maps predicted by TruFor on GDD in-domain test images, comparing the official pretrained weights with weights after training on GDD. Examples include splice-inpainting (S-I), fully-regenerated (FR), copy-move-object (CM-O), and splice-object (S-O) manipulations.}
    \label{fig:trufor-qual}
\end{figure}

As further detailed in Supplementary Section~\ref{supp:manip-results}, copy-move-object and fully-regenerated manipulations remain the most challenging variants to localize for both models, even after domain-specific training on GDD. Figure~\ref{fig:trufor-qual} provides qualitative localization maps on GDD-ID test samples, illustrating how GDD fine-tuning corrects false positives and sharpens boundary delineation across challenging manipulations including splice-inpainting (S-I), fully-regenerated (FR), copy-move-object (CM-O), and splice-object (S-O) manipulations. Even after training on GDD, some manipulations remain difficult to localize.

\begin{table}[t]
    \centering
    \resizebox{\columnwidth}{!}{
    \begin{tabular}{l cc cc}
    \toprule
    & \multicolumn{2}{c}{\textbf{FLDCF}}
    & \multicolumn{2}{c}{\textbf{GDD-ID}} \\
    \cmidrule(lr){2-3} \cmidrule(lr){4-5} 
    \textbf{Method} & Best & Fixed & Best & Fixed \\
    \midrule
    \rowcolor{gray!30}
    \textbf{Pretrained Checkpoint} & & & & \\
    TruFor \cite{guillaro2023trufor} & \cellcolor{verypoor}.557 & \cellcolor{verypoor}.486 & \cellcolor{fair}.703 & \cellcolor{poor}.620 \\
    MMFusion \cite{triaridis2024exploring} & \cellcolor{verypoor}.531 & \cellcolor{verypoor}.454 & \cellcolor{poor}.679 & \cellcolor{verypoor}.578 \\
    \rowcolor{gray!30}
    \textbf{Trained on FLDCF} & & & &  \\
    TruFor & \cellcolor{great}.959 & \cellcolor{great}.951 &  \cellcolor{verypoor}.350 &  \cellcolor{verypoor}.266 \\
    \rowcolor{gray!30}
    \textbf{Trained on GDD} & & & & \\
    TruFor & \cellcolor{verypoor}.487 & \cellcolor{verypoor}.411 & \cellcolor{good}.837 & \cellcolor{fair}.795 \\
    \bottomrule
    \end{tabular}
    }
    \caption{Pixel-level F1 manipulation localization evaluation across FLDCF dataset and GDD In-Domain (ID) test split.}
    \label{tab:manipulation-localization-our-dataset}
\end{table}

\section{Conclusion}
\label{sec:conclusion}

In this paper, we introduce GDD, a dataset for training and evaluating deepfake detection and manipulation localization models in the RS domain. GDD consists of images derived from a diverse selection of real RS image datasets, generative models, and manipulation procedures. Unlike existing RS deepfake benchmarks, which often prioritize dataset size at the expense of visual realism, diversity, and evaluations, GDD incorporates rigorous OOD testing protocols to assess model robustness under realistic distribution shifts. We demonstrate its utility through evaluations across diverse test splits that stress-test deepfake detection models against shifts in real image sources and generative models. For these evaluations, we utilize several state-of-the-art models for deepfake detection and manipulation localization. Our results highlight GDD's ability to facilitate the development and evaluation of generalizable models while exposing challenging test cases that motivate further research in RS deepfake detection and manipulation localization. Despite these contributions, GDD has some limitations. First, inpainting manipulations are generated using only a single generative inpainting model. Second, the dataset currently lacks cross-domain and cross-manipulation test splits for manipulation localization. Future work will address these limitations by incorporating additional generative and inpainting models, enabling more comprehensive and diverse training and testing scenarios for manipulation localization. Overall, GDD provides a foundation for advancing research and development in RS deepfake detection and manipulation localization.

\section*{LLM Usage Statement}
We used LLMs to assist with manuscript preparation which included refining grammar, language, writing, structure, and LaTeX formatting. We also used LLMs to assist with code implementation and debugging. The authors reviewed and verified all AI-assisted content and retain full responsibility for the manuscript's scientific claims, experimental results, and conclusions.

\section*{Acknowledgments}
We acknowledge that this manuscript has been authored by UT-Battelle, LLC under Contract No. DE-AC05-00OR22725 with the U.S. Department of Energy. The United States Government retains and the publisher, by accepting the article for publication, acknowledges that the United States Government retains a non-exclusive, paid-up, irrevocable, world-wide license to publish or reproduce the published form of this manuscript, or allow others to do so, for United States Government purposes. DOE will provide public access to these results of federally sponsored research in accordance with the DOE Public Access Plan (http://energy.gov/downloads/doe-public-access-plan). 

Research sponsored by the Laboratory Directed Research and Development Program of Oak Ridge National Laboratory, managed by UT-Battelle, LLC, for the U. S. Department of Energy. 

{
    \small
    \bibliographystyle{ieeenat_fullname}
    \bibliography{main}
}

\clearpage
\setcounter{page}{1}
\maketitlesupplementary


\begin{table*}[ht!]
    \centering
    \scriptsize
    \begin{tabular}{llccccc}
        \toprule
        \textbf{Type} & \textbf{Source / Generator / Manipulation} & \textbf{Train} & \textbf{Validation} & \textbf{In-Domain Test}  & \textbf{Cross-Generator Test} & \textbf{Cross-Domain Test} \\
        \midrule
        \multirow{6}{*}{Real}
            & fMoW & 1668 & 250 & 250 & 334 & -- \\
            & Million-AID & 1666 & 250 & 250 & 333 & -- \\
            & xBD & 1666 & 250 & 250 & 333 & -- \\
            & OrbitalNet & -- & -- & -- & -- & 333 \\
            & DOTA & -- & -- & -- & -- & 333 \\
            & INRIA & -- & -- & -- & -- & 334 \\
        \midrule
        \multirow{8}{*}{Fully Synthesized}
            & DiffusionSat & 1000 & 150 & 150 & -- & -- \\
            & R3GAN & 1000 & 150 & 150 & -- & -- \\
            & Flux2-dev & 1000 & 150 & 150 & -- & -- \\
            & Flux2-klein-base-4b & 1000 & 150 & 150 & -- & -- \\
            & NeDS & 1000 & 150 & 150 & -- & -- \\
            & RSPaint & -- & -- & -- & 333 & 333 \\
            & SD 3.5 Medium & -- & -- & -- & 333 & 333 \\
            & GeoSynth & -- & -- & -- & 334 & 334 \\
        \midrule
        \multirow{15}{*}{Manipulation}
            & Splice Simple (Million-AID to Million-AID) & 63 & 21 & 21 & -- & -- \\
            & Splice Simple (Million-AID to fMoW) & 63 & 20 & 20 & -- & -- \\
            & Splice Simple (fMoW to Million-AID) & 62 & 21 & 21 & -- & -- \\
            & Splice Simple (fMoW to fMoW) & 62 & 21 & 21 & -- & -- \\
            & Copy-move Simple (Million-AID) & 125 & 41 & 41 & -- & -- \\
            & Copy-move Simple (fMoW) & 125 & 42 & 42 & -- & -- \\
            & Splice Object (Million-AID to Million-AID) & 40 & 21 & 21 & -- & -- \\
            & Splice Object (Million-AID to fMoW) & 40 & 21 & 20 & -- & -- \\
            & Splice Object (fMoW to Million-AID) & 40 & 21 & 21 & -- & -- \\
            & Splice Object (fMoW to fMoW) & 40 & 20 & 21 & -- & -- \\
            & Copy-move Object (Million-AID) & 125 & 42 & 42 & -- & -- \\
            & Copy-move Object (fMoW) & 125 & 41 & 41 & -- & -- \\
            & Splice Inpaint (RSPaint to fMoW) & 125 & 42 & 42 & -- & -- \\
            & Splice Inpaint (RSPaint to Million-AID) & 125 & 42 & 42 & -- & -- \\
            & Fully Regenerated (RSPaint) & 250 & 84 & 84 & -- & -- \\
        \midrule
        \textbf{Total} & -- & 11410 & 2000 & 2000 & 2000 & 2000 \\
        \bottomrule
    \end{tabular}
    \caption{Number of images present in GDD and their sources across train, validation, in-domain test, cross-generator test, and cross-domain test splits.}
    \label{tab:dataset-split-details}
\end{table*}

\subsection{Dataset Composition}
\label{appendix:dataset-comp}
We present the counts of images present in each split along with their sources in Table~\ref{tab:dataset-split-details}. In Table~\ref{tab:real-dataset-info}, we describe the real source image datasets that are included in GDD. These datasets offer diversity in sensor, senor type, spatial resolution, and image format.

\begin{table*}[h]
    \centering
    \begin{tabular}{lcccc}
        \toprule
         \textbf{Dataset} & \textbf{Source/Sensors} & \textbf{Sensor Type} & \textbf{Spatial Resolution (m)} & \textbf{Image Format} \\
         \midrule
         fMoW & WV-02, WV-03, QB-02, GE-01 & Satellite & NA & GeoTIFF \\
         Million-AID & Google Earth & Satellite and Aerial &  0.5 to 153 & JPEG \\
         xBD & WV-02, WV-03, QB-02, GE-01 & Satellite & $~$0.5 & GeoTIFF \\
         DOTA & Google Earth, GF-2, JL-1 & Satellite and Aerial & NA & PNG \\
         INRIA & NA / (Aerial Orthophoto) & Aerial & 0.3 &  GeoTIFF \\ 
         OrbitalNet & WV-02, WV-03, QB-02, GE-01 & Satellite & 0.3 - 0.6 & GeoTIFF \\
         \bottomrule
    \end{tabular}
    \caption{Remote sensing datasets used for populating the ``real`` portion of GDD.}
    \label{tab:real-dataset-info}
\end{table*}






\subsection{Qualitative Examples of Samples in GDD}
\label{appendix:visual-examples}
\begin{figure*}[t]
    \centering
    \begin{subfigure}[b]{0.16\textwidth}
        \centering
        \includegraphics[width=\textwidth]{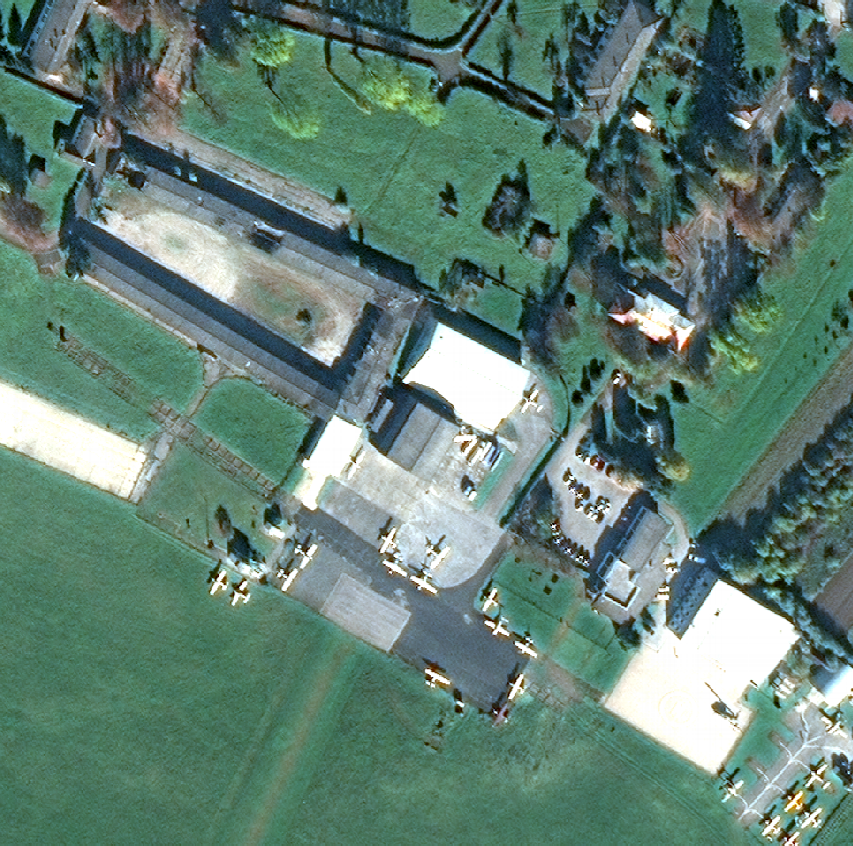}
        \caption{}
        \label{fig:1}
    \end{subfigure}
    \hfill
    \begin{subfigure}[b]{0.16\textwidth}
        \centering
        \includegraphics[width=\textwidth]{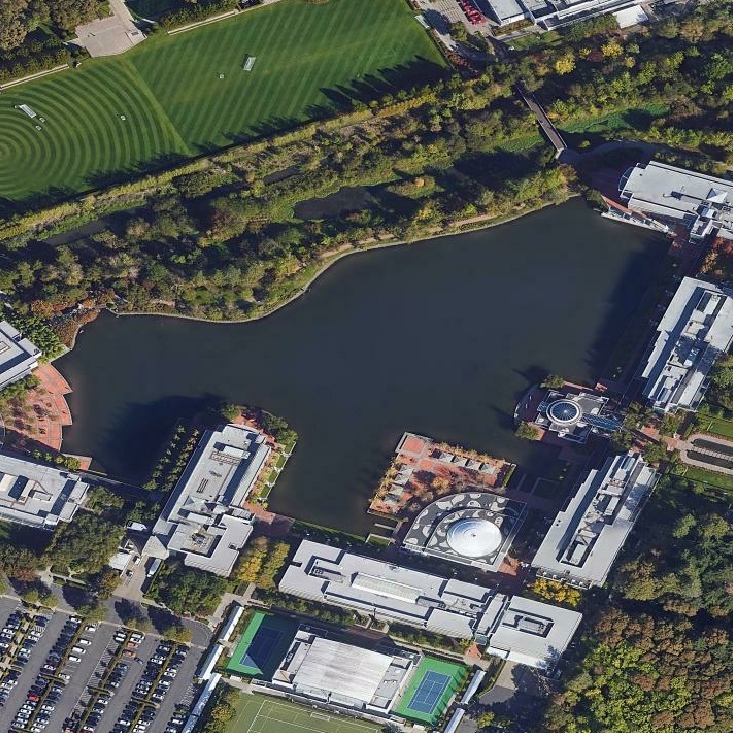}
        \caption{}
        \label{fig:2}
    \end{subfigure}
    \hfill
    \begin{subfigure}[b]{0.16\textwidth}
        \centering
        \includegraphics[width=\textwidth]{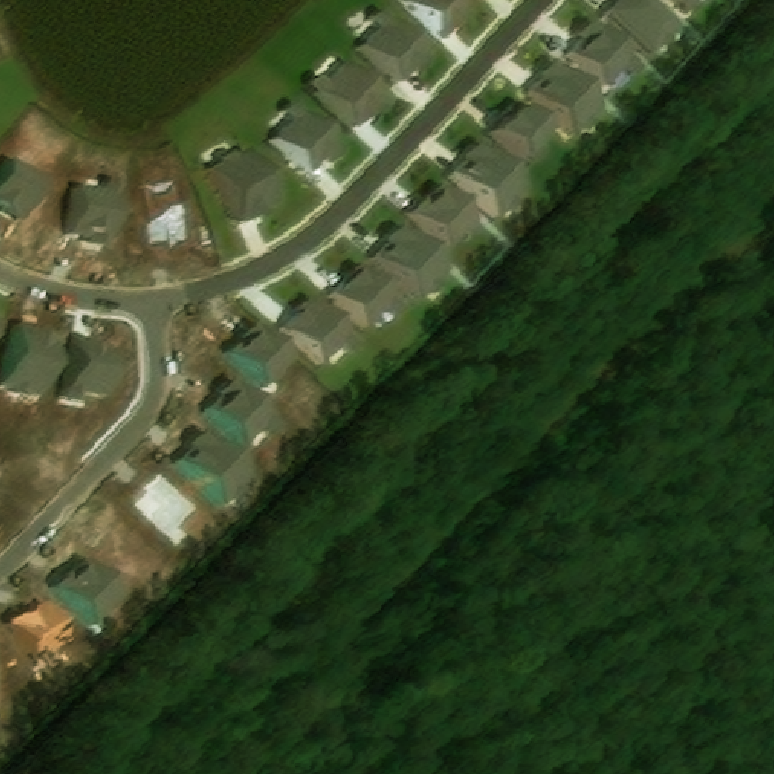}
        \caption{}
        \label{fig:3}
    \end{subfigure}
    \hfill
    \begin{subfigure}[b]{0.16\textwidth}
        \centering
        \includegraphics[width=\textwidth]{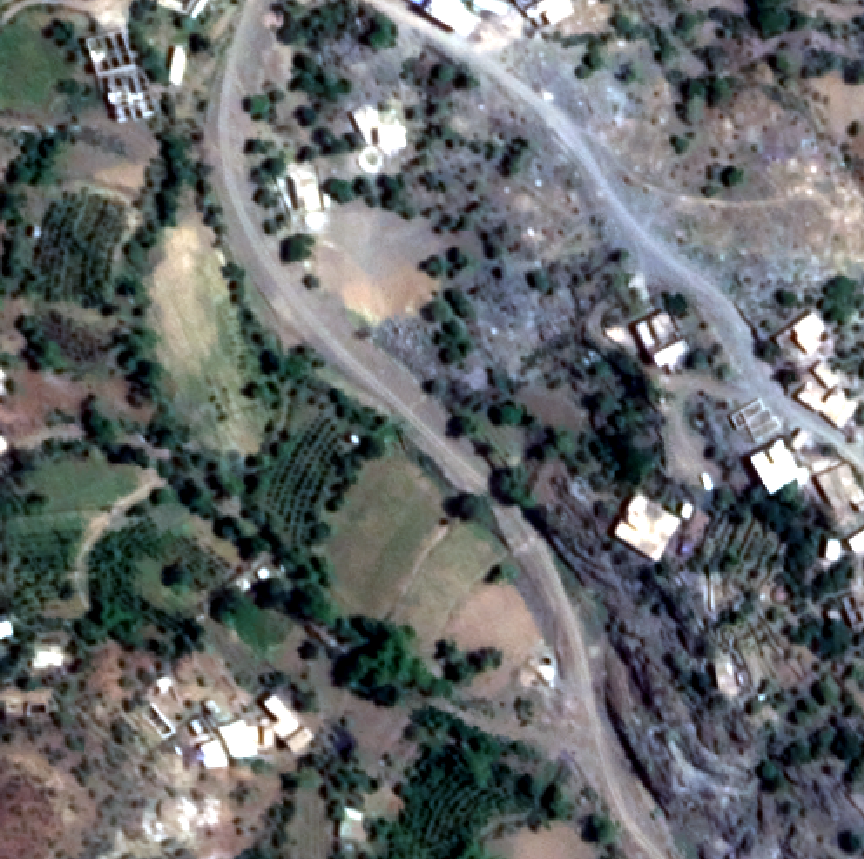}
        \caption{}
        \label{fig:4}
    \end{subfigure}
    \hfill
    \begin{subfigure}[b]{0.16\textwidth}
        \centering
        \includegraphics[width=\textwidth]{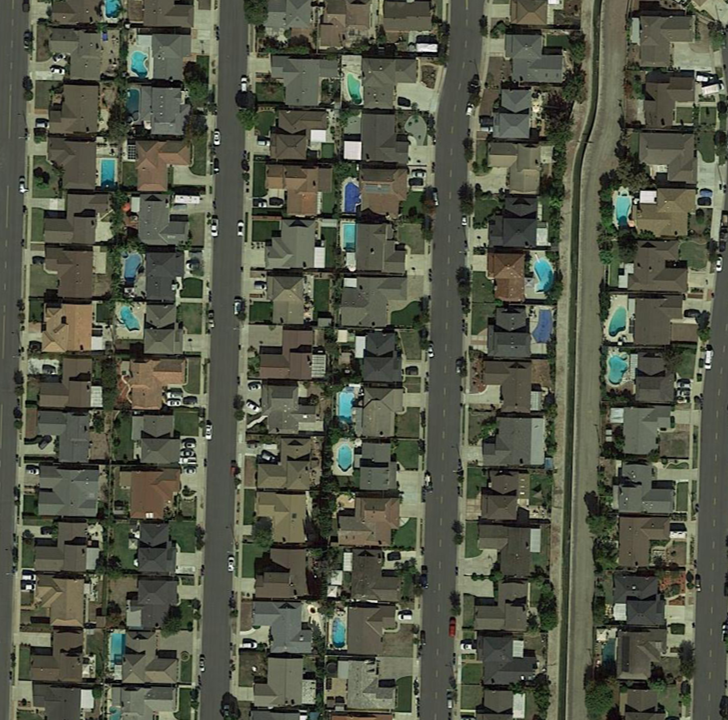}
        \caption{}
        \label{fig:5}
    \end{subfigure}
    \hfill
    \begin{subfigure}[b]{0.16\textwidth}
        \centering
        \includegraphics[width=\textwidth]{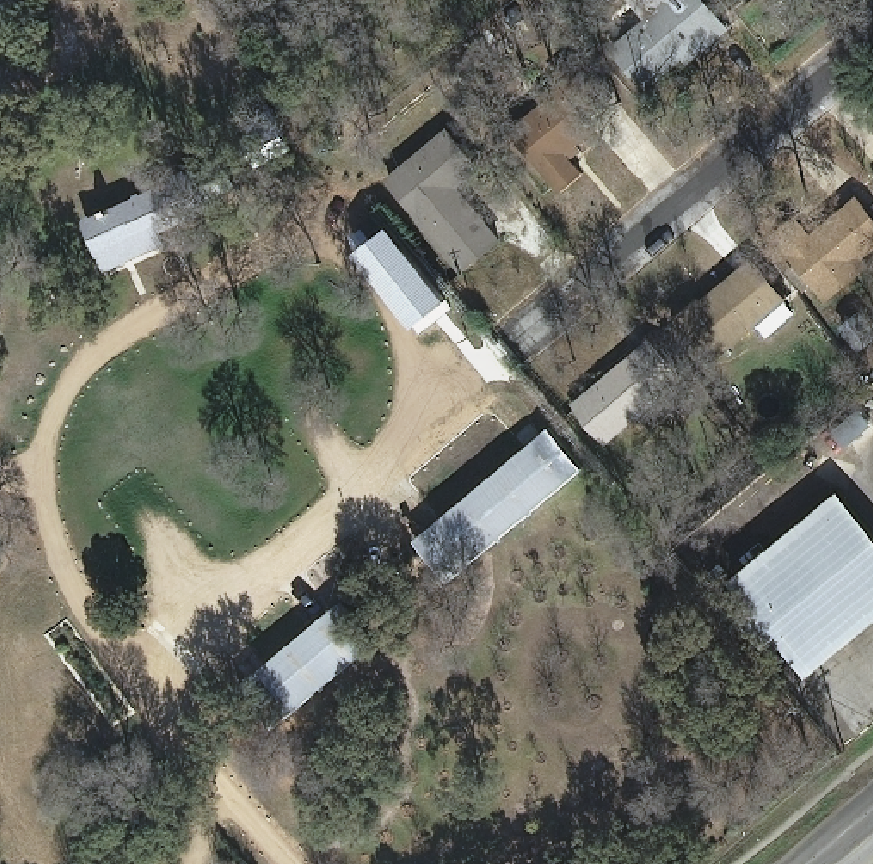}
        \caption{}
        \label{fig:6}
    \end{subfigure}
    \caption{Examples of real images included in GDD. The examples shown are drawn from the following source datasets, from left to right: fMoW, Million-AID, xBD, OrbitalNet, DOTA, INRIA.}
    \label{fig:qual-real-examples}
\end{figure*}

\begin{figure*}[t]
    \centering
    \begin{subfigure}[b]{0.12\textwidth}
        \centering
        \includegraphics[width=\textwidth]{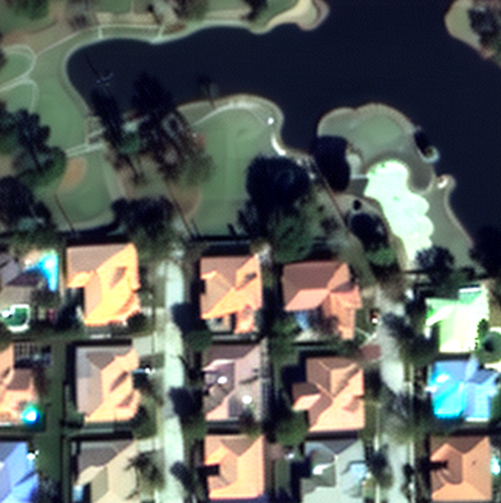}
        \caption{}
        \label{fig:1}
    \end{subfigure}
    \hfill
    \begin{subfigure}[b]{0.12\textwidth}
        \centering
        \includegraphics[width=\textwidth]{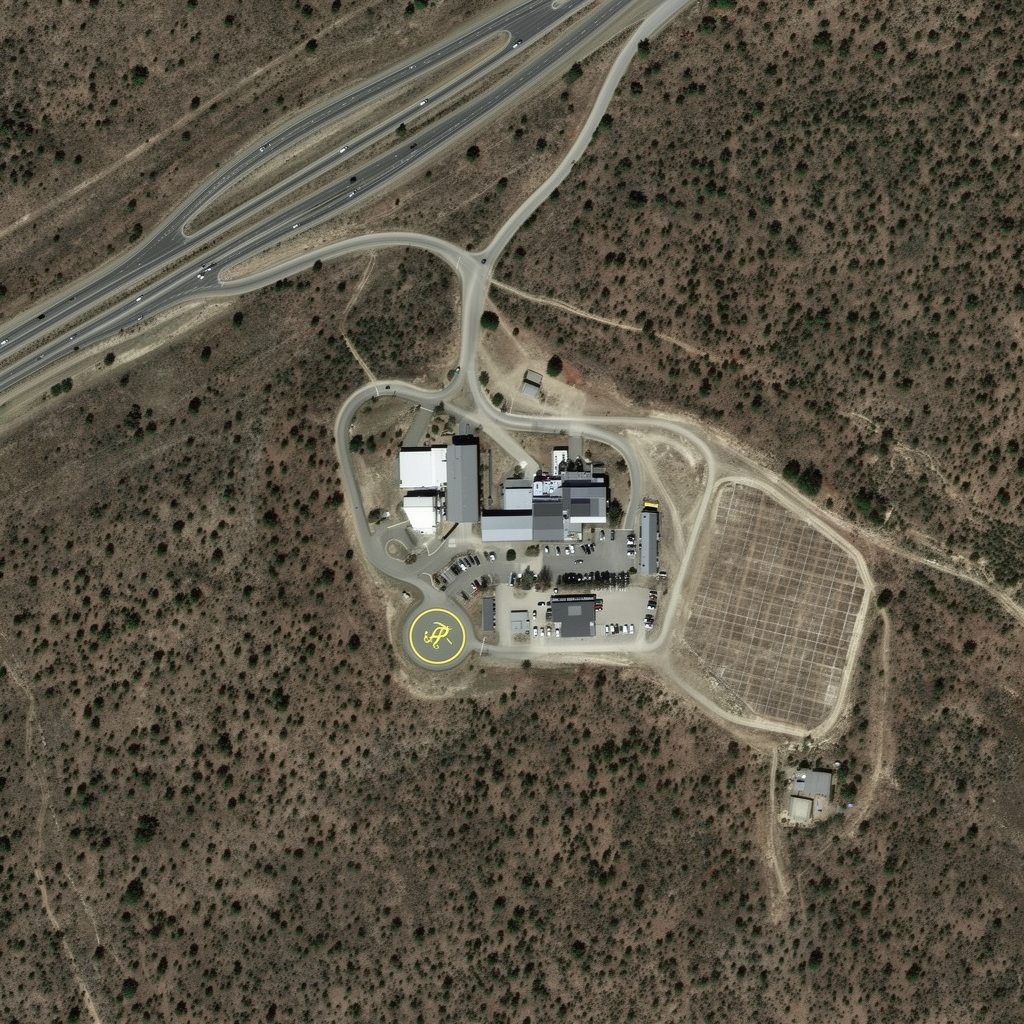}
        \caption{}
        \label{fig:2}
    \end{subfigure}
    \hfill
    \begin{subfigure}[b]{0.12\textwidth}
        \centering
        \includegraphics[width=\textwidth]{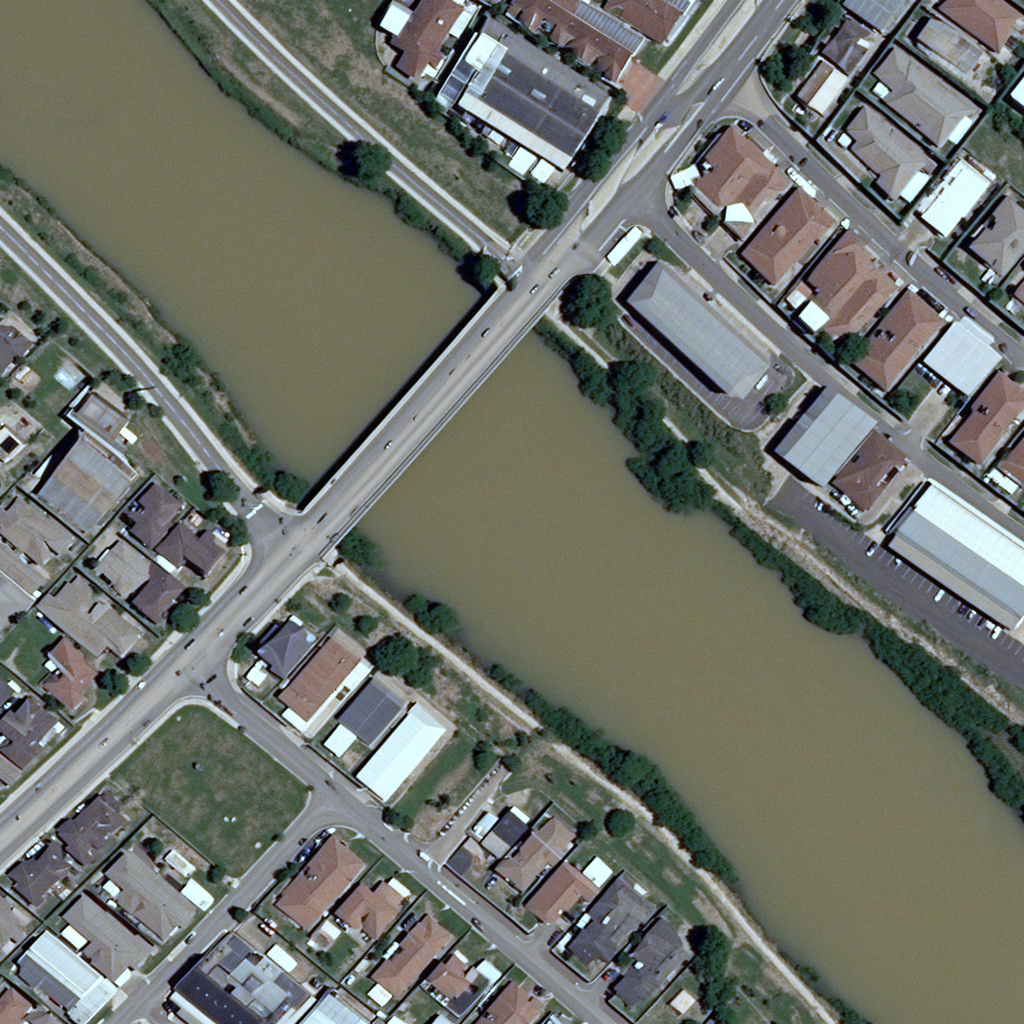}
        \caption{}
        \label{fig:3}
    \end{subfigure}
    \hfill
    \begin{subfigure}[b]{0.12\textwidth}
        \centering
        \includegraphics[width=\textwidth]{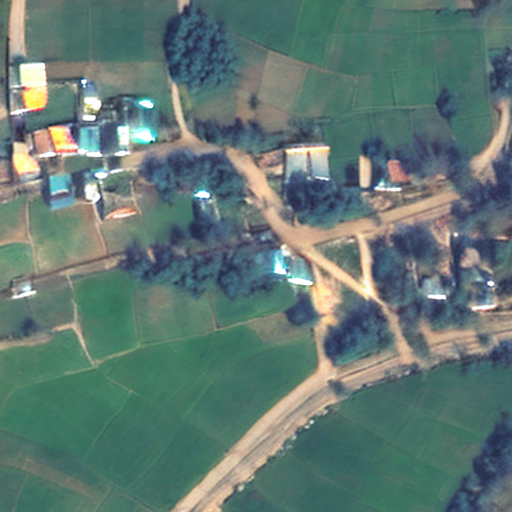}
        \caption{}
        \label{fig:4}
    \end{subfigure}
    \hfill
    \begin{subfigure}[b]{0.12\textwidth}
        \centering
        \includegraphics[width=\textwidth]{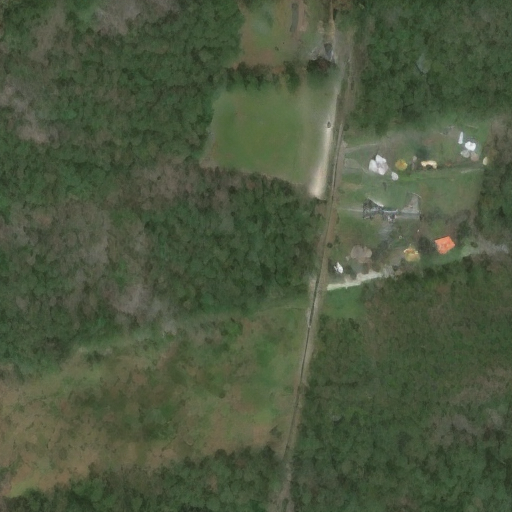}
        \caption{}
        \label{fig:5}
    \end{subfigure}
    \hfill
    \begin{subfigure}[b]{0.12\textwidth}
        \centering
        \includegraphics[width=\textwidth]{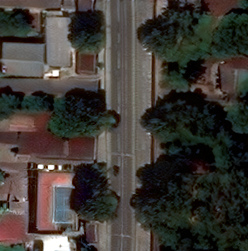}
        \caption{}
        \label{fig:6}
    \end{subfigure}
    \hfill
    \begin{subfigure}[b]{0.12\textwidth}
        \centering
        \includegraphics[width=\textwidth]{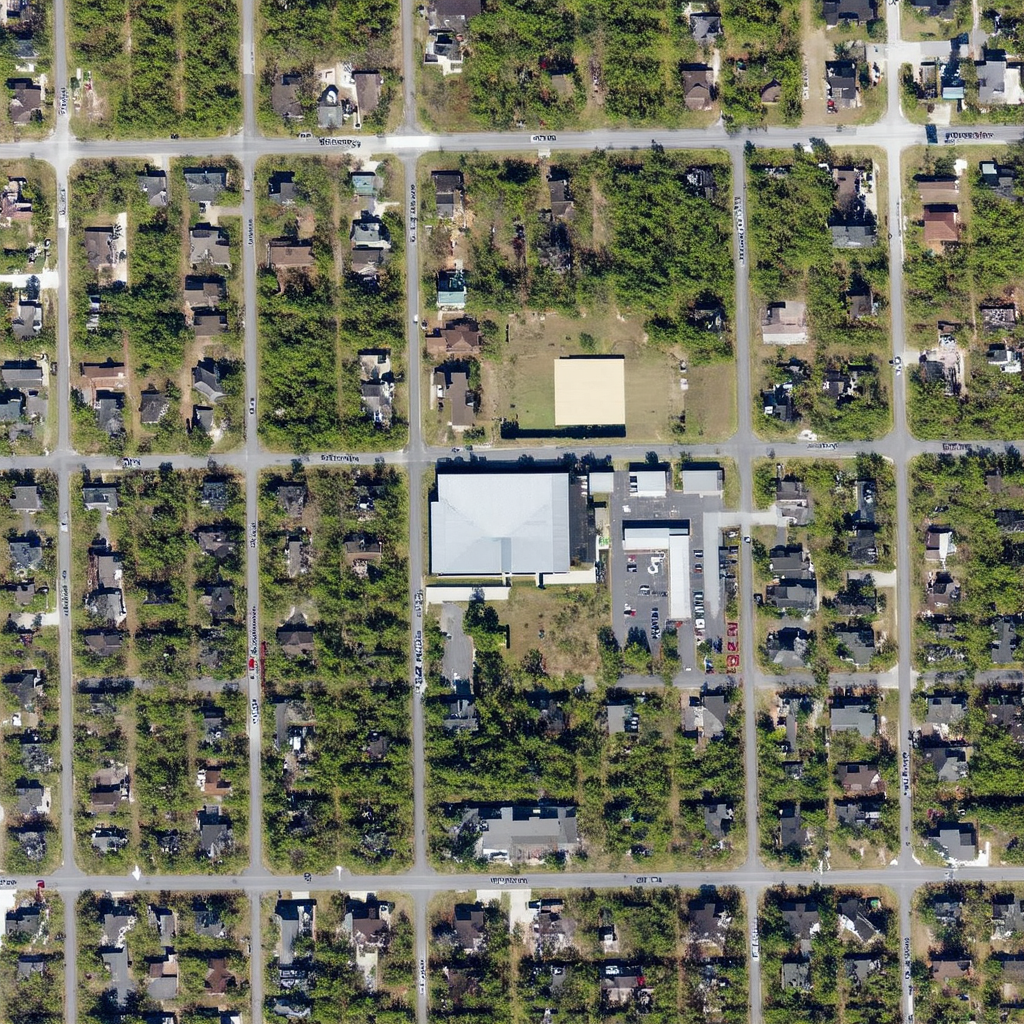}
        \caption{}
        \label{fig:7}
    \end{subfigure}
    \hfill
    \begin{subfigure}[b]{0.12\textwidth}
        \centering
        \includegraphics[width=\textwidth]{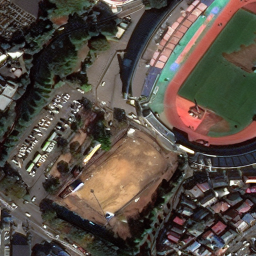}
        \caption{}
        \label{fig:8}
    \end{subfigure}
    \caption{Examples of fully synthesized images included in GDD. Images are produced from the following generators, from left to right: DiffusionSat, Flux2-dev, Flux2-dev-klein-base-4b, GeoSynth, NeDS, R3GAN, Stable Diffusion 3.5 Medium, and RSPaint.}
    \label{fig:qual-generated-examples}
\end{figure*}

\begin{figure*}[t]
    \centering
    \begin{subfigure}[b]{0.32\textwidth}
        \centering
        \includegraphics[width=\textwidth]{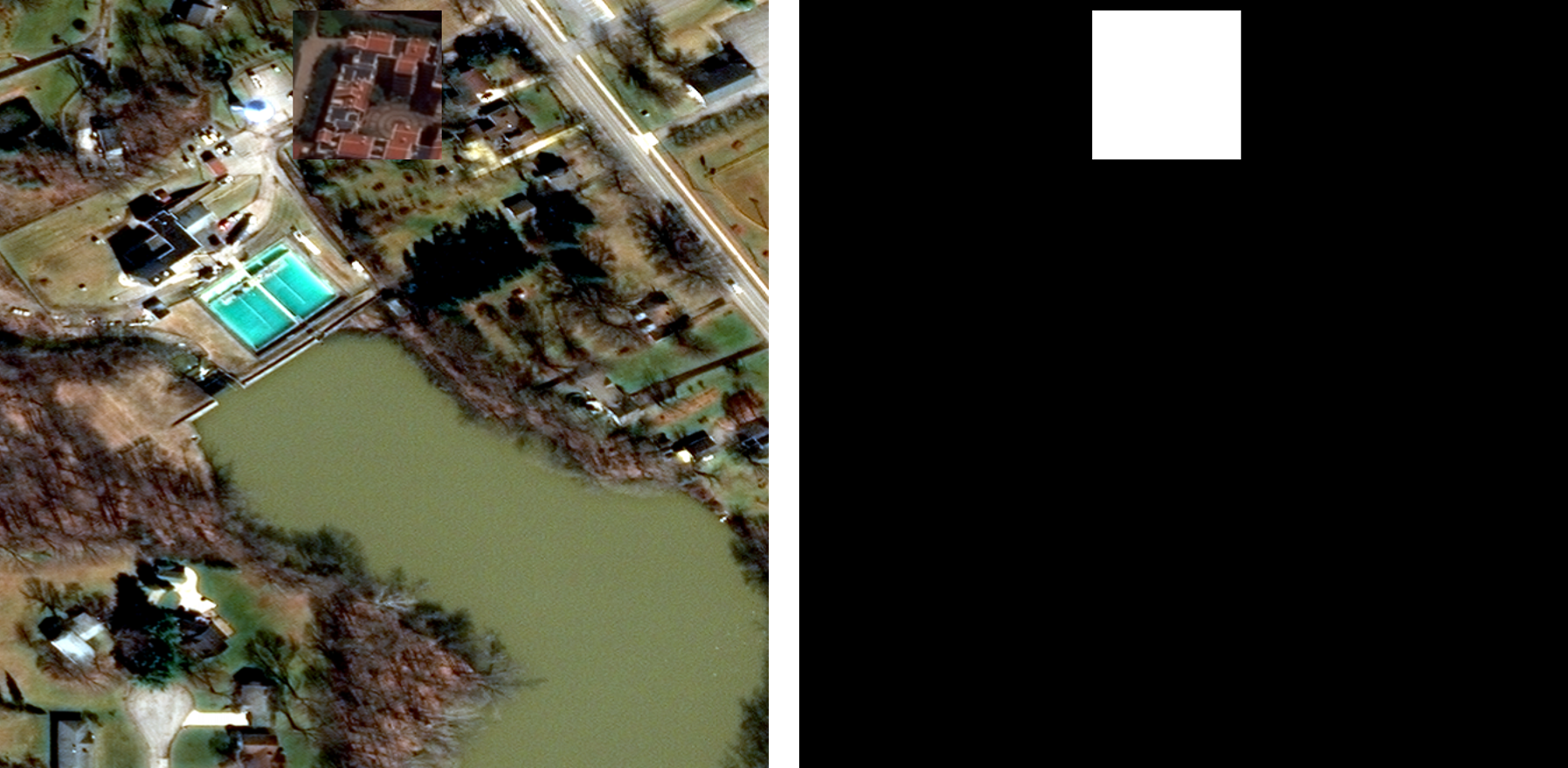}
        \caption{Splice Simple}
        \label{fig:1}
    \end{subfigure}
    \hfill
    \begin{subfigure}[b]{0.32\textwidth}
        \centering
        \includegraphics[width=\textwidth]{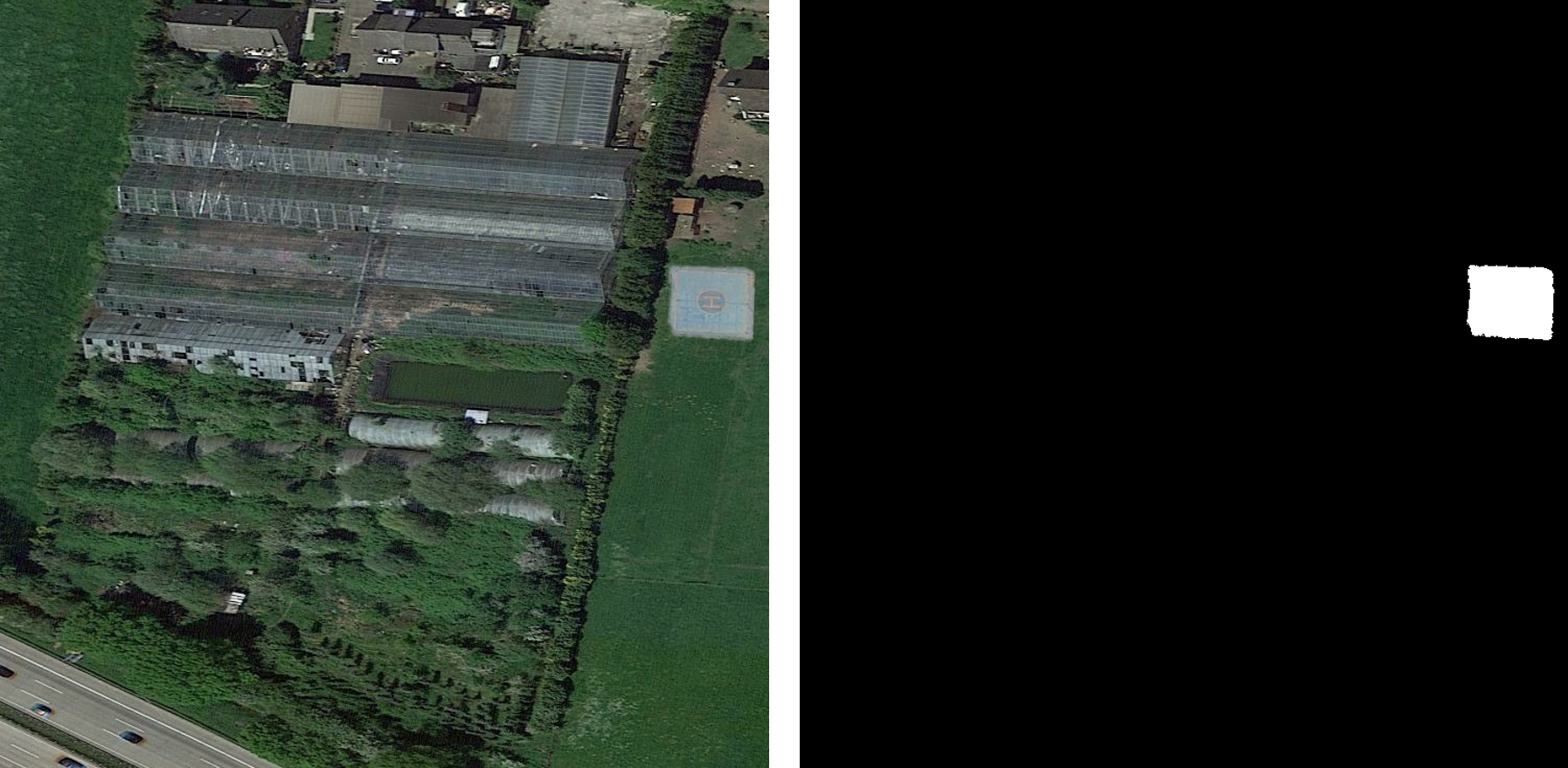}
        \caption{Splice Object}
        \label{fig:2}
    \end{subfigure}
    \hfill
    \begin{subfigure}[b]{0.32\textwidth}
        \centering
        \includegraphics[width=\textwidth]{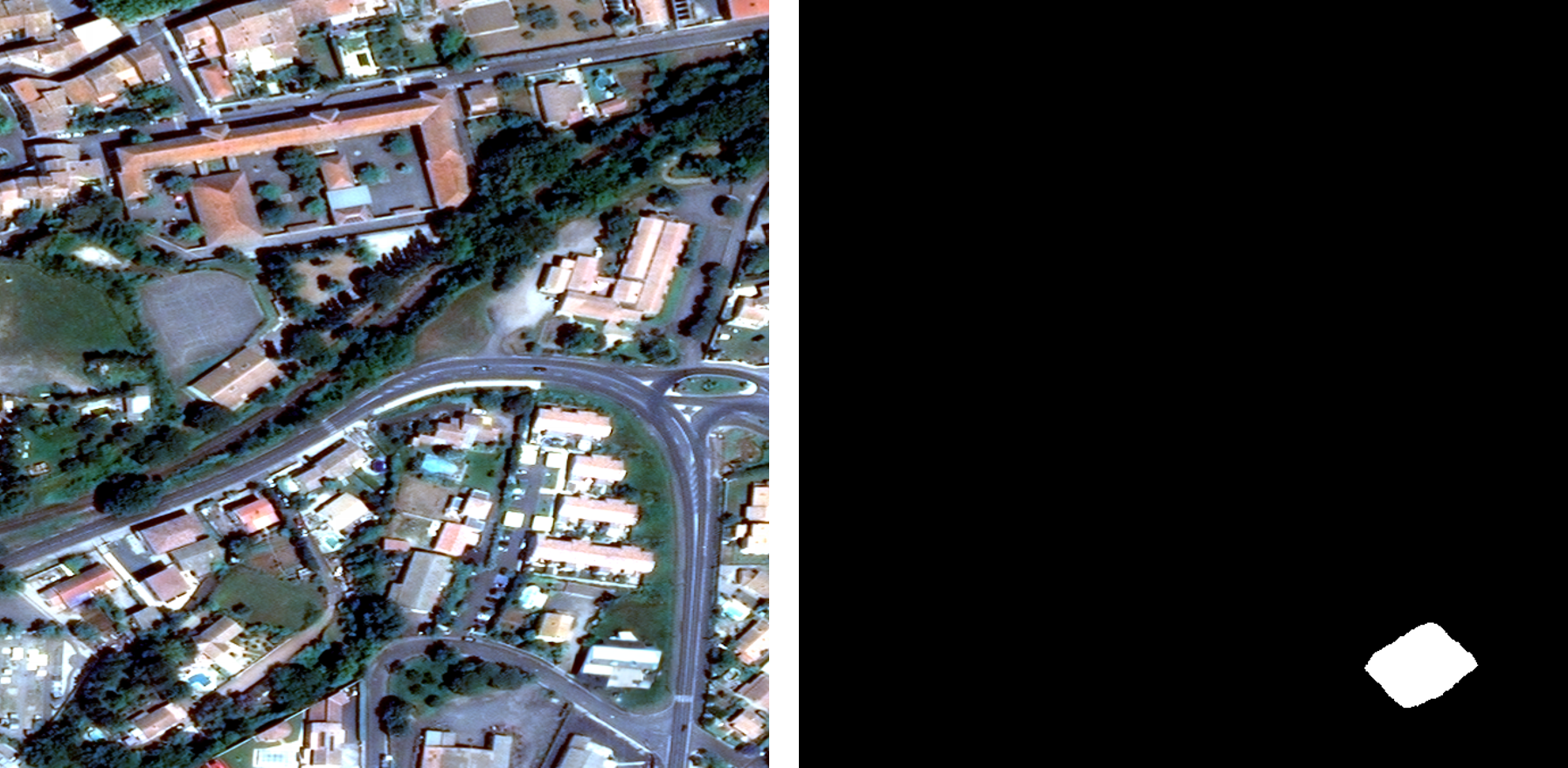}
        \caption{Splice Inpaint}
        \label{fig:3}
    \end{subfigure}

    \vspace{0.5em} 

    \begin{subfigure}[b]{0.32\textwidth}
        \centering
        \includegraphics[width=\textwidth]{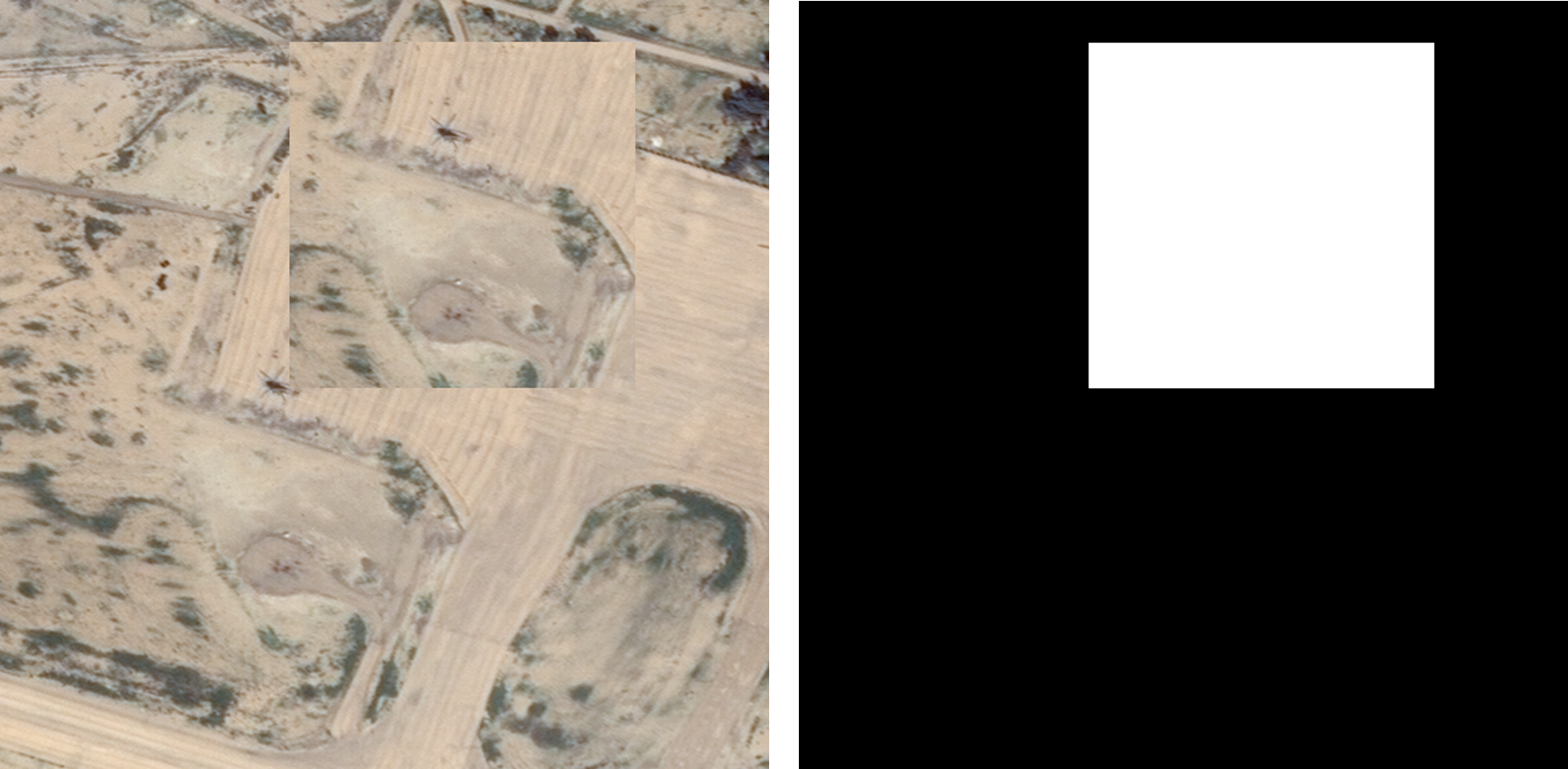}
        \caption{Copy-move Simple}
        \label{fig:4}
    \end{subfigure}
    \hfill
    \begin{subfigure}[b]{0.32\textwidth}
        \centering
        \includegraphics[width=\textwidth]{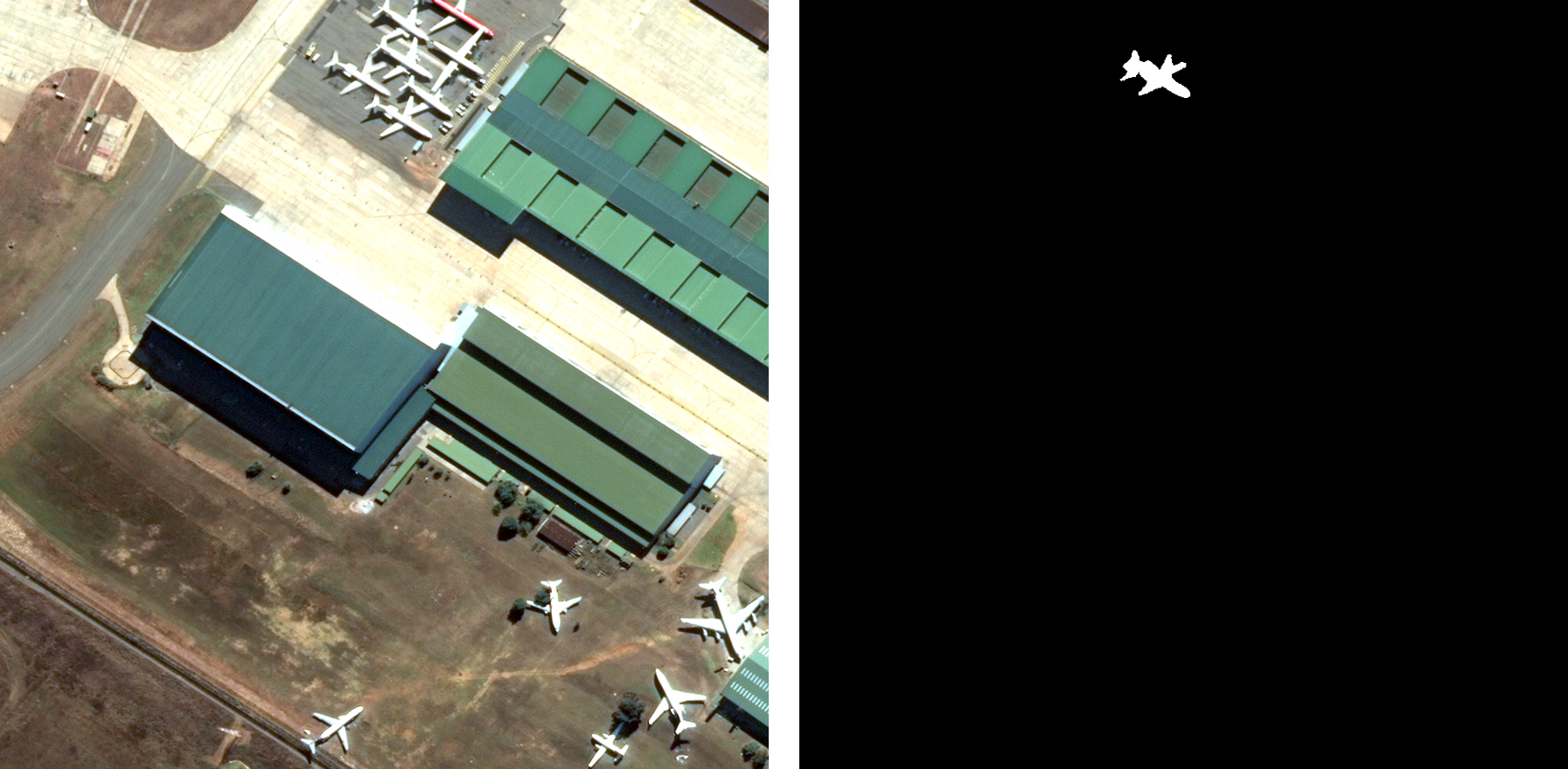}
        \caption{Copy-move Object}
        \label{fig:5}
    \end{subfigure}
    \hfill
    \begin{subfigure}[b]{0.32\textwidth}
        \centering
        \includegraphics[width=\textwidth]{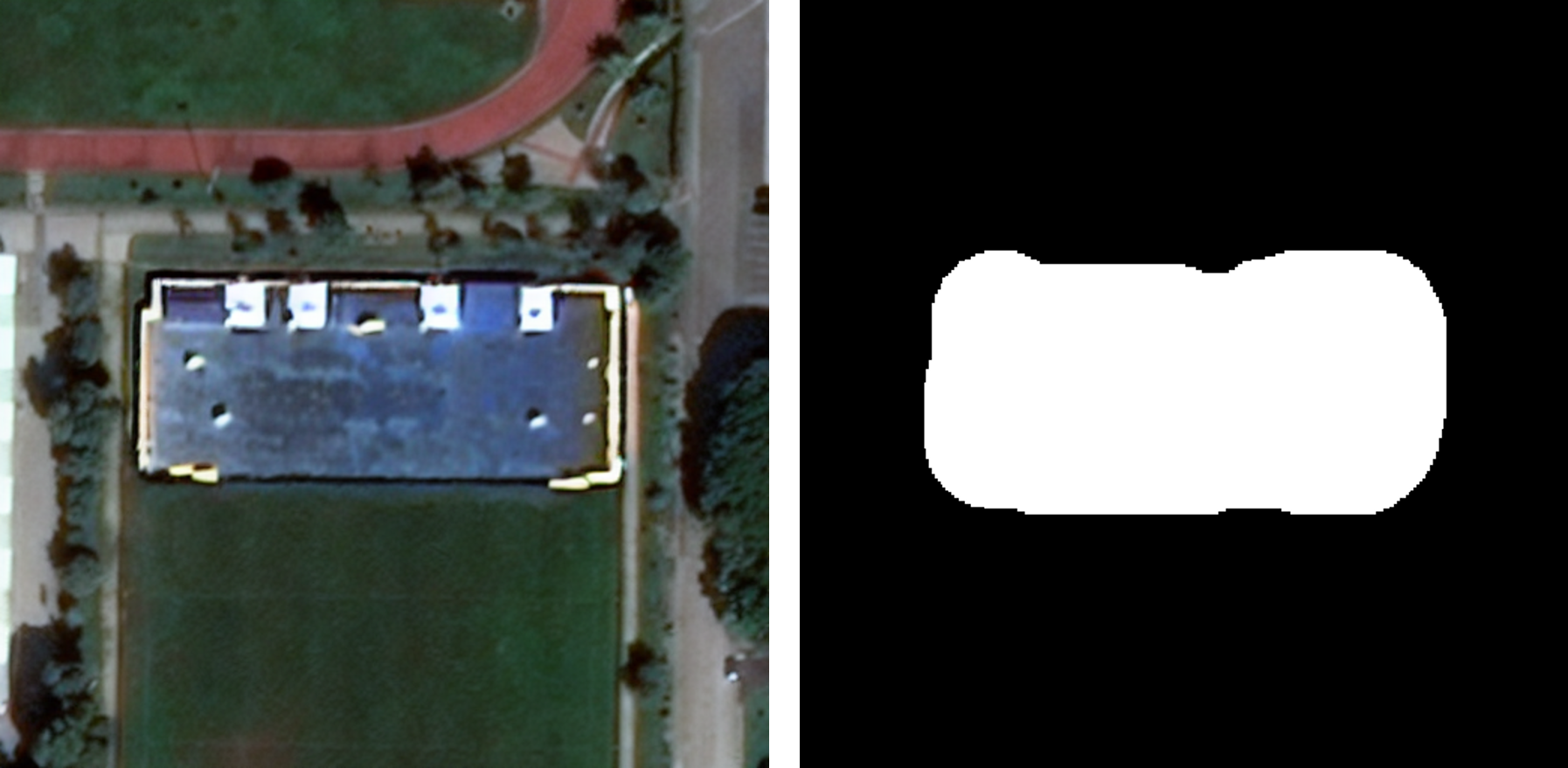}
        \caption{Fully Regenerated}
        \label{fig:6}
    \end{subfigure}
    \caption{Examples of manipulation types and variants included in GDD. We show the manipulated image and the corresponding ground truth mask that indicates which pixels are fake/manipulation in the image.}
    \label{fig:qual-manipulation-examples}
\end{figure*}

Figure~\ref{fig:qual-real-examples} presents examples of real images in GDD. Figure~\ref{fig:qual-generated-examples} displays examples of fully synthesized images present in the fake portion of GDD. Finally, Figure~\ref{fig:qual-manipulation-examples} presents examples of each of the six image manipulations that are included in the dataset.

\subsubsection{Text Prompt}
\label{appendix:text-prompt}
We use the following prompt and Qwen3-VL-8B-Instruct to auto generate captions for satellite images in the fMoW dataset. The automatically generated captions are used to prompt the text-to-image generative models for producing synthetic satellite images. \\

{\footnotesize\ttfamily
\noindent You are an expert satellite image analyst. Analyze this overhead/satellite image and generate a caption for training a text-to-image diffusion model.

\noindent\textbf{CRITICAL CONSTRAINTS:}\\
- Your caption MUST be 40--60 words maximum (approximately 50--70 tokens).\\
- DO NOT exceed 60 words under any circumstances.\\
- DO NOT use subjective language (beautiful, peaceful, serene, picturesque).\\
- DO NOT start with ``The image shows'' or ``This is''.\\
- The caption MUST convey that this is a high-altitude satellite image captured in nadir (top-down) view, NOT a drone or low-altitude aerial photograph.

\noindent\textbf{REQUIRED INFORMATION (in order of priority):}\\
1. Viewing perspective: establish that this is a satellite image with a top-down nadir view.\\
2. Primary land cover type (urban, suburban, rural, agricultural, forest, industrial, etc.).\\
3. Key objects present (buildings, roads, parking lots, vegetation, water bodies, specific facilities).\\
4. Spatial arrangement and density (dense, sparse, clustered, grid-like, scattered).\\
5. Spatial relationships between objects.\\
6. Notable positions.\\
\noindent Write as a single flowing paragraph. Be specific about object types and include
approximate counts or density where relevant.
}

\subsection{Dataset Analysis via t-SNE Visualization}
\label{appendix:tsne}
\begin{figure}[htbp]
    \centering
    \begin{subfigure}[t]{0.23\textwidth}
        \centering
        \includegraphics[width=\linewidth]{figures/tsnes/suite_tests_dinov3_tsne_a_real_by_source.png}
        \caption{Real images by dataset}
        \label{fig:tsne_real}
    \end{subfigure}
    \hfill
    \begin{subfigure}[t]{0.23\textwidth}
        \centering
        \includegraphics[width=\linewidth]{figures/tsnes/suite_tests_dinov3_tsne_b_fake_by_source.png}
        \caption{Fakes by dataset}
        \label{fig:tsne_fakes}
    \end{subfigure}
    \hfill
    \begin{subfigure}[t]{0.23\textwidth}
        \centering
        \includegraphics[width=\linewidth]{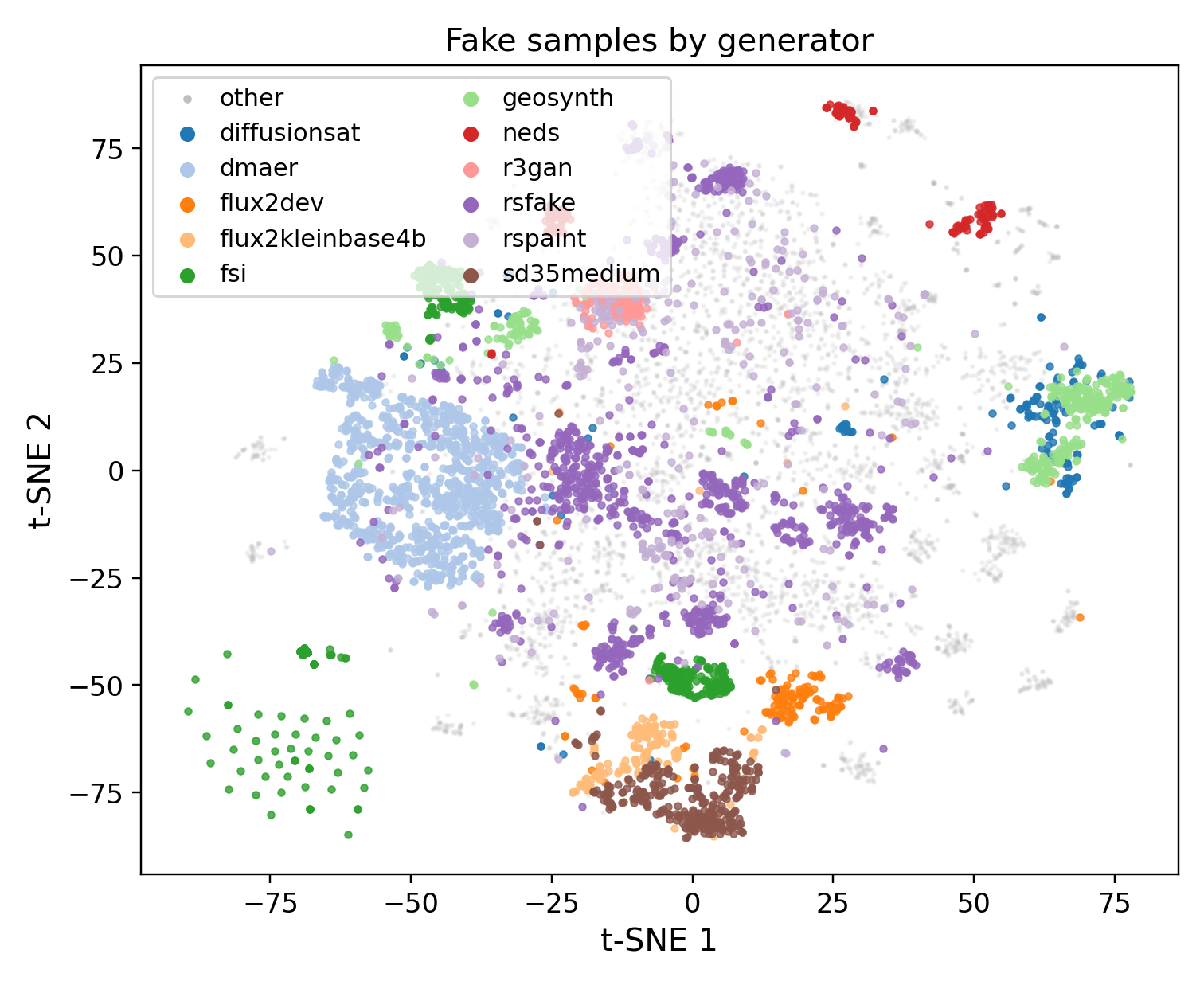}
        \caption{Fakes by generator}
        \label{fig:tsne_generator}
    \end{subfigure}
    \hfill
    \begin{subfigure}[t]{0.23\textwidth}
        \centering
        \includegraphics[width=\linewidth]{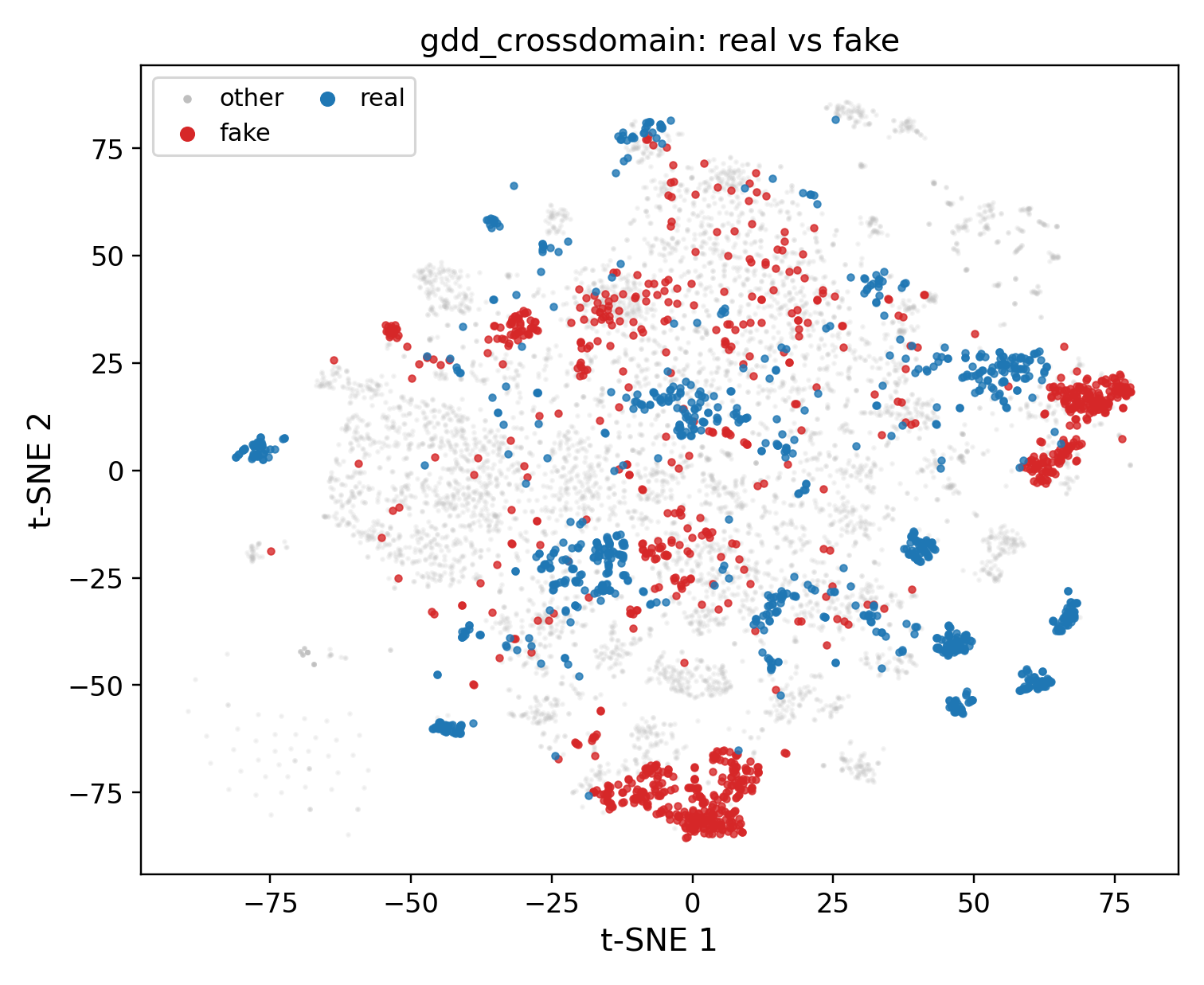}
        \caption{GDD (CD): Real/Fake}
        \label{fig:tsne_gddcross}
    \end{subfigure}
    
    \medskip 
    
    \begin{subfigure}[b]{0.23\textwidth}
        \centering
        \includegraphics[width=\linewidth]{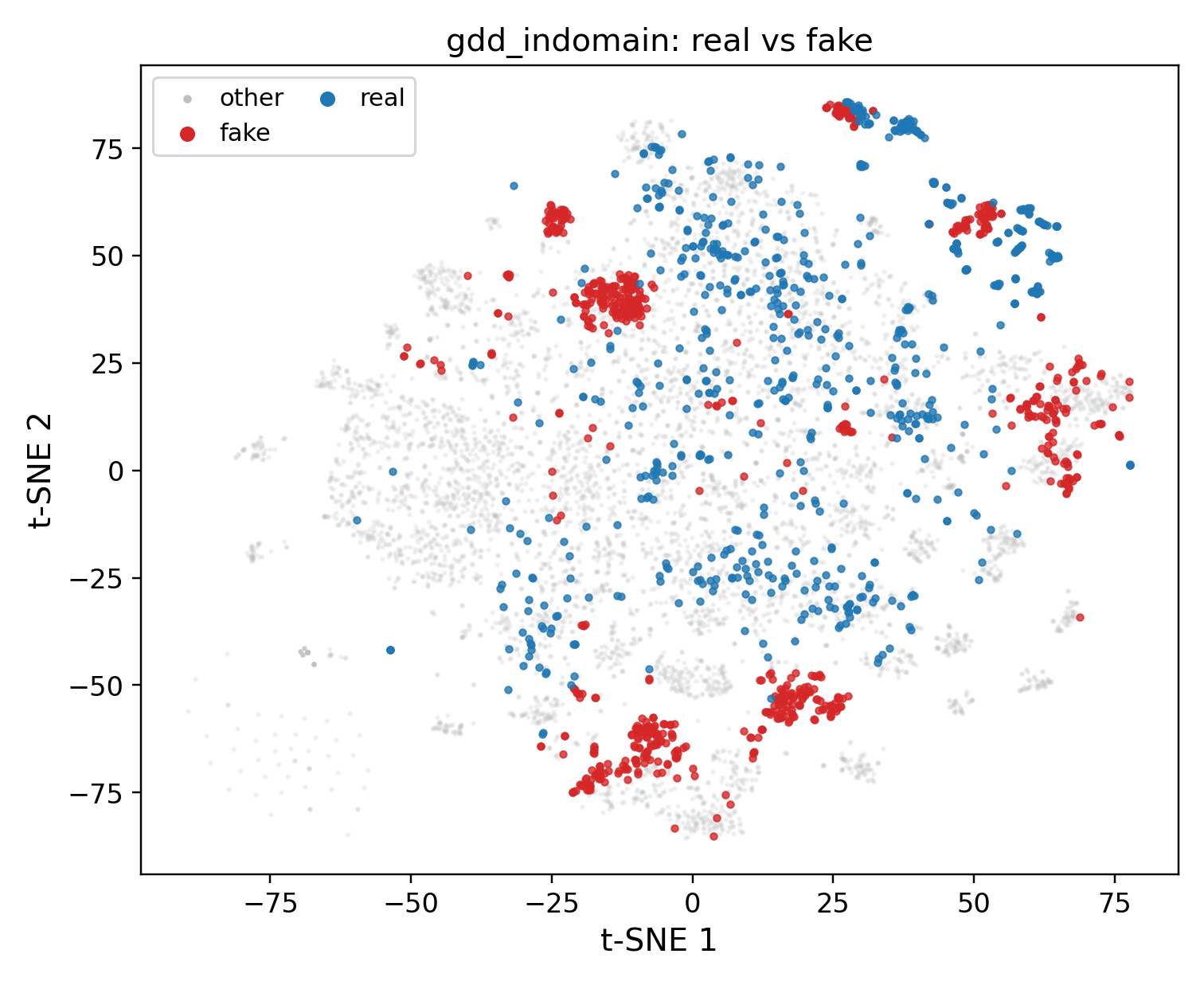}
        \caption{GDD (ID): Real/Fake}
        \label{fig:tsne_gddin}
    \end{subfigure}
    \hfill
    \begin{subfigure}[b]{0.23\textwidth}
        \centering
        \includegraphics[width=\linewidth]{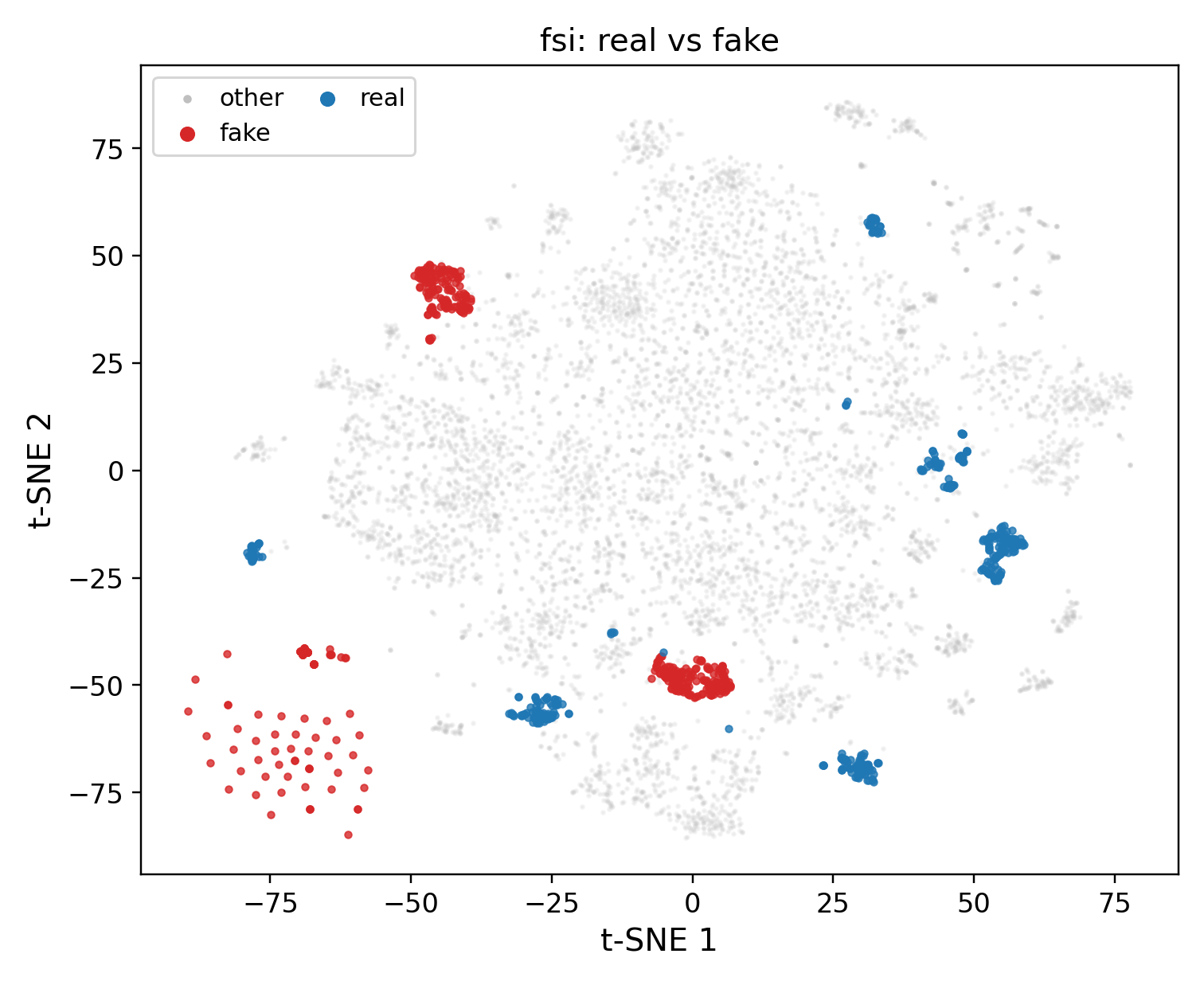}
        \caption{FSI: Real/Fake}
        \label{fig:tsne_fsi}
    \end{subfigure}
    \hfill
    \begin{subfigure}[b]{0.23\textwidth}
        \centering
        \includegraphics[width=\linewidth]{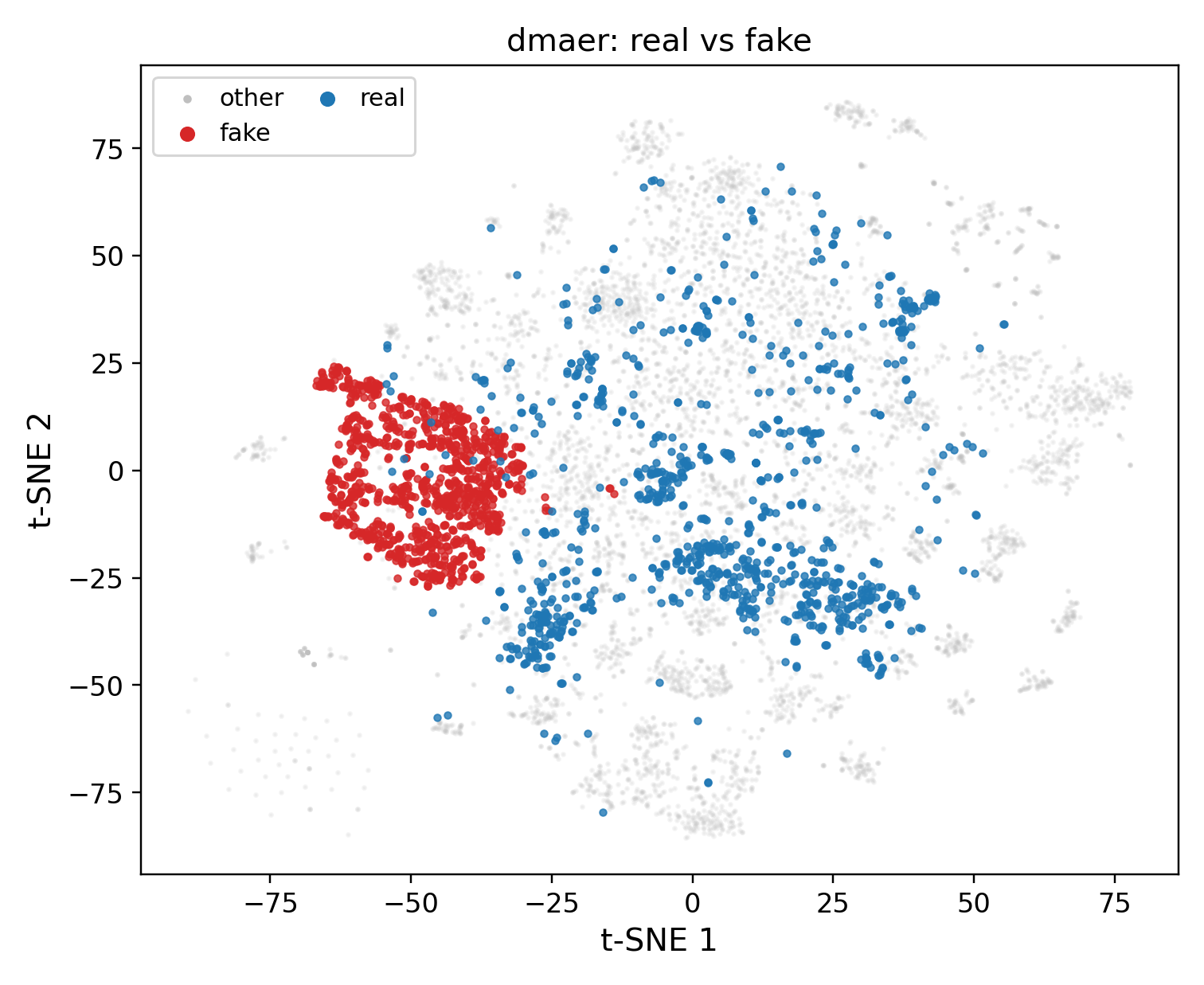}
        \caption{DM-AER: Real/Fake}
        \label{fig:tsne_dmaer}
    \end{subfigure}
    \hfill
    \begin{subfigure}[b]{0.23\textwidth}
        \centering
        \includegraphics[width=\linewidth]{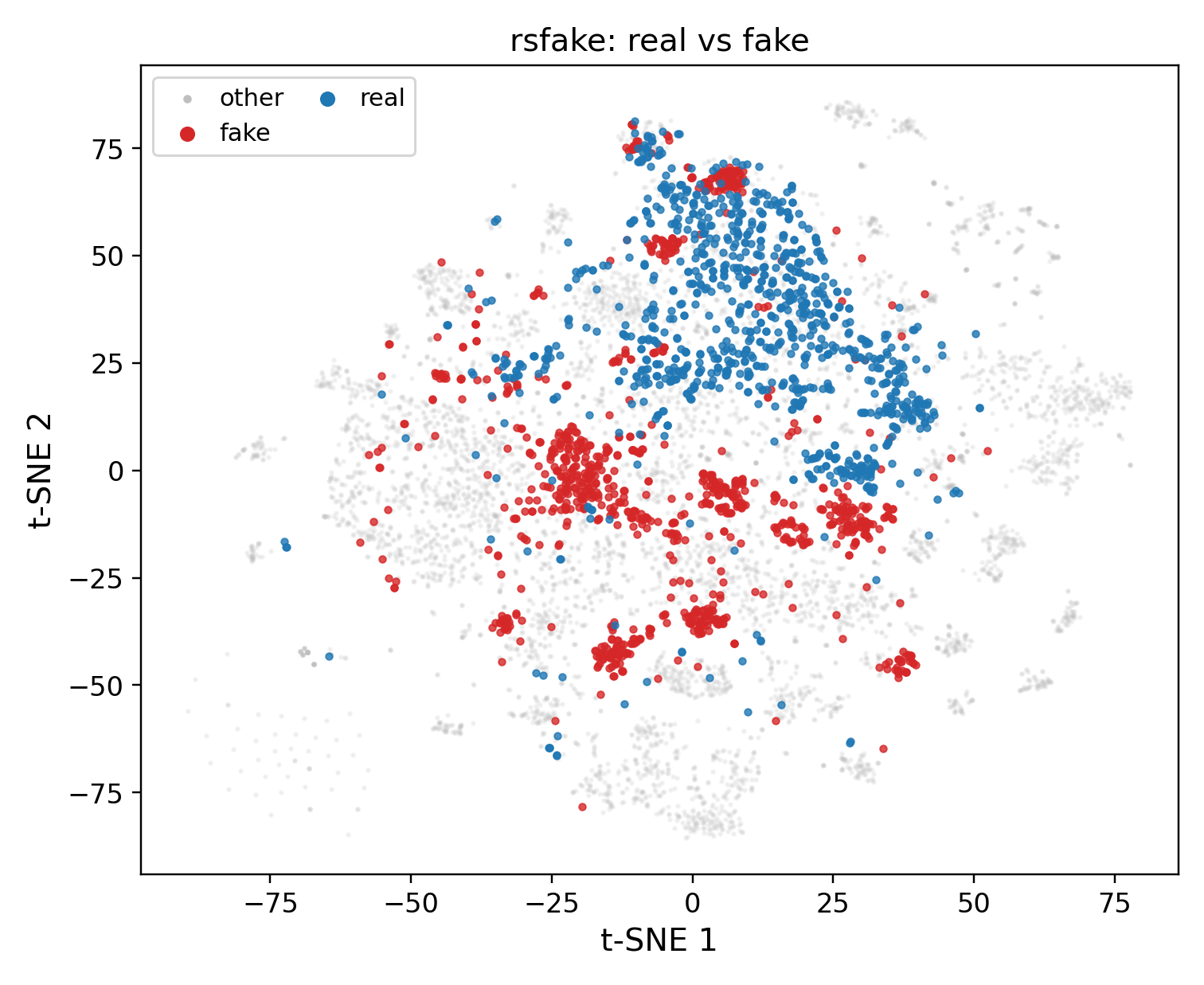}
        \caption{RSFAKE-1M: Real/Fake}
        \label{fig:tsne_rsfake}
    \end{subfigure}
    
    \caption{Illustration of t-SNE representations using DINOv3-Large-SAT for samples composing test sets of GDD (ID: In-Domain and CD: Cross-Domain), FSI, RSFAKE-1M, and DM-AER datasets}
    \label{fig:tsne}
\end{figure}

\begin{figure}[htbp]
    \centering
    \begin{subfigure}[t]{0.23\textwidth}
        \centering
        \includegraphics[width=\linewidth]{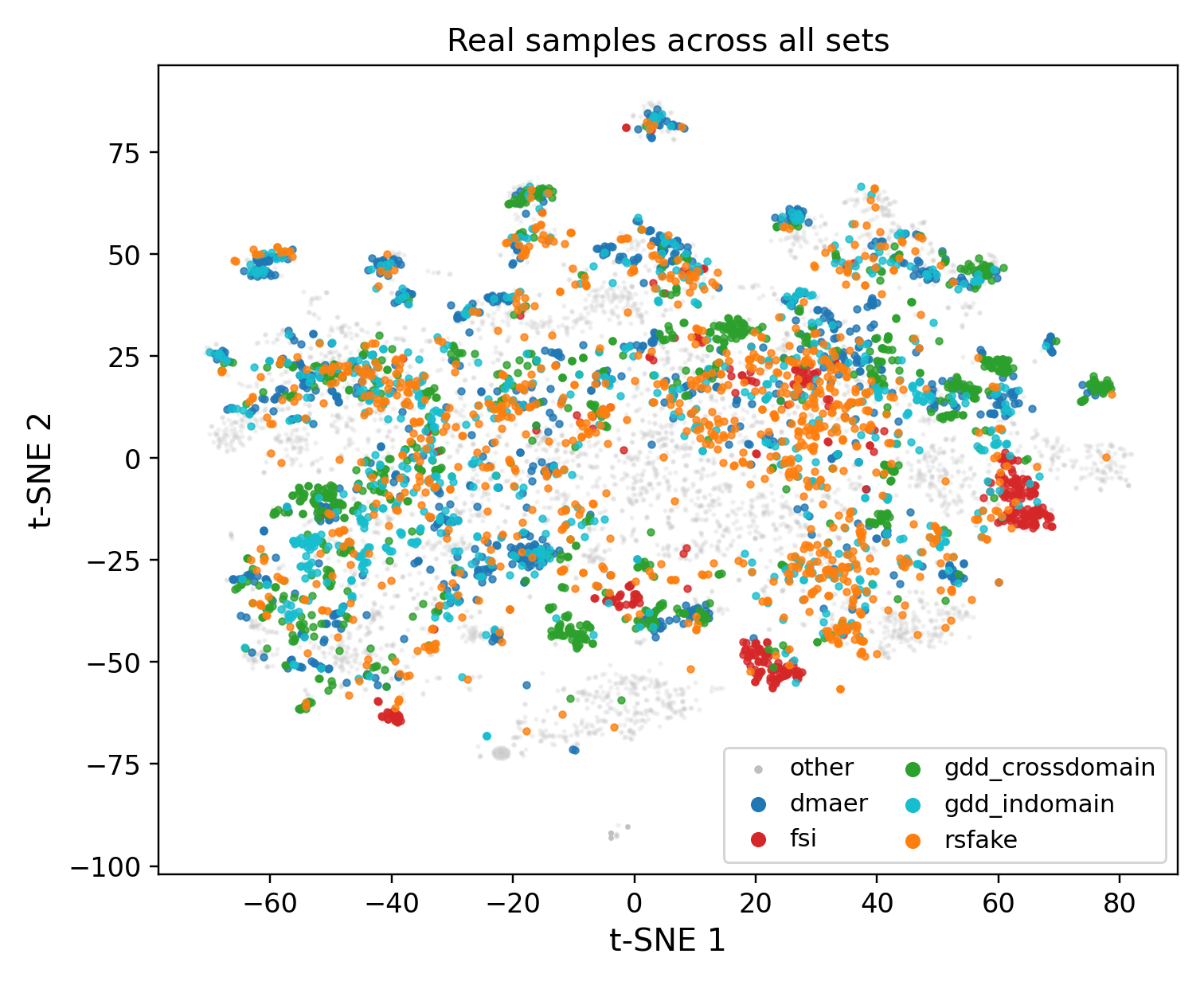}
        \caption{Real images by dataset}
        \label{fig:lvd_tsne_real}
    \end{subfigure}
    \hfill
    \begin{subfigure}[t]{0.23\textwidth}
        \centering
        \includegraphics[width=\linewidth]{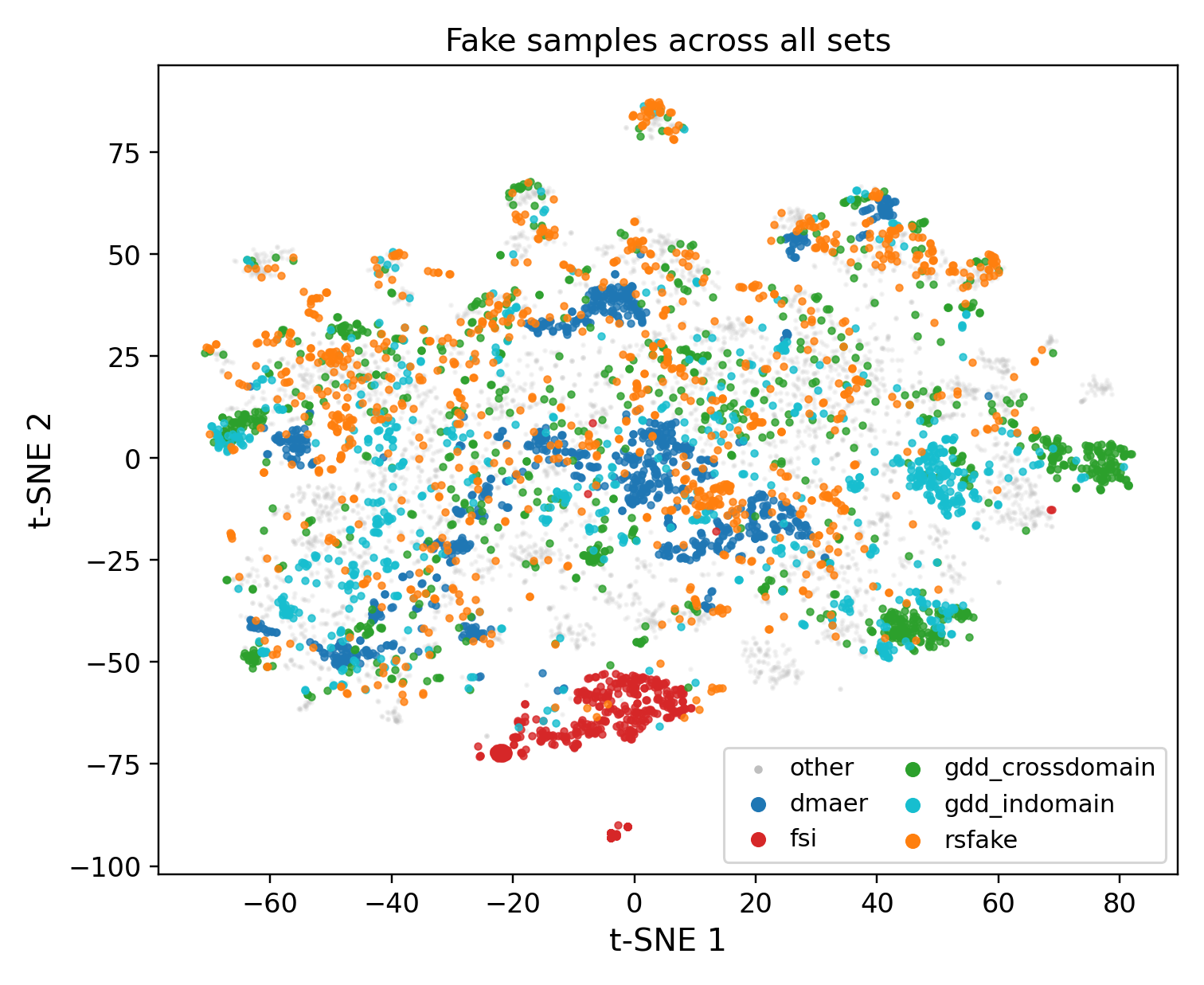}
        \caption{Fakes by dataset}
        \label{fig:lvd_tsne_fakes}
    \end{subfigure}
    \hfill
    \begin{subfigure}[t]{0.23\textwidth}
        \centering
        \includegraphics[width=\linewidth]{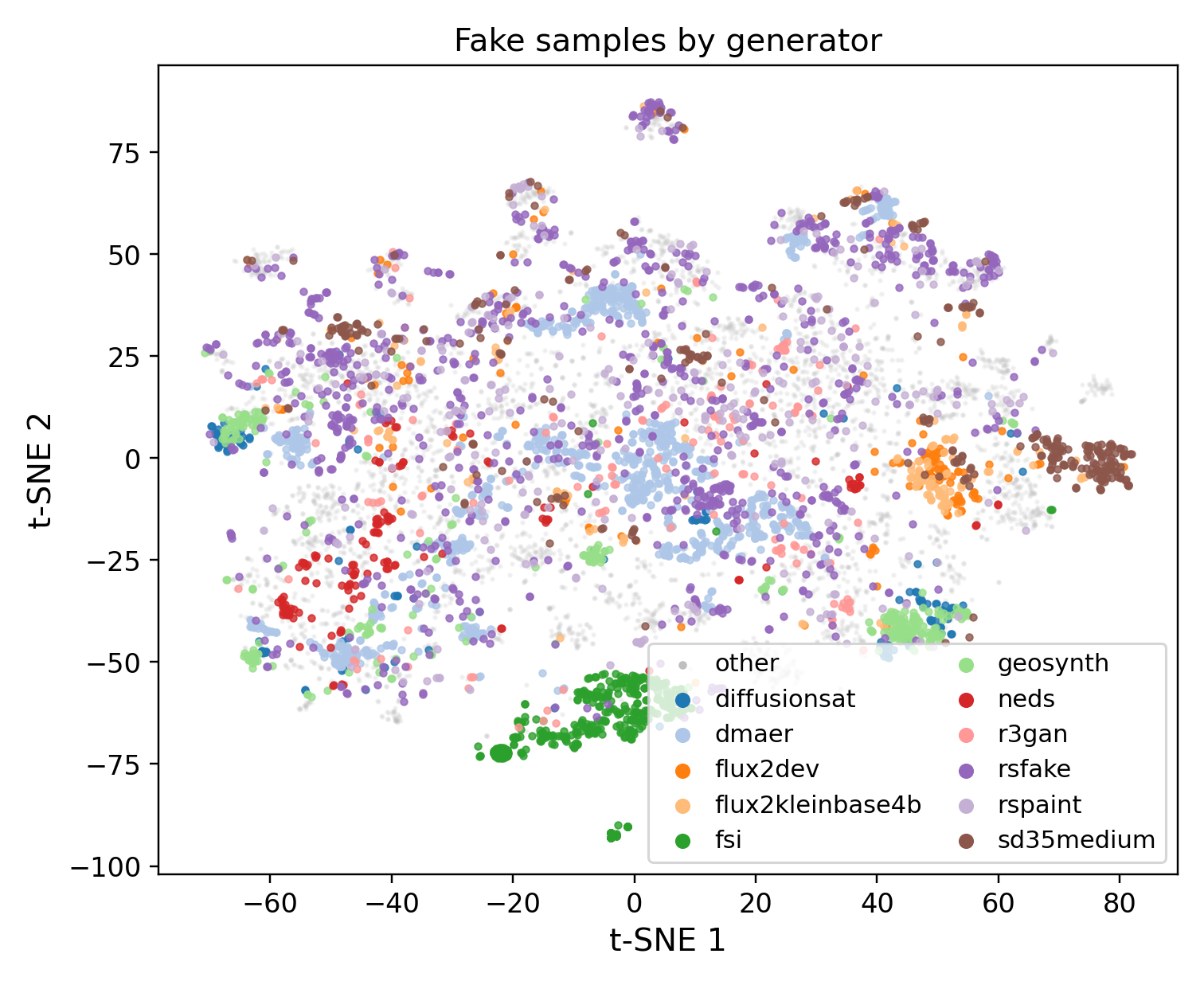}
        \caption{Fakes by generator}
        \label{fig:lvd_tsne_generator}
    \end{subfigure}
    \hfill
    \begin{subfigure}[t]{0.23\textwidth}
        \centering
        \includegraphics[width=\linewidth]{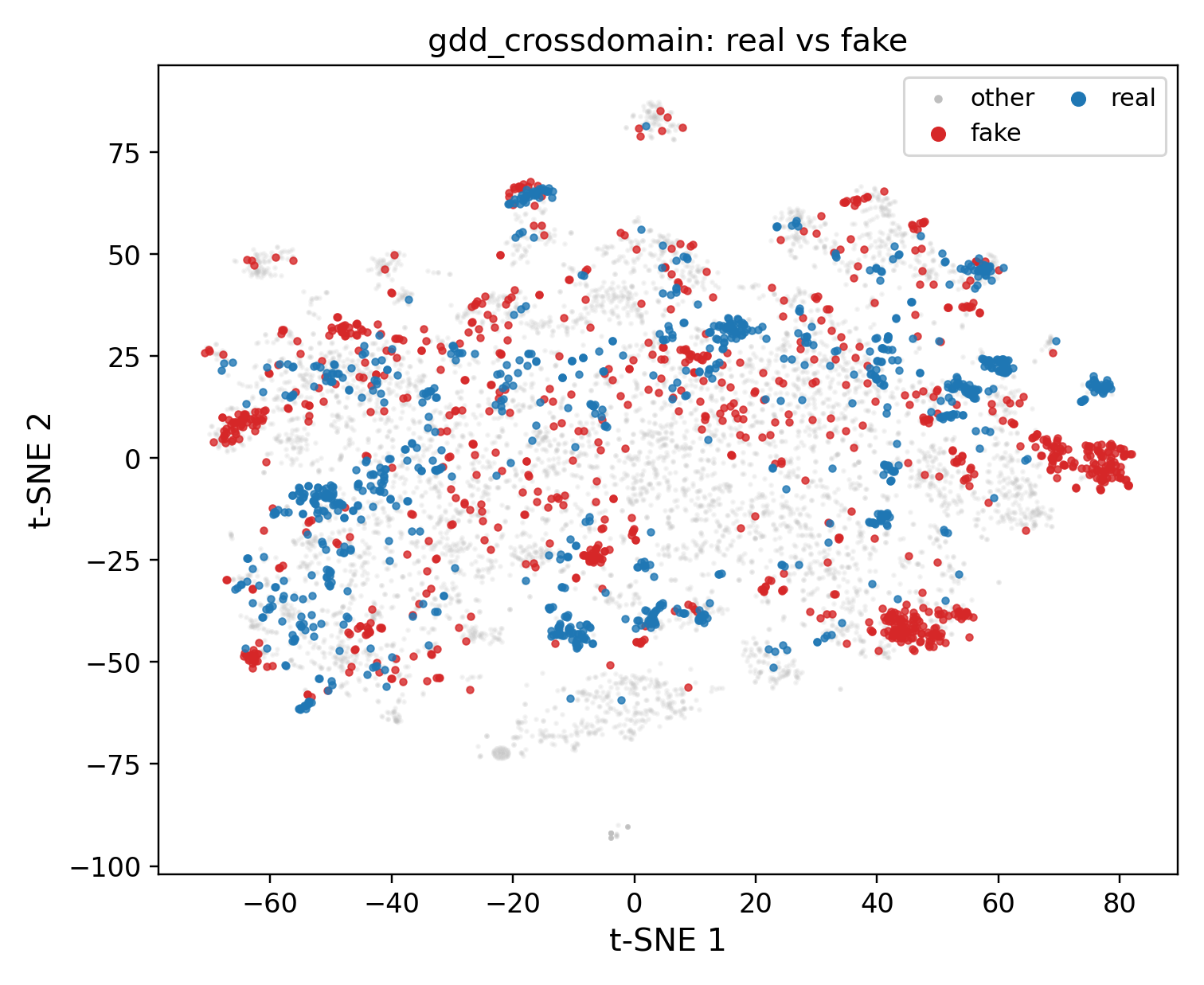}
        \caption{GDD (CD): Real/Fake}
        \label{fig:lvd_tsne_gddcross}
    \end{subfigure}
    
    \medskip 
    
    \begin{subfigure}[b]{0.23\textwidth}
        \centering
        \includegraphics[width=\linewidth]{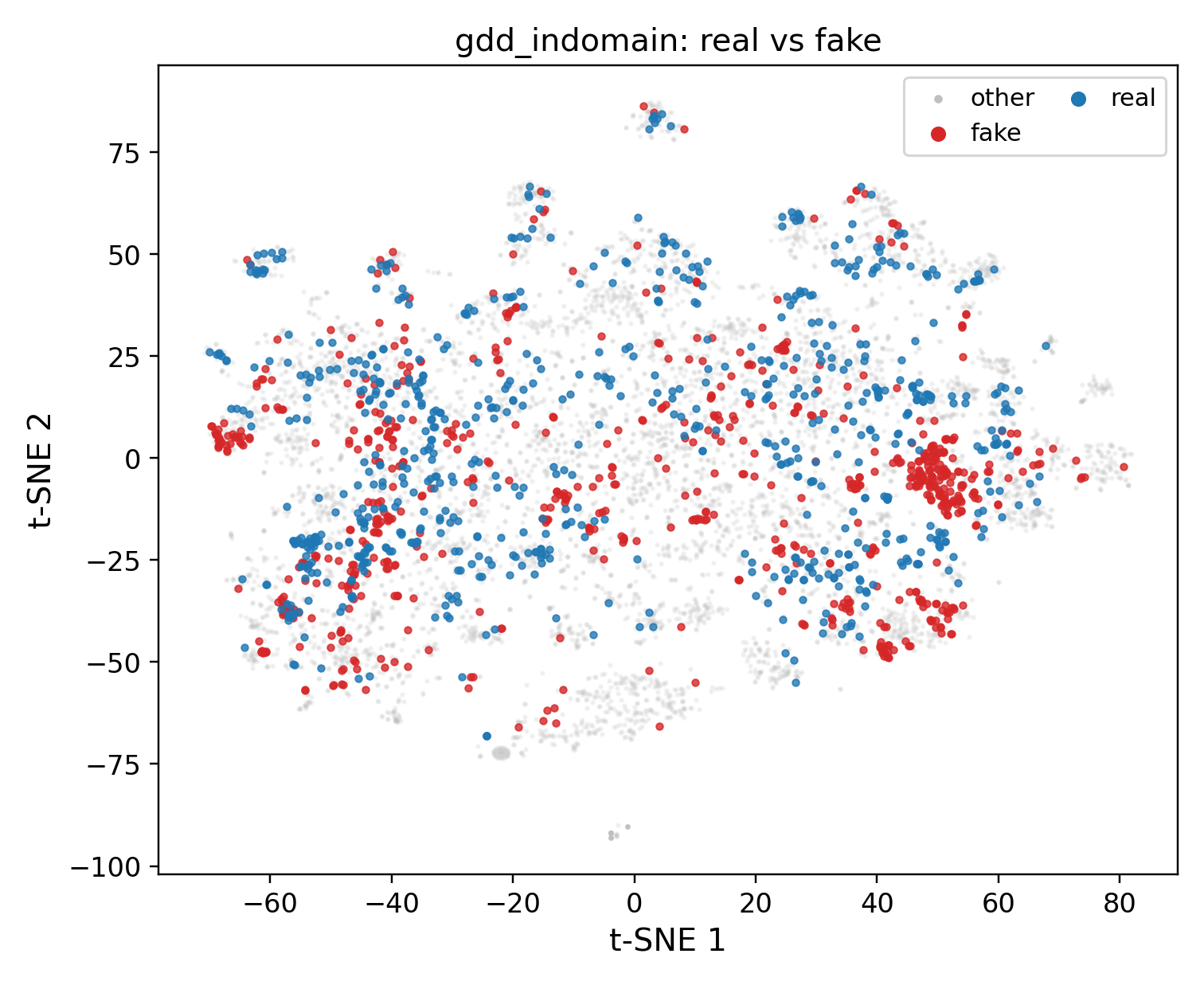}
        \caption{GDD (ID): Real/Fake}
        \label{fig:lvd_tsne_gddin}
    \end{subfigure}
    \hfill
    \begin{subfigure}[b]{0.23\textwidth}
        \centering
        \includegraphics[width=\linewidth]{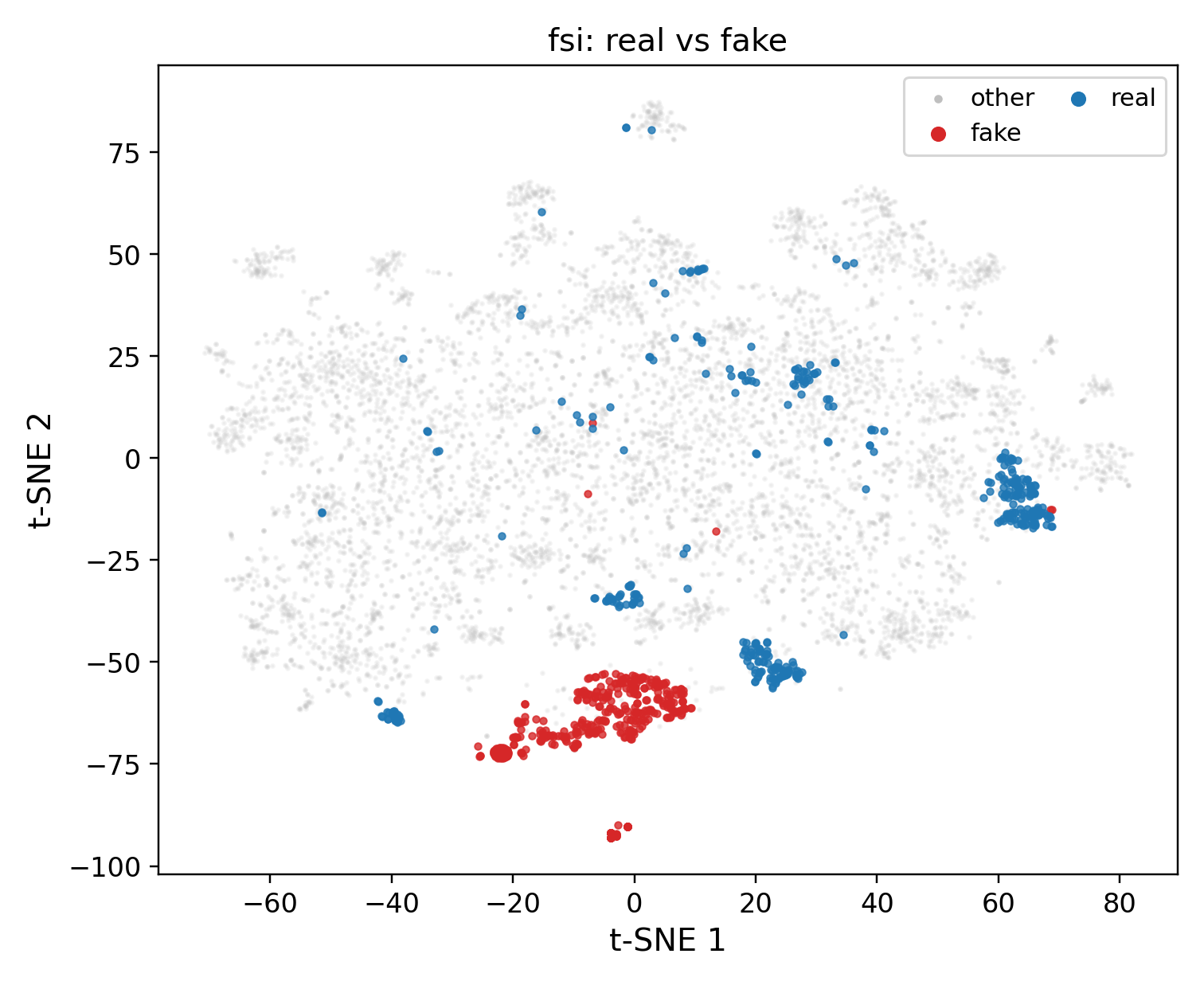}
        \caption{FSI: Real/Fake}
        \label{fig:lvd_tsne_fsi}
    \end{subfigure}
    \hfill
    \begin{subfigure}[b]{0.23\textwidth}
        \centering
        \includegraphics[width=\linewidth]{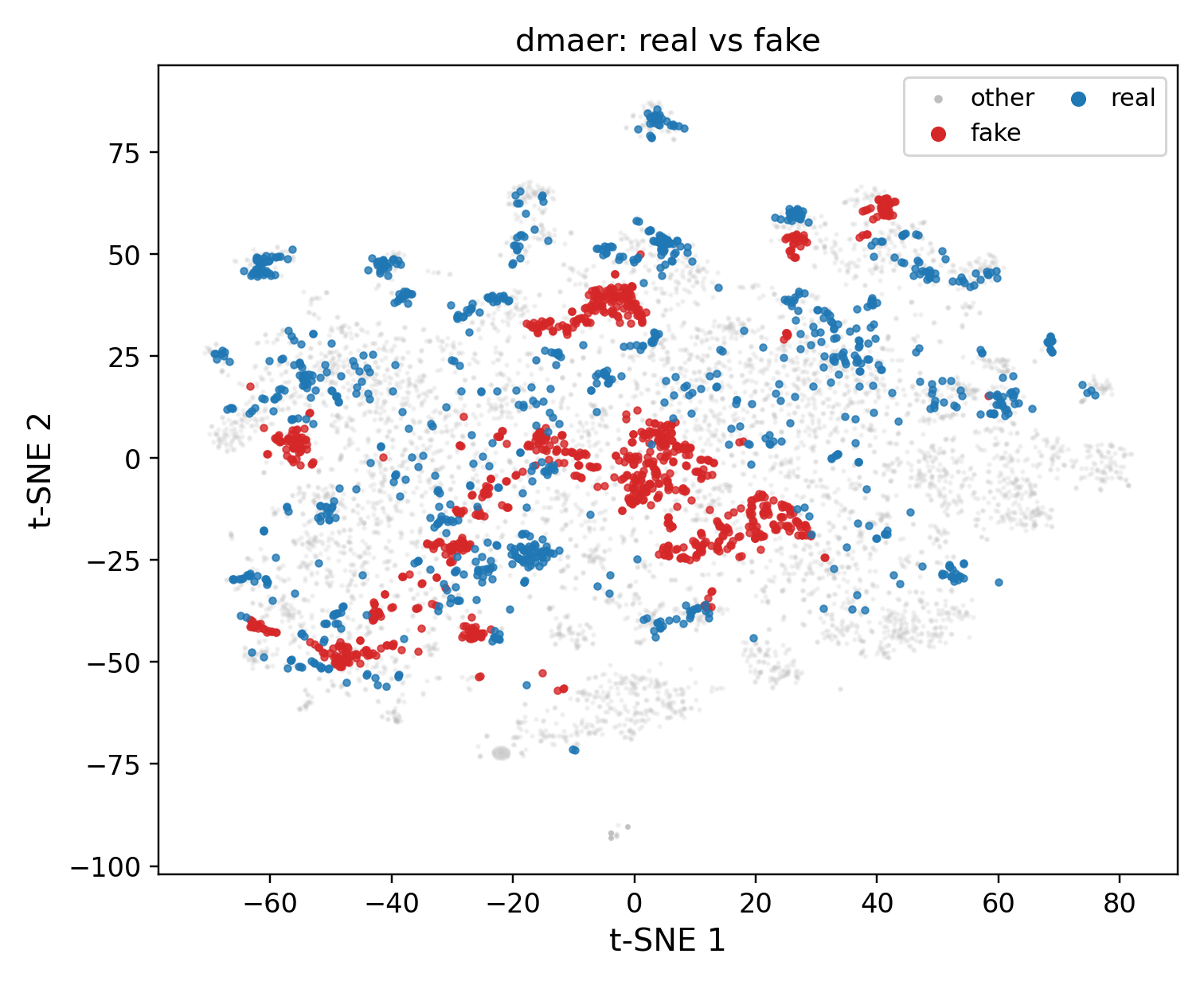}
        \caption{DM-AER: Real/Fake}
        \label{fig:lvd_tsne_dmaer}
    \end{subfigure}
    \hfill
    \begin{subfigure}[b]{0.23\textwidth}
        \centering
        \includegraphics[width=\linewidth]{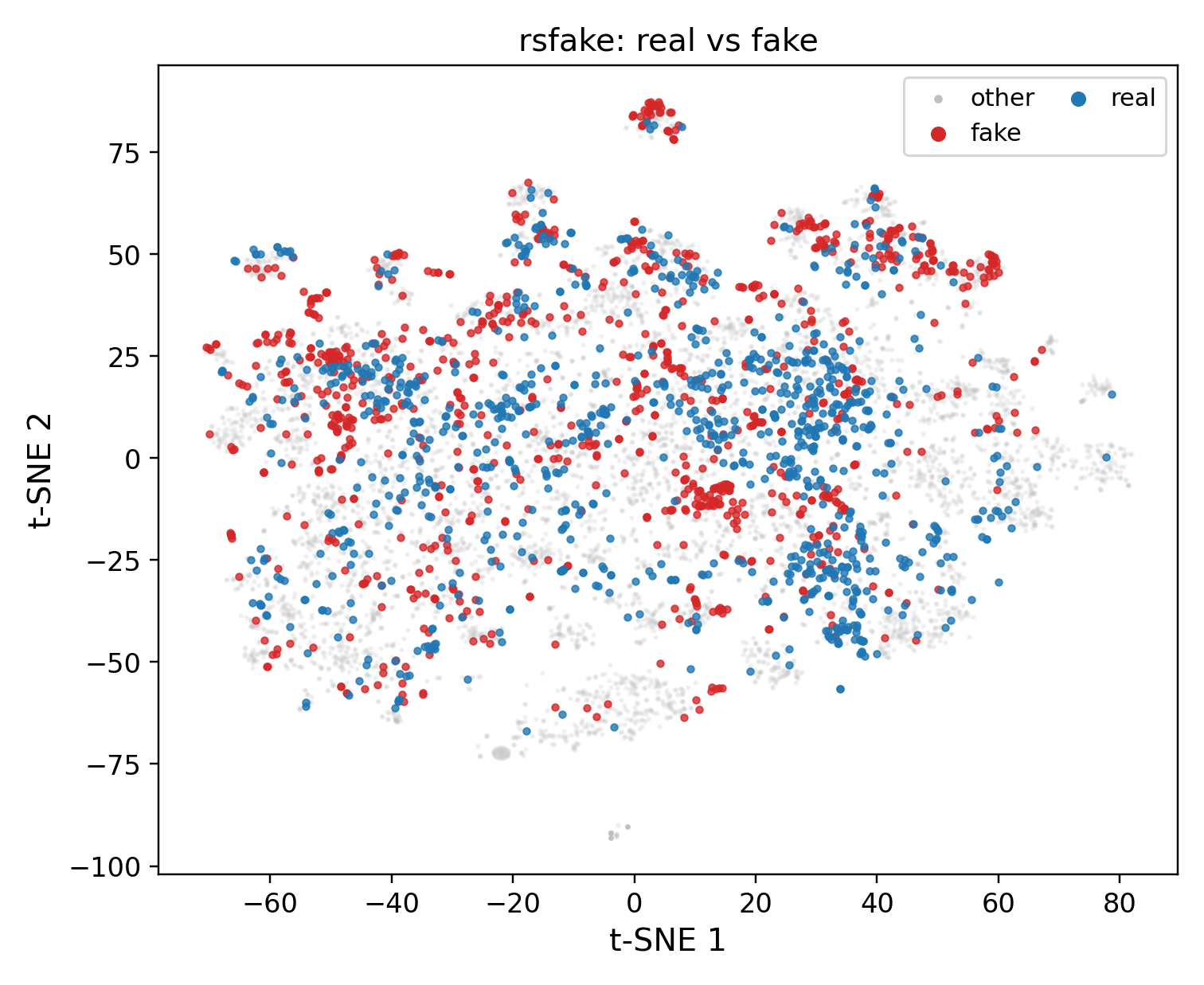}
        \caption{RSFAKE-1M: Real/Fake}
        \label{fig:lvd_tsne_rsfake}
    \end{subfigure}
    
    \caption{Illustration of t-SNE representations using DINOv3-Large-LVD for samples composing test sets of GDD (ID: In-Domain and CD: Cross-Domain), FSI, RSFAKE-1M, and DM-AER datasets}
    \label{fig:lvd_tsne}
\end{figure}

To help illustrating and guiding discussions on dataset diversity and transferability, in computed t-SNE plots based on embedded representations collected from [CLS] token when processing inputs with the DINOv3-Large-SAT and DINOv3-Large-LVD models, in Figures ~\ref{fig:tsne} and ~\ref{fig:lvd_tsne} respectively. Each Figure display samples across test samples composing each datasets, with decompositions per dataset, per fake generator, as well as contrasting real vs fake images for each dataset under a single shared t-SNE projection space. For RSFAKE-1M and DM-AER, we randomly sample 1000 real and 1000 fake images from corresponding test sets.

Subplots~\ref{fig:tsne_real}-~\ref{fig:tsne_fakes} demonstrate the diversity of GDD samples, which compose multiple clusters spread across both real and fake image distribution spaces. Compared to for example FSI, several GDD samples are close in the embedding space to samples from other datasets, contributing to its higher transferability observed in experimental results.

The subplots per dataset ~\ref{fig:tsne_gddcross}-~\ref{fig:tsne_rsfake} illustrate how samples composing GDD and RSFAKE-1M are less trivial to classify between real and fake classes as compared to DM-AER and particularly FSI. Compared to RSFAKE-1M, GDD's higher diversity in terms of real data sources also becomes evident and contributes to harder classification.  

\subsection{Deepfake Detection}
\label{supp:deepfake-detection}
\subsubsection{Implementation Details}
For a fairer comparison across the different deepfake detection models assessed in our benchmarking experiments, we limited all models to $10,000$ training iterations, batch size $BS=128$. Validation is conducted every $1000$ iterations, with training early stopping set available after a minimum of $2,000$ iterations has been completed, with a patience of $4,000$ iterations (4 validation steps) based on validation loss. Verification of loss curves indicate convergence of all models in each scenario. Learning rates are set based on original implementation of each model, with $LR=1e-4$ for UnivFD, AIDE, SimLBR, and DINOv3-supervised models. For Co-SPY, $LR=2e-4$ for semantic and artifact modules and $LR=2E-2$ for fusion module, which corresponds to scaling by $2\times$ its original LR listed in official implementation for a $BS=32$ (i.e., we scale the LR by the square-root of factor of batch size increase). We maintain the augmentation choices made by each model's original implementation: AIDE adopts random JPEG compression (QF sampled uniformly $[30,100]$) and random Gaussian blur ($\sigma$ sampled uniformly $[0.1,3.0]$, UnivFD uses Blur + JPEG with $0.5$ probability, while Co-SPY only uses feature interpolation, SimLBR only latent blending regularization.

\subsubsection{Ablation on DINOv3 Checkpoints.} 
Using the GDD dataset, we performed experiments with three DINOv3 variants: Base-LVD weights, Large-LVD weights, and Large-SAT. While LVD variants are trained on $1.689Bi$ general web images, the SAT version is trained on $493M$ high-resolution RGB ortho-rectified satellite images ($512\times512px$ at $0.6m/px$). 
Table~\ref{tab:dinov3_ablation} indicates ViT-Large variants outperform the base version, but surprisingly, both the supervised model as well as SimLBR variants using DINOv3-Large-LVD outperform the Large-SAT counterparts. This contrasts to the assumption that a model trained exclusively on remote sensing imagery would yield better descriptors, such as argued in SimLBR~\cite{dhakal2026simlbr}. Instead, the higher sample diversity of LVD seems to provide features that enable higher generalization for cross-domain deepfake detection for remote sensing imagery. 

\begin{table}[t]
    \centering
    \small
    \setlength{\tabcolsep}{4pt}
    \renewcommand{\arraystretch}{1.2}
    \begin{tabular}{l ccc}
    \toprule
    \textbf{Method} & \textbf{ID} & \textbf{CG} & \textbf{CD} \\
    \midrule
    Supervised-Base 
        & \cellcolor{great}0.98 \scriptsize{(0.98 / 0.98)} 
        & \cellcolor{good}0.82 \scriptsize{(0.99 / 0.65)} 
        & \cellcolor{fair}0.79 \scriptsize{(0.92 / 0.65)} \\
    Supervised-LVD 
        & \cellcolor{excellent}0.99 \scriptsize{(0.97 / 0.99)} 
        & \cellcolor{good}0.85 \scriptsize{(0.97 / 0.72)} 
        & \cellcolor{good}0.82 \scriptsize{(0.93 / 0.72)} \\
    Supervised-SAT 
        & \cellcolor{excellent}0.99 \scriptsize{(0.99 / 0.99)} 
        & \cellcolor{good}0.85 \scriptsize{(0.99 / 0.70)} 
        & \cellcolor{fair}0.77 \scriptsize{(0.84 / 0.70)} \\
    SimLBR-LVD 
        & \cellcolor{great}0.94 \scriptsize{(0.92 / 0.95)} 
        & \cellcolor{good}0.85 \scriptsize{(0.92 / 0.78)} 
        & \cellcolor{fair}0.79 \scriptsize{(0.79 / 0.78)} \\
    SimLBR-SAT 
        & \cellcolor{great}0.98 \scriptsize{(0.95 / 1.00)} 
        & \cellcolor{good}0.89 \scriptsize{(0.95 / 0.83)} 
        & \cellcolor{poor}0.64 \scriptsize{(0.45 / 0.83)} \\
    \bottomrule
    \end{tabular}
    \caption{GDD in-domain (\textbf{ID}), cross-generator (\textbf{CG}), and cross-domain (\textbf{CD}) performance for supervised and SimLBR models under various DINOv3 version. Scores indicate overall accuracy and class-wise \textit{Real / Fake} accuracy in parentheses.}
    \label{tab:dinov3_ablation}
\end{table}

\subsubsection{Comparing non-RS Pretrained Deepfake Detection Models}
Table~\ref{tab:aide_cospy_gdd_conditions} contrast detection results on GDD test sets provided by ``off-the-shelf'' AIDE and Co-SPY models, in comparison with our versions trained on GDD train set. Note that the concept of in-domain vs cross-domain/-generator only applies for GDD trained models. For variants trained on other datasets, the three test splits equally represent domain transfer from non-RS to RS deepfake detection. Unsurprisingly, both models perform best when trained on GDD and evaluated in in-domain (ID). For AIDE, the GDD trained variant also performs best in terms of overall accuracy (OA) when transferred to cross-generator (CG) and cross-domain (CD) scenarios, being outperformed only in terms of Fake Accuracy by the variants trained on ProGAN and GenImage. For both AIDE and Co-SPY the variants trained on ProGAN yield better Fake Accuracy than the SD1.4-trained counterparts. This is rather surprising, since fakes in CG/CD splits are diffusion based. We conjecture that ProGAN's higher sample count/diversity may be a reason. In terms of overall accuracy, Co-SPY trained on SD1.4 performs better than the ProGAN, but with trade-offs in real vs fake accuracy: the variant trained on SD14 identifies real images well, while the variant trained on ProGAN yields better fake accuracy.

\begin{table}[t]
    \centering
    \footnotesize
    \setlength{\tabcolsep}{3pt}
    \renewcommand{\arraystretch}{1.2}
    \begin{tabular}{ll ccc}
    \toprule
    \textbf{Method} & \textbf{Train} & \textbf{ID} & \textbf{CG} & \textbf{CD} \\
    \midrule
    \multirow{4}{*}{AIDE} 
        & ProGAN 
        & \cellcolor{good}0.80 \scriptsize{(0.82 / 0.77)} 
        & \cellcolor{good}0.83 \scriptsize{(0.81 / 0.85)} 
        & \cellcolor{fair}0.73 \scriptsize{(0.61 / 0.85)} \\
        & SD14 
        & \cellcolor{poor}0.61 \scriptsize{(0.72 / 0.50)} 
        & \cellcolor{fair}0.75 \scriptsize{(0.73 / 0.77)} 
        & \cellcolor{poor}0.65 \scriptsize{(0.52 / 0.77)} \\
        & GenImage 
        & \cellcolor{fair}0.73 \scriptsize{(0.74 / 0.72)} 
        & \cellcolor{good}0.80 \scriptsize{(0.73 / 0.87)} 
        & \cellcolor{poor}0.64 \scriptsize{(0.42 / 0.87)} \\
        & GDD 
        & \cellcolor{excellent}\textbf{0.99} \scriptsize{(0.99 / 0.99)} 
        & \cellcolor{good}\textbf{0.89} \scriptsize{(0.98 / 0.79)} 
        & \cellcolor{good}\textbf{0.81} \scriptsize{(0.82 / 0.79)} \\
    \midrule
    \multirow{3}{*}{Co-SPY} 
        & ProGAN 
        & \cellcolor{poor}0.61 \scriptsize{(0.41 / 0.82)} 
        & \cellcolor{poor}0.60 \scriptsize{(0.43 / 0.77)} 
        & \cellcolor{verypoor}0.55 \scriptsize{(0.33 / 0.77)} \\
        & SD14 
        & \cellcolor{poor}0.64 \scriptsize{(0.83 / 0.44)} 
        & \cellcolor{fair}0.72 \scriptsize{(0.84 / 0.60)} 
        & \cellcolor{fair}0.73 \scriptsize{(0.85 / 0.60)} \\
        & GDD 
        & \cellcolor{great}\textbf{0.95} \scriptsize{(0.90 / 1.00)} 
        & \cellcolor{good}\textbf{0.86} \scriptsize{(0.89 / 0.83)} 
        & \cellcolor{poor}0.69 \scriptsize{(0.54 / 0.83)} \\
    \bottomrule
    \end{tabular}
    \caption{AIDE and Co-SPY performance across training sets on GDD In-Domain (\textbf{ID}), Cross-Generator (\textbf{CG}), and Cross-Domain (\textbf{CD}) evaluations. Scores indicate overall accuracy and class-wise \textit{Real / Fake} accuracy in parentheses.}
    \label{tab:aide_cospy_gdd_conditions}
\end{table}

\subsubsection{Results per Generator and Qualitative Examples}

\begin{figure}[h]
    \centering
    \includegraphics[width=\linewidth]{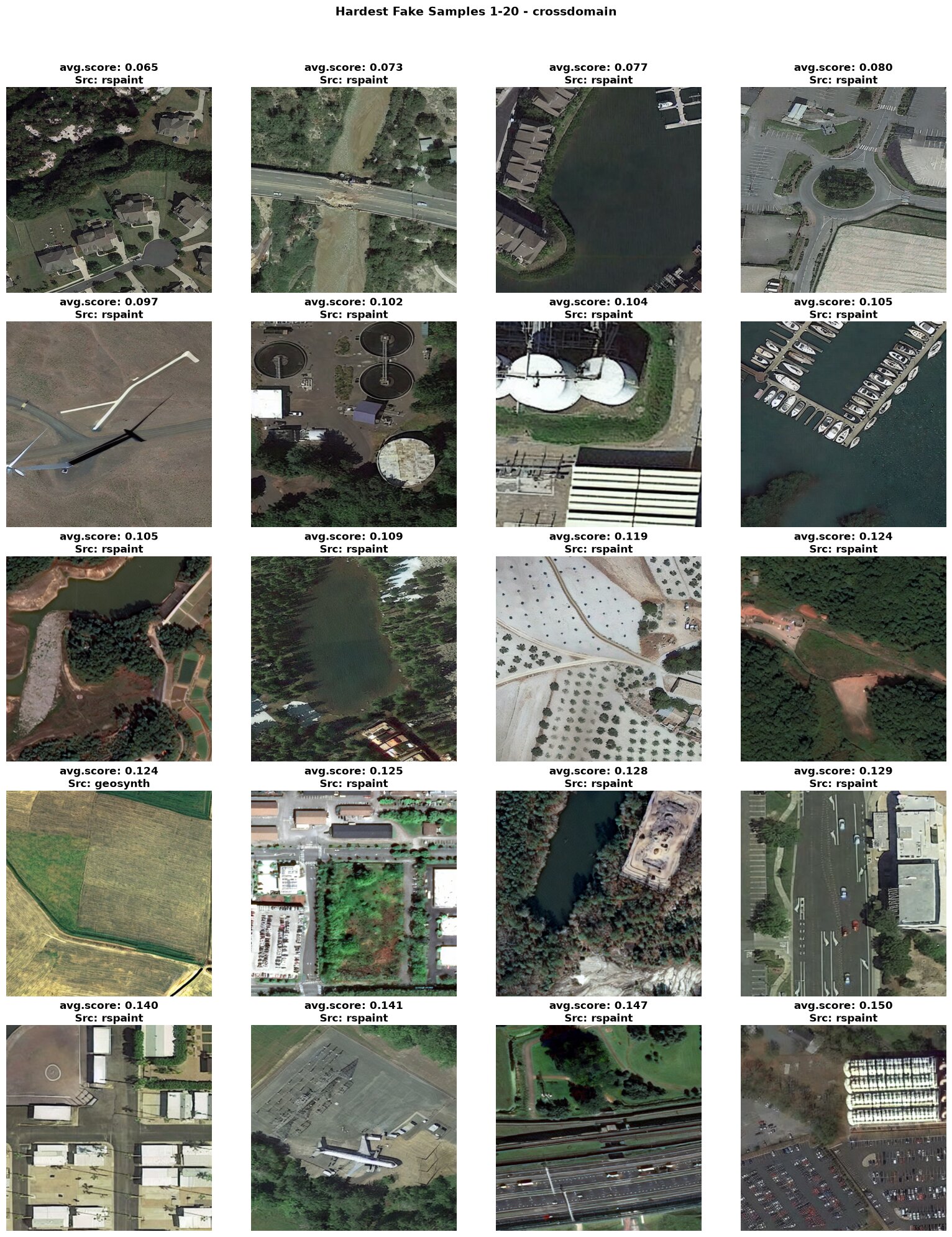}
    \caption{Top-20 hardest fake samples in GDD cross-domain test set, as measured based on average predicted score by UnivFD, AIDE, Co-SPY, DINOv3-sup, and SimLBR trained on GDD train set.}
    \label{fig:hard-cd-fake}
\end{figure}

\begin{figure}[h]
    \centering
    \includegraphics[width=\linewidth]{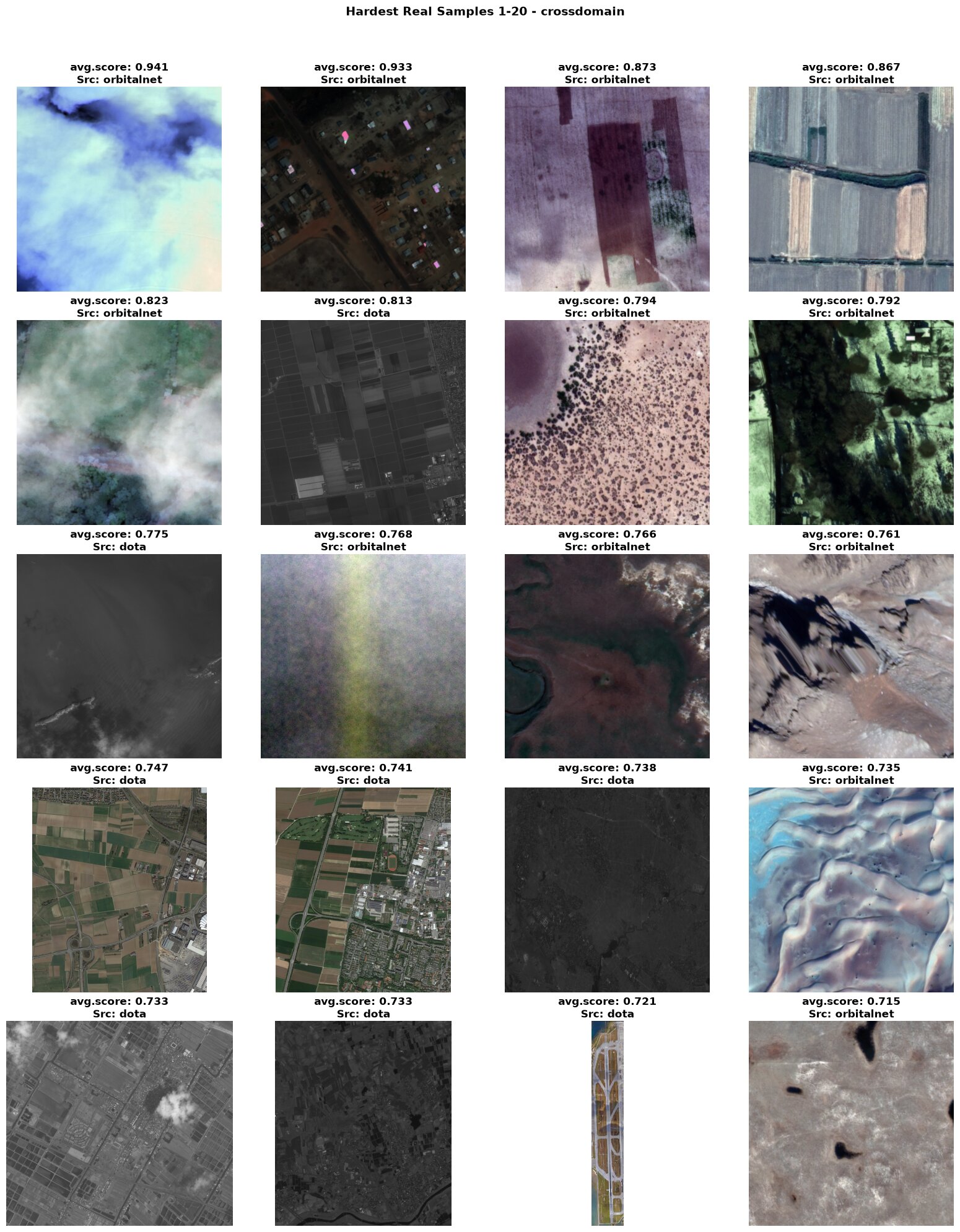}
    \caption{Top-20 hardest real samples in GDD cross-domain test set, as measured with same criteria as Fig \ref{fig:hard-cd-fake}.}
    \label{fig:hard-cd-real}
\end{figure}

\begin{figure}[h]
    \centering
    \includegraphics[width=\linewidth]{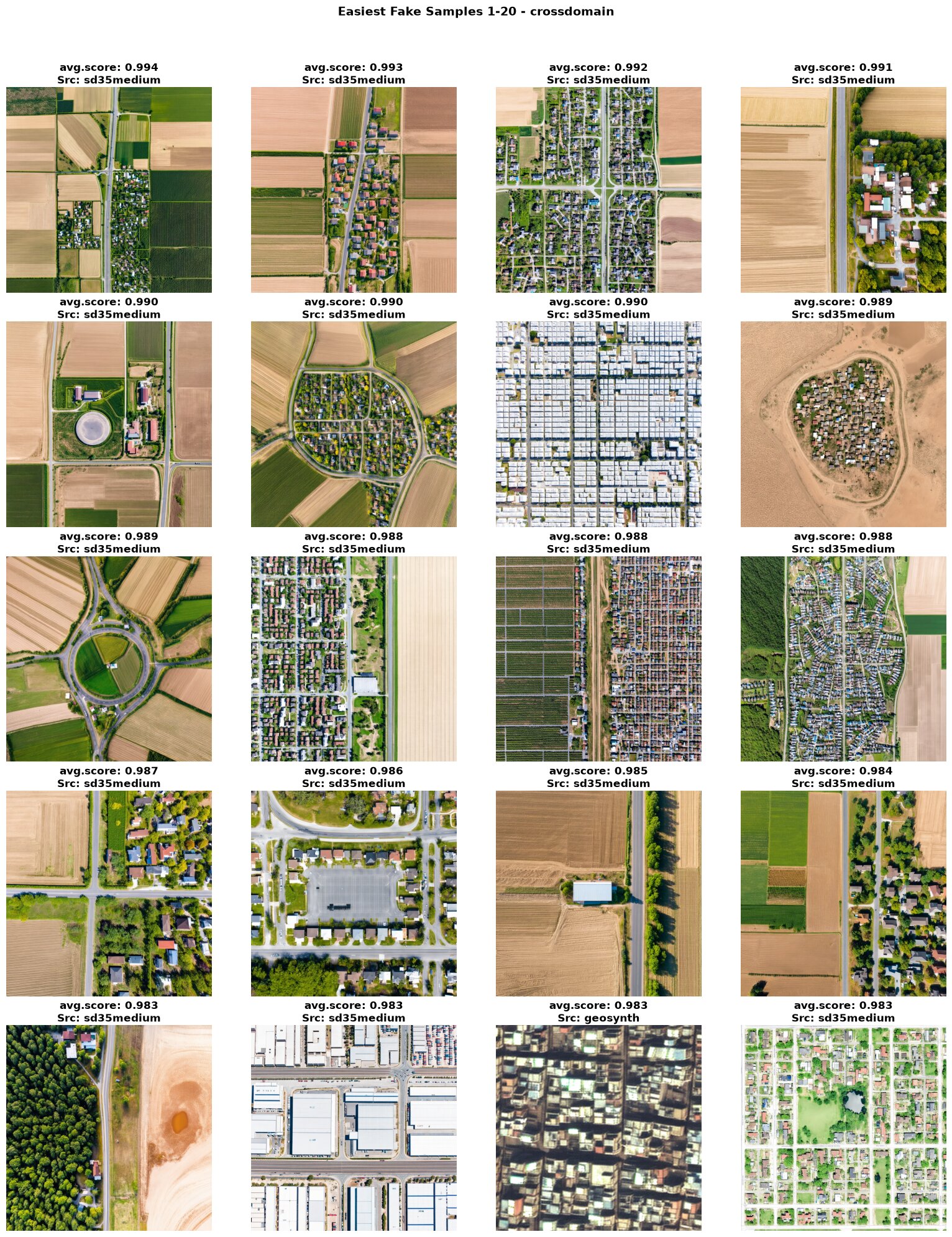}
    \caption{Top-20 easiest fake samples in GDD cross-domain test set, as measured with same criteria as Fig \ref{fig:hard-cd-fake}.}
    \label{fig:easy-cd-fake}
\end{figure}

\begin{figure}[h]
    \centering
    \includegraphics[width=\linewidth]{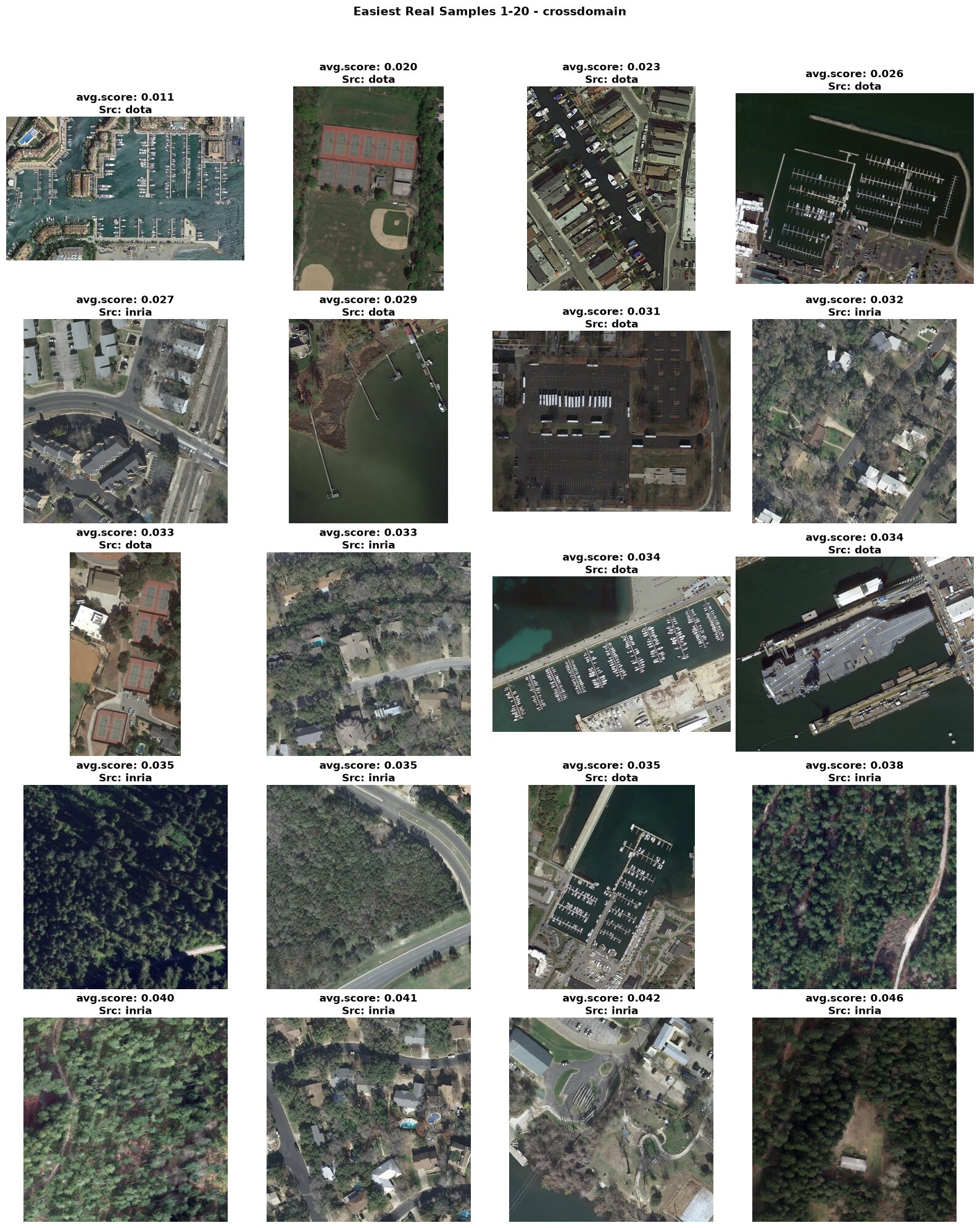}
    \caption{Top-20 easiest real samples in GDD cross-domain test set, as measured with same criteria as Fig \ref{fig:hard-cd-fake}.}
    \label{fig:easy-cd-real}
\end{figure}

\begin{figure}[h]
    \centering
    \includegraphics[width=\linewidth]{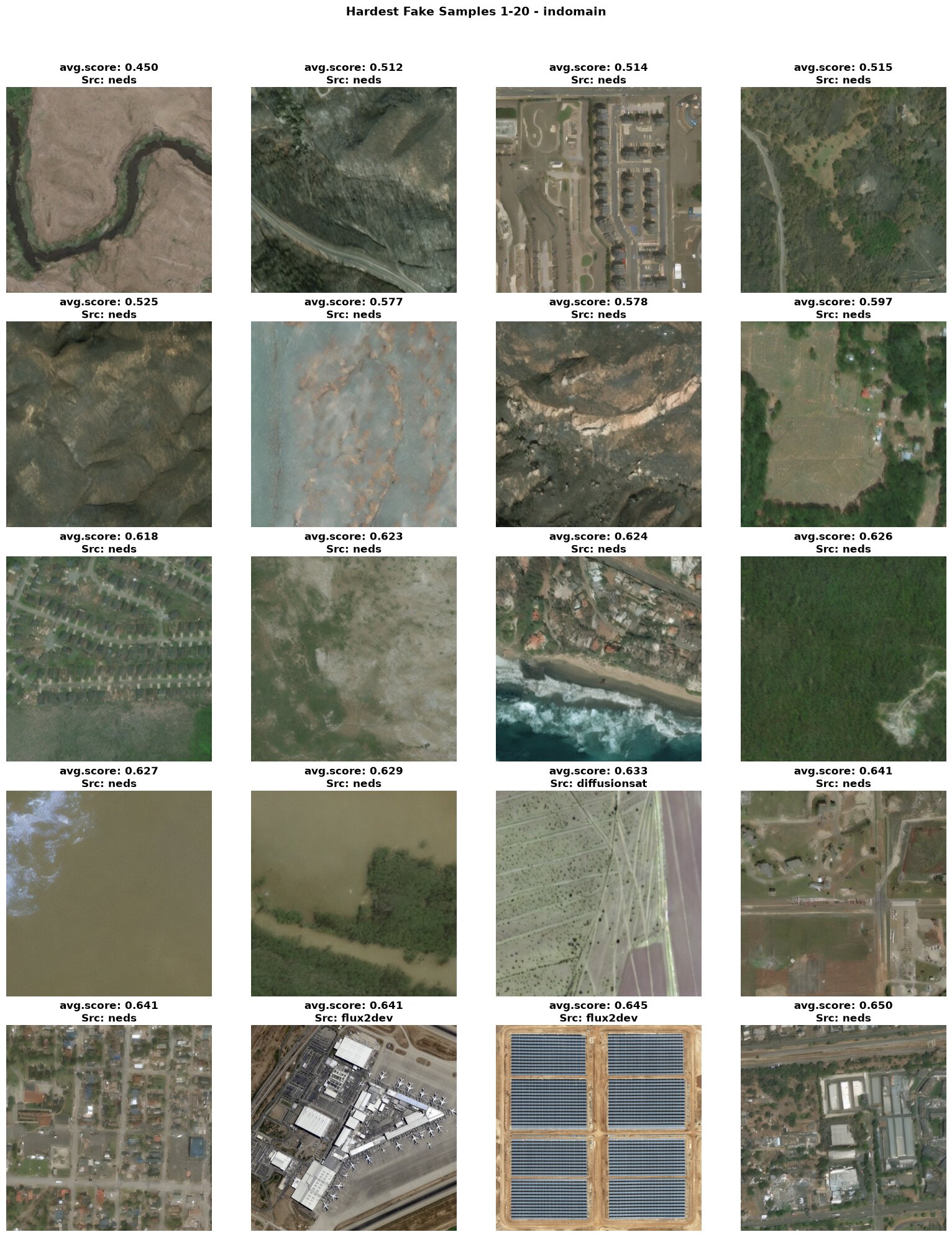}
    \caption{Top-20 hardest fake samples in GDD in-domain test set, as measured with same criteria as Fig \ref{fig:hard-cd-fake}.}
    \label{fig:hard-id-fake}
\end{figure}

\begin{figure}[h]
    \centering
    \includegraphics[width=\linewidth]{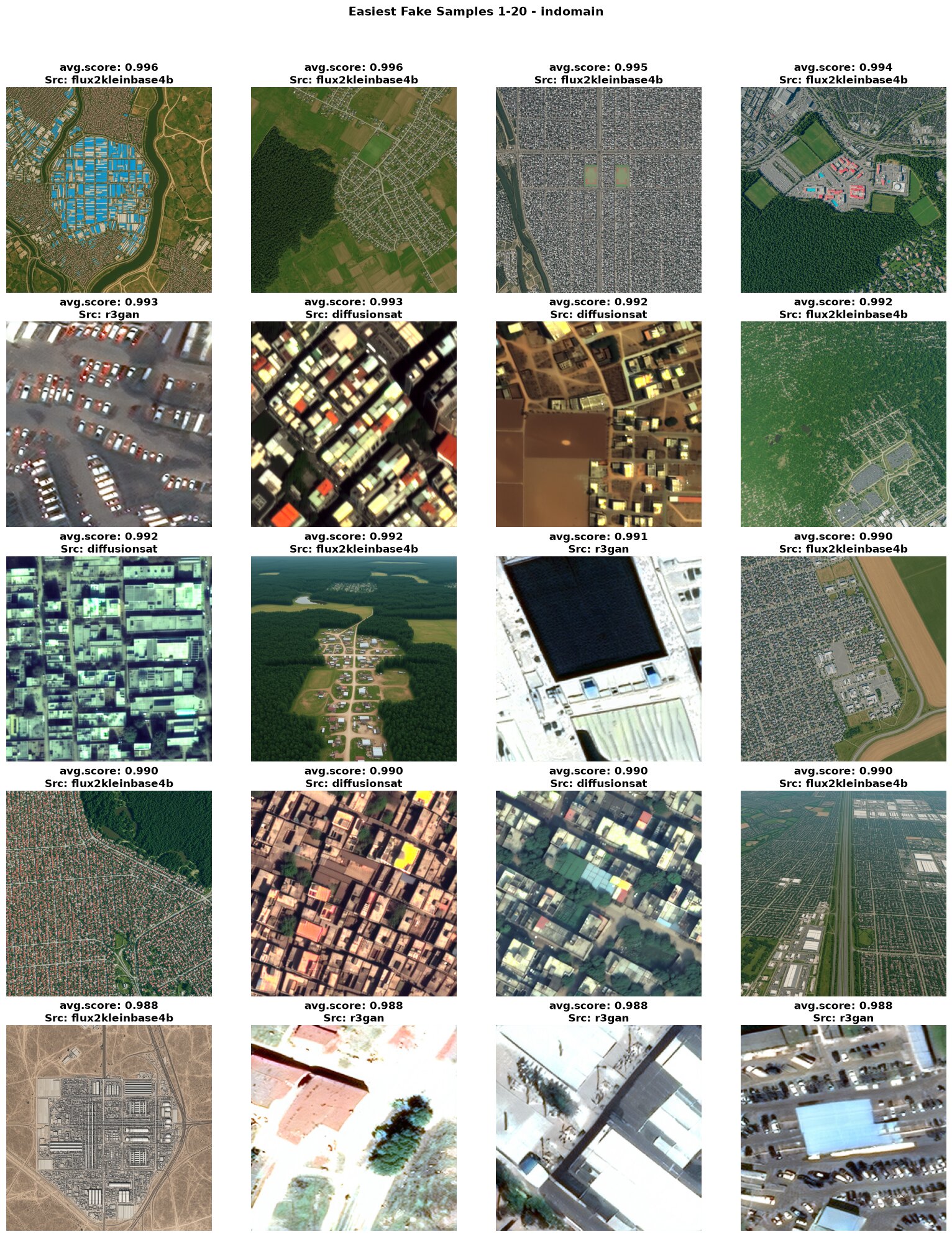}
    \caption{Top-20 easiest fake samples in GDD in-domain test set, as measured with same criteria as Fig \ref{fig:hard-cd-fake}.}
    \label{fig:easy-id-fake}
\end{figure}

Table \ref{tab:cd_confusion_matrix_source_dataset} provides results in terms of confusion matrix counts across each source dataset/generator for real and fakes in the GDD in-domain test split, while Table \ref{tab:id_confusion_matrix_source_dataset} provides analogous information for GDD cross-domain samples. Figures \ref{fig:hard-cd-fake}-\ref{fig:easy-id-fake} provide qualitative examples of the top-20 hardest/easiest fake/real samples for both test splits. 

In-domain analysis shows \textit{neds} as containing most of the hardest samples, while cross-domain analysis highlights RSPaint samples. In common, both datasets consist of samples generated by diffusion models tasked with image manipulation (localized disaster simulation and inpainting, respectively). Although both models fully synthesize image based on editing prompts, as corroborated by visual analysis their outputs are highly realistic when contrasted to xBD and fmow real samples, for instance. On the other side of the spectrum, several \textit{sd35medium}-generated fakes are notoriously easy to detect, sharing patterns such as symmetry with respect to image center, and grid-like distribution of tiny buildings. For indomain samples, multiple \textit{flux2klein}-generated fakes are easy to spot, whether from from the rather off-nadir view angle, or the unusual texture of small structures

Regarding real samples, Fig.~\ref{fig:easy-cd-real} shows detectors are very accurate for samples with high-resolution and presence of multiple structures such as trees, boats, vehicles, sports courts, which provide rich features both from low-level and semantic perspectives. In turns, Table~\ref{tab:cd_confusion_matrix_source_dataset} shows transferability to \textit{orbitalnet} and, to a lower extent, \textit{dota} samples as the hardest cases. Examples in Fig.~\ref{fig:hard-cd-real} show that hardest samples from orbitalnet include cases of cloud-cover, ``featureless'' landscapes (i.e., without buildings/trees), and low contrast. For DOTA, hard examples include grayscale (panchromatic-derived) images.

\begin{table*}[h]
\centering
\caption{Confusion-matrix counts by source dataset, for models trained on GDD and evaluated on GDD in-domain.}
\label{tab:id_confusion_matrix_source_dataset}
\resizebox{\textwidth}{!}{%
\begin{tabular}{lcccccccccccccccc}
\toprule
\textbf{Model} & \multicolumn{10}{c}{\textbf{fake}} & \multicolumn{6}{c}{\textbf{real}} \\
\cmidrule(lr){2-11} \cmidrule(lr){12-17}
 & \multicolumn{2}{c}{\textbf{diffusionsat}} & \multicolumn{2}{c}{\textbf{flux2dev}} & \multicolumn{2}{c}{\textbf{flux2kleinbase4b}} & \multicolumn{2}{c}{\textbf{neds}} & \multicolumn{2}{c}{\textbf{r3gan}} & \multicolumn{2}{c}{\textbf{fmow}} & \multicolumn{2}{c}{\textbf{millionaid}} & \multicolumn{2}{c}{\textbf{xbd}} \\
\cmidrule(lr){2-3} \cmidrule(lr){4-5} \cmidrule(lr){6-7} \cmidrule(lr){8-9} \cmidrule(lr){10-11} \cmidrule(lr){12-13} \cmidrule(lr){14-15} \cmidrule(lr){16-17}
 & \textbf{TP} & \textbf{FN} & \textbf{TP} & \textbf{FN} & \textbf{TP} & \textbf{FN} & \textbf{TP} & \textbf{FN} & \textbf{TP} & \textbf{FN} & \textbf{TN} & \textbf{FP} & \textbf{TN} & \textbf{FP} & \textbf{TN} & \textbf{FP} \\
\midrule
UnivFD & 140 & 10 & 85 & 65 & 136 & 14 & 56 & 94 & 148 & 2 & 62 & 188 & 227 & 23 & 126 & 124 \\
AIDE & 148 & 2 & 150 & 0 & 150 & 0 & 148 & 2 & 150 & 0 & 244 & 6 & 247 & 3 & 250 & 0 \\
Co-SPY & 150 & 0 & 149 & 1 & 150 & 0 & 149 & 1 & 150 & 0 & 237 & 13 & 232 & 18 & 202 & 48 \\
DINOv3-sup & 149 & 1 & 148 & 2 & 150 & 0 & 143 & 7 & 150 & 0 & 248 & 2 & 247 & 3 & 247 & 3 \\
SimLBR & 149 & 1 & 146 & 4 & 150 & 0 & 122 & 28 & 149 & 1 & 227 & 23 & 235 & 15 & 236 & 14 \\
\bottomrule
\end{tabular}%
}
\end{table*}

\begin{table}[h]
\centering
\caption{Confusion-matrix counts by source dataset, for models trained on GDD and evaluated on GDD cross-domain.}
\label{tab:cd_confusion_matrix_source_dataset}
\resizebox{\columnwidth}{!}{%
\begin{tabular}{lcccccccccccc}
\toprule
\textbf{Model} & \multicolumn{6}{c}{\textbf{fake}} & \multicolumn{6}{c}{\textbf{real}} \\
\cmidrule(lr){2-7} \cmidrule(lr){8-13}
 & \multicolumn{2}{c}{\textbf{geosynth}} & \multicolumn{2}{c}{\textbf{rspaint}} & \multicolumn{2}{c}{\textbf{sd35medium}} & \multicolumn{2}{c}{\textbf{dota}} & \multicolumn{2}{c}{\textbf{inria}} & \multicolumn{2}{c}{\textbf{orbitalnet}} \\
\cmidrule(lr){2-3} \cmidrule(lr){4-5} \cmidrule(lr){6-7} \cmidrule(lr){8-9} \cmidrule(lr){10-11} \cmidrule(lr){12-13}
 & \textbf{TP} & \textbf{FN} & \textbf{TP} & \textbf{FN} & \textbf{TP} & \textbf{FN} & \textbf{TN} & \textbf{FP} & \textbf{TN} & \textbf{FP} & \textbf{TN} & \textbf{FP} \\
\midrule
UnivFD & 265 & 69 & 156 & 177 & 300 & 33 & 255 & 78 & 306 & 28 & 115 & 218 \\
AIDE & 293 & 41 & 173 & 160 & 323 & 10 & 287 & 46 & 300 & 34 & 233 & 100 \\
Co-SPY & 310 & 24 & 192 & 141 & 332 & 1 & 221 & 112 & 255 & 79 & 61 & 272 \\
DINOv3-sup & 291 & 43 & 101 & 232 & 324 & 9 & 308 & 25 & 332 & 2 & 293 & 40 \\
SimLBR & 294 & 40 & 169 & 164 & 321 & 12 & 229 & 104 & 293 & 41 & 246 & 87 \\
\bottomrule
\end{tabular}%
}
\end{table} 

\noindent\textbf{Can Contrast Enhancement Improve Results in OrbitalNet?} Motivated by the identification of several low-contrast images as false positives across detectors, we evaluated whether contrast enhancement as a pre-processing could improve results for OrbitalNet-sourced samples. Results in Table~\ref{tab:orbitalnet_real_accuracy} show significant reduction in false positives by both AIDE and SimLBR when OrbitalNet samples undergo contrast enhancement. For the top-performing DINOv3-sup, however, a rather increase in $+4$ FPs was observed, indicating that preprocessing operations should be considered in a case-by-case fashion. 
\begin{table}[h]
\centering
\caption{Real-source performance on the OrbitalNet-sourced real samples, with vs without contrast enhancement.}
\label{tab:orbitalnet_real_accuracy}
\resizebox{\columnwidth}{!}{%
\begin{tabular}{lrrrr}
\toprule
\textbf{Model} & \textbf{TN} & \textbf{FP} & \textbf{Real acc. ($\%$)} & \textbf{Avg. score} \\
\midrule
AIDE & 233 & 100 & 69.97 & 0.3585 \\
AIDE+ContrastEnh & 252 & 81 & 75.68 & 0.3511 \\
DINOv3-sup & 293 & 40 & 87.99 & 0.1595 \\
DINOv3-sup+ContrastEnh & 289 & 44 & 86.79 & 0.1596 \\
SimLBR & 246 & 87 & 73.87 & 0.3301 \\
SimLBR+ContrastEnh & 270 & 63 & 81.08 & 0.2709 \\
\bottomrule
\end{tabular}%
}
\end{table}

\subsection{Manipulation Localization Additional Results}
\label{supp:manip-results}
\textbf{Performance by Manipulation Type.} We present localization results stratified by manipulation type in Table~\ref{tab:localization-extra}. Here, we see great variability in localization performance as a result of the manipulation type. Using pretrained TruFor and MMFusion checkpoints, \textit{Copy-Move Simple} and \textit{Fully Regenerated} manipulations are the most difficult manipulations to localize, while the easiest are \textit{Splice Simple} manipulations. After training TruFor on GDD, localization of \textit{Copy-Move Object} and \textit{Fully Regenerated} manipulations are greatly improved.

\begin{table*}[t]
    \centering
    \small
    \caption{Evaluation scores across manipulation types on GDD in-domain test split. Scores indicate pixel-level F1  computed using the best threshold per image and using a fixed threshold of 0.5. S-S: splice-simple, S-O: splice-object, S-I: splice-inpainting, CM-S: copy-move-simple, CM-O: copy-move-object, FR: fully-regenerated.}
    \label{tab:localization-extra}
    \begin{tabular}{l cc cc cc cc cc cc cc}
    \toprule
    & \multicolumn{2}{c}{\textbf{S-S}} 
    & \multicolumn{2}{c}{\textbf{S-O}} 
    & \multicolumn{2}{c}{\textbf{S-I}} 
    & \multicolumn{2}{c}{\textbf{CM-S}} 
    & \multicolumn{2}{c}{\textbf{CM-O}} 
    & \multicolumn{2}{c}{\textbf{FR}} 
    & \multicolumn{2}{c}{\textbf{Avg.}} \\
    \cmidrule(lr){2-3} \cmidrule(lr){4-5} \cmidrule(lr){6-7} \cmidrule(lr){8-9} \cmidrule(lr){10-11} \cmidrule(lr){12-13} \cmidrule(lr){14-15}
    \textbf{Method} & Best & Fixed & Best & Fixed & Best & Fixed & Best & Fixed & Best & Fixed & Best & Fixed & Best & Fixed \\
    \midrule
    \rowcolor{blue!8}
    \textbf{Pretrained Checkpoint} & &&&&&&&&&&&&& \\
    TruFor \cite{guillaro2023trufor} & 
    \cellcolor{good}.877 & \cellcolor{good}.823 & 
    \cellcolor{fair}.739 & \cellcolor{poor}.689 & 
    \cellcolor{fair}.769 & \cellcolor{poor}.684 & 
    \cellcolor{good}.804 & \cellcolor{fair}.731 & 
    \cellcolor{verypoor}.495 & \cellcolor{verypoor}.438 & 
    \cellcolor{verypoor}.537 & \cellcolor{verypoor}.354 & 
    \cellcolor{fair}.703 & \cellcolor{poor}.620 \\
    MMFusion \cite{triaridis2024exploring} & 
    \cellcolor{good}.890 & \cellcolor{good}.857 & 
    \cellcolor{fair}.749 & \cellcolor{poor}.663 & 
    \cellcolor{fair}.712 & \cellcolor{verypoor}.584 & 
    \cellcolor{fair}.764 & \cellcolor{fair}.701 & 
    \cellcolor{verypoor}.391 & \cellcolor{verypoor}.304 & 
    \cellcolor{verypoor}.567 & \cellcolor{verypoor}.364 & 
    \cellcolor{poor}.679 & \cellcolor{verypoor}.578 \\
    \rowcolor{blue!8}
    \textbf{Trained on GDD} & &&&&&&&&&&&&& \\
    TruFor & \cellcolor{good}.888 & \cellcolor{good}.867 & \cellcolor{good}.880 & \cellcolor{good}.824 & \cellcolor{great}.917 & \cellcolor{good}.890 & \cellcolor{good}.835 & \cellcolor{good}.802 & \cellcolor{fair}.700 & \cellcolor{poor}.641 & \cellcolor{good}.803 & \cellcolor{fair}.747 & \cellcolor{good}.837 & \cellcolor{fair}.795 \\
    \bottomrule
    \end{tabular}
\end{table*}

\begin{figure}[h!]
    \centering
    \begin{subfigure}{\linewidth}
        \centering
        \includegraphics[width=\linewidth]{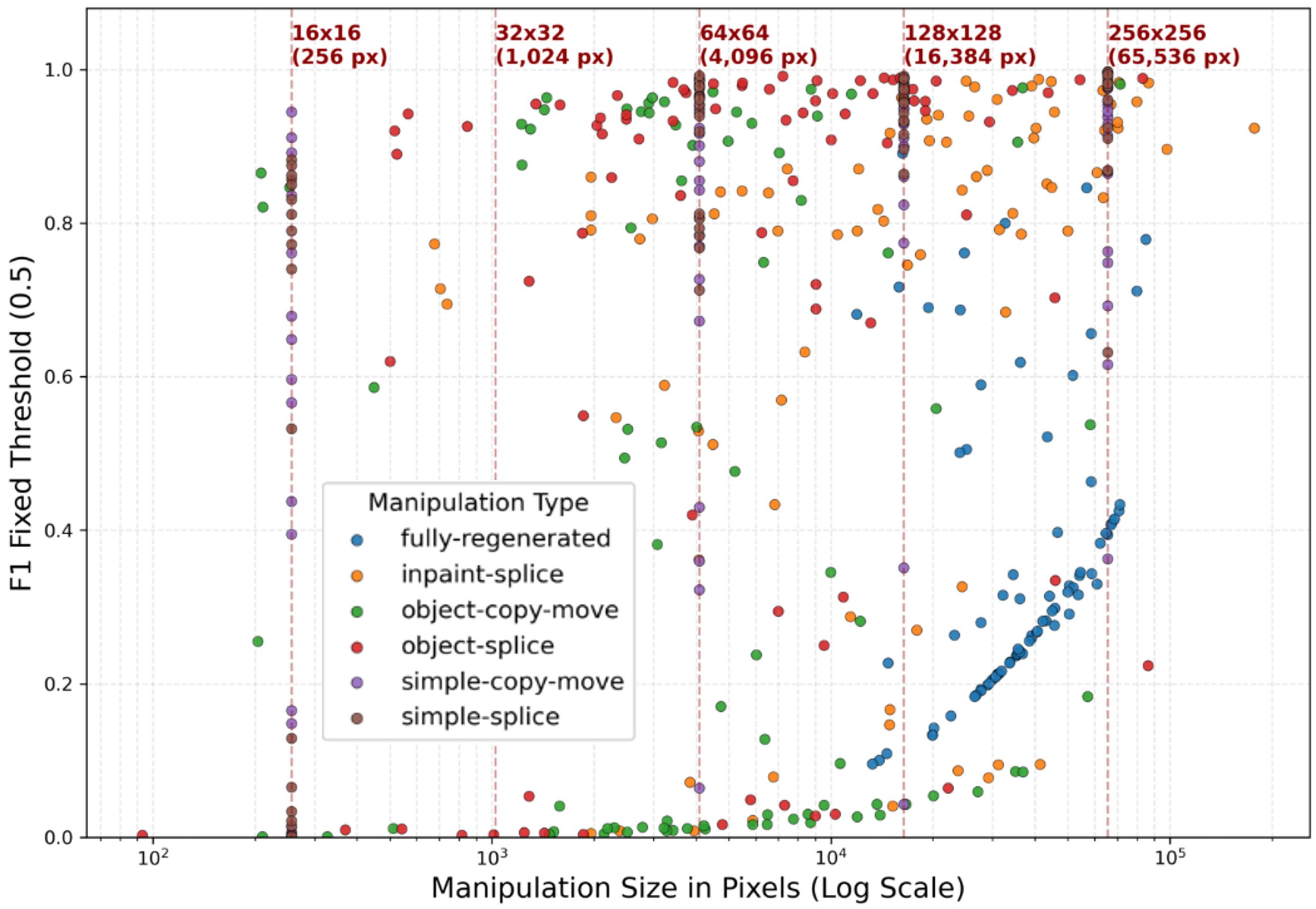}
        \caption{Pretrained TruFor.}
        \label{fig:pretrained-trufor-size}
    \end{subfigure}
    \vspace{0.3cm}
    \begin{subfigure}{\linewidth}
        \centering
        \includegraphics[width=\linewidth]{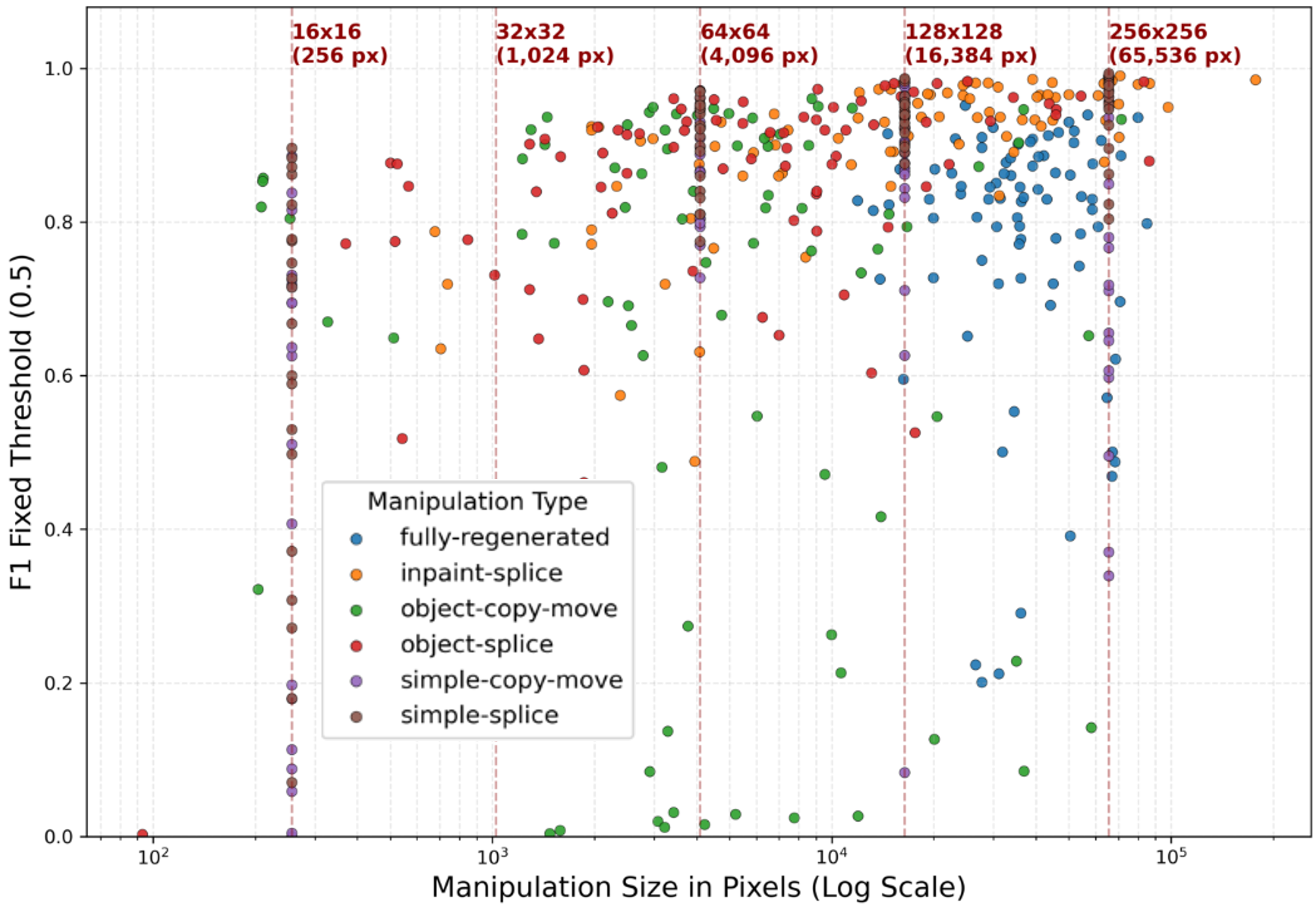}
        \caption{TruFor trained on GDD training split.}
        \label{fig:gdd-trufor-size}
    \end{subfigure}
    \caption{F1 scores (fixed threshold of 0.5) on a per image basis as a function of manipulation size for each image in the GDD In-Domain test split.}
    \label{fig:trufor-score-by-size}
\end{figure}

\textbf{Performance by Manipulation Size.} Additionally, we analyze localization performance as a function of manipulation size, indicated by the number of non-zero pixels in the corresponding ground-truth manipulation mask. Figure~\ref{fig:trufor-score-by-size} presents the per-image F1 scores on GDD In-Domain test split for pretrained TruFor and TruFor trained on the GDD training split.

For pretrained TruFor, manipulation type appears to be more strongly associated with localization performance than manipulation size. However, for fully regenerated manipulations, we observe a distinct trend of improving localization performance with increasing manipulation size. Comparing these results with those obtained using TruFor trained on GDD, we observe a general improvement in localization performance across manipulation sizes. Nevertheless, for both models, manipulation size alone does not appear to be a primary indicator of localization performance. Instead, the results suggest that other factors, including image characteristics, the manipulation procedure, and image content, may play a more significant role in determining localization performance.


\end{document}